\pdfoutput=1

\documentclass[11pt]{article}
\usepackage{fix-cm}
\usepackage[]{acl}

\usepackage{tikz}

\usepackage{xfrac}
 
\usepackage{mathtools}
\mathtoolsset{showonlyrefs,showmanualtags}
\usepackage{amsfonts}
\usepackage{bm}
\usepackage{times}
\usepackage{latexsym}
\usepackage{amsfonts,amsmath,amsthm, amssymb}
\usepackage{graphicx}
\usepackage{subfig}
\usepackage{array}
\usepackage{ifthen}
\usepackage{tabularx}
\usepackage{booktabs}
\usepackage{xcolor}
\usepackage{relsize}
\usepackage{stfloats}
\usepackage{float}
\usepackage{multirow}
\newcommand{\rv}[1]{\mathbf{#1}}
\newcommand{\myvec}[1]{\mathbf{#1}}
\newcommand{\mymat}[1]{\bm{#1}}

\newcommand{\mat}[1]{\bm{#1}}
\newcommand{\subsetswithout}[1]{\textbf{\underline{\textcolor{blue}{#1}}}}
\newcommand{\subsetswith}[1]{\textbf{\underline{\textcolor{purple}{#1}}}}

\usepackage[T1]{fontenc}

\usepackage[utf8]{inputenc}

\usepackage{microtype}

\newcommand{\ignore}[1]{}

\definecolor{mygreenua}{HTML}{F1F5EB}
\definecolor{myredda}{HTML}{FFE6E6}

\newcommand{\uahelper}[1]{\colorbox{myredda}{\smaller$\uparrow$#1}}
\newcommand{\dahelper}[1]{\colorbox{mygreenua}{\smaller$\downarrow$#1}}

\newcommand{\uaghelper}[1]{\colorbox{mygreenua}{\smaller$\uparrow$#1}}
\newcommand{\dabhelper}[1]{\colorbox{myredda}{\smaller$\downarrow$#1}}

\newcommand{\ua}[1]{\ifthenelse{\equal{#1}{0.00} \or \equal{#1}{0.000}}{}{\uahelper{#1}}}
\newcommand{\da}[1]{\ifthenelse{\equal{#1}{0.00} \or \equal{#1}{0.000}}{}{\dahelper{#1}}}
\newcommand{\uag}[1]{\ifthenelse{\equal{#1}{0.00} \or \equal{#1}{0.000}}{}{\uaghelper{#1}}}
\newcommand{\dab}[1]{\ifthenelse{\equal{#1}{0.00} \or \equal{#1}{0.000}}{}{\dabhelper{#1}}}

\newtheorem{assumption}{Assumption}

\newenvironment{itemizesquish}[2]{\begin{list}{\labelitemi}{\setlength{\itemsep}{#1}\setlength{\labelwidth}{#2}\setlength{\leftmargin}{\labelwidth}\addtolength{\leftmargin}{\labelsep}}}{\end{list}}

\author{Shun Shao$^1$ \qquad Yftah Ziser$^{3,4}$ \qquad Zheng Zhao$^2$ \qquad Yifu Qiu$^2$ \\
\textbf{Shay B. Cohen}$^2$ \qquad \textbf{Anna Korhonen}$^1$ \\
$^1$University of Cambridge \qquad $^2$University of Edinburgh \\
\qquad $^3$NVIDIA Research \qquad $^4$University of Groningen \\
\texttt{ss3047@cam.ac.uk} \quad 
\texttt{yziser@nvidia.com} \quad
\texttt{zheng.zhao@ed.ac.uk} \\
\texttt{yifu.qiu@ed.ac.uk} \quad \texttt{scohen@inf.ed.ac.uk} \quad \texttt{alk23@cam.ac.uk}}

\title{Iterative Multilingual Spectral Attribute Erasure}

\begin{document}
\maketitle

\begin{abstract}

Multilingual representations embed words with similar meanings to share a common semantic space across languages, creating opportunities to transfer debiasing effects between languages. However, existing methods for debiasing are unable to exploit this opportunity because they operate on individual languages.
We present Iterative Multilingual Spectral Attribute Erasure (IMSAE), which identifies and mitigates joint bias subspaces across multiple languages through iterative SVD-based truncation. Evaluating IMSAE across eight languages and five demographic dimensions, we demonstrate its effectiveness in both standard and zero-shot settings, where target language data is unavailable, but linguistically similar languages can be used for debiasing. Our comprehensive experiments across diverse language models (BERT, Llama, Mistral) show that IMSAE outperforms traditional monolingual and cross-lingual approaches while maintaining model utility. \footnote{Code is available at \url{https://github.com/jasonshaoshun/IMSAE}.}

\end{abstract}

\section{Introduction}

\begin{figure}[t]
    \centering
    \includegraphics[width=\linewidth]{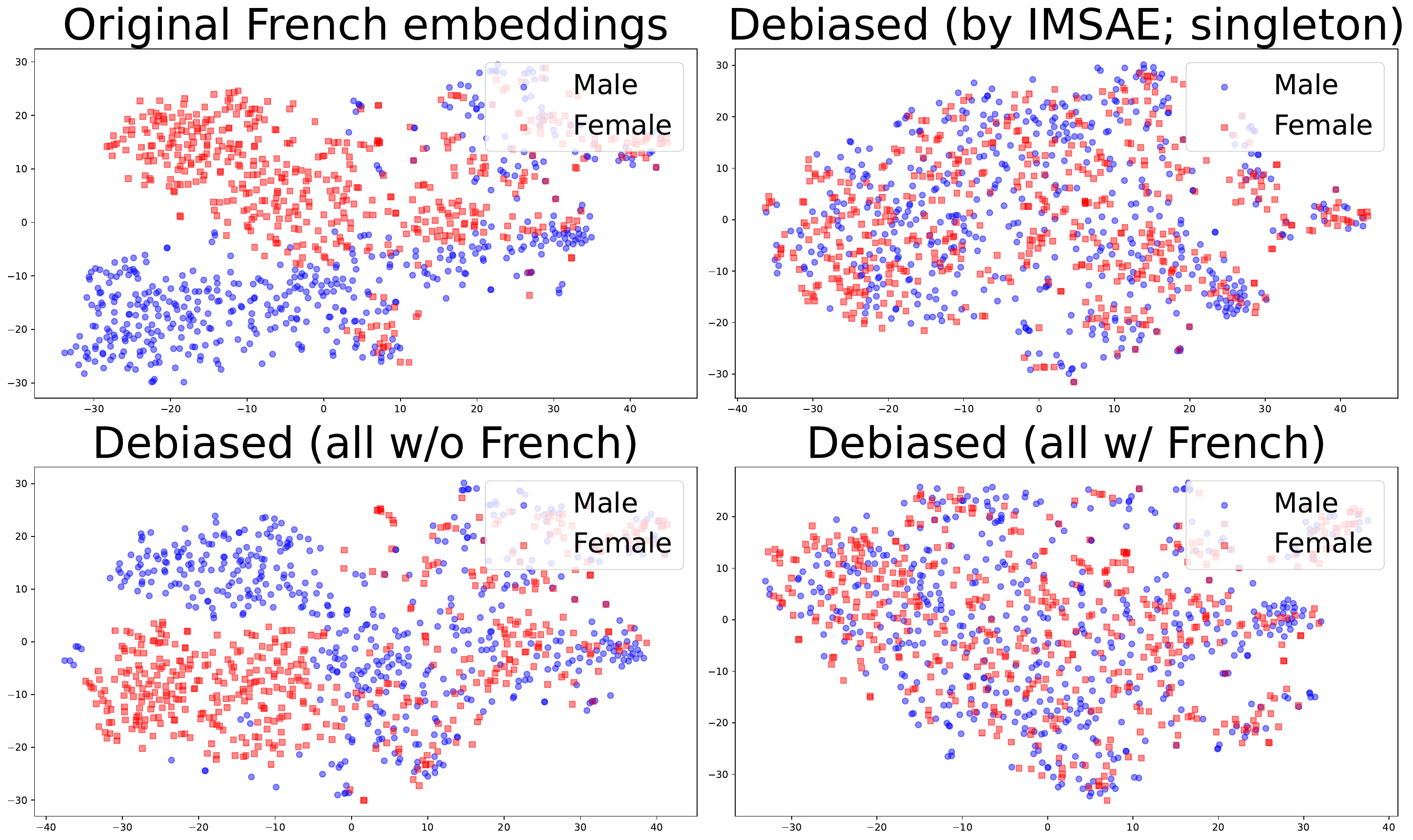}

    \caption{Visualization of gender bias in mBERT French embeddings using t-SNE (blue, male; red, female). Top left: Original embeddings showing clear gender clustering. Top right and bottom: Results after applying IMSAE, demonstrating effective elimination of gender-based patterns through both monolingual (French-only) and cross-lingual (other languages) debiasing approaches. For additional LLM embedding visualizations across different datasets and models, see Figure \ref{fig:dataset-lang}.
}

    \label{fig:t-sne}
\end{figure}

Large language models (LLMs) may make biased decisions or generate unwanted outputs at various stages of training and deployment \cite{hovy-etal-2021, chu2024fairness}, raising ethical concerns in downstream applications \cite{lauscher-etal-2021-sustainable-modular}. Debiasing methods aim to mitigate this by reducing models' reliance on demographic patterns and promoting fairness. Most approaches require pairing texts with authors' protected attributes to remove sensitive information from model representations \cite{reusens2023investigating, liang-etal-2020-monolingual}. However, because large-scale demographic labels are difficult to obtain, most fairness studies have focused exclusively on English datasets \cite{orgad-belinkov-2023-blind}.

To address this, \textbf{multilingual debiasing} leverages transfer learning to mitigate bias in a target language by incorporating information from multiple source languages. Existing approaches typically identify a small set of protected attribute directions, such as gender, in a single source language and apply debiasing to the target language by nullifying projections into these directions \cite{liang-etal-2020-monolingual}. Methods include null space projection \cite{gonen-etal-2022-analyzing}, semantic gender shifting \cite{zhou-etal-2019-examining}, and aligning embeddings across representational spaces \cite{zhao-etal-2020-gender}. This line of work frames multilingual debiasing as a cross-lingual transfer problem: detecting bias in one language and applying the learned debiasing transformation to another. However, state-of-the-art methods remain limited in their ability to fully remove bias through transfer learning \cite{vashishtha-etal-2023-evaluating}, as they fail to account for cultural nuances and demographic variations across languages \cite{talat-etal-2022-reap}.

While prior work has shown the existence of joint gender subspaces across languages \citep{gonen-etal-2022-analyzing}, how to effectively leverage these subspaces for cross-lingual bias mitigation remains an open question. To address this gap, we propose Iterative Multilingual Spectral Attribute Erasure (IMSAE). IMSAE is a structural extension of the monolingual debiasing method by \citet{shao-etal-2022-SAL}. IMSAE is specifically designed to address the issues of directly applying monolingual techniques in multilingual contexts (see \S\ref{section:fullyjoint} for a detailed discussion). IMSAE iteratively debiases subsets of source languages using singular value decomposition (SVD) truncation. These subsets may overlap, and at each step, a shared subspace capturing the guarded attribute across languages is identified and neutralized. Our approach can debias representations without direct access to target language data. IMSAE addresses this challenge in a principled manner by leveraging shared linguistic structures. This effectiveness is visualized in the t-SNE plots in Figure~\ref{fig:t-sne}: while the original embeddings (top left) show clear gender clustering, applying IMSAE (remaining plots) successfully obscures these gender-based patterns. 
IMSAE reduces bias in French embeddings both with (bottom left) and without (bottom right) access to French data, demonstrating its ability to identify and mitigate bias patterns across languages.

 In addition, to properly evaluate our approach, we introduce the Multilingual Stack Exchange Fairness (MSEFair) dataset, which offers two advantages: verified protected attributes and authentic Russian language usage rather than translations. This dataset can serve as a robust evaluation framework for cross-lingual debiasing techniques in linguistically diverse, non-Western linguistic contexts \cite{ramesh-etal-2023-fairness, vashishtha-etal-2023-evaluating}.

We validate IMSAE's effectiveness across eight languages and five demographic dimensions using comprehensive multilingual fairness benchmarks, including our newly developed MSEFair dataset. Our evaluation compares IMSAE against three state-of-the-art post-hoc debiasing methods across diverse language families and multiple model architectures (Llama, Mistral, and BERT), demonstrating consistent performance improvements particularly in zero-shot scenarios.

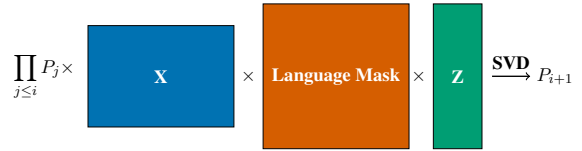
\begin{figure}[t]
    \centering

\scalebox{0.64}{
\begin{tikzpicture}
    \definecolor{cb1}{RGB}{0,114,178}  %
    \definecolor{cb2}{RGB}{213,94,0}   %
    \definecolor{cb3}{RGB}{0,158,115}  %

    \node[draw=black, fill=cb1, text=white, minimum width=3cm, minimum height=2.1cm] (rect1) at (0,0) {\textbf{X}};
    \node[draw=black, fill=cb2, text=white, minimum width=3cm, minimum height=3cm] (rect2) at (3.6,0) {\textbf{Language Mask}};
    \node[draw=black, fill=cb3, text=white, minimum height=3cm, minimum width=1cm] (rect3) at (6.1,0) {\textbf{Z}};

    \node[left of=rect1, node distance=2.4cm] (label_left) {$\displaystyle\prod_{j\le i}P_j \times$};
    \node[right of=rect3, node distance=2cm] (label_right) {$P_{i+1}$};
    \node[right of=rect1, node distance=1.8cm] (label_times) {$\times$};
    \node[right of=rect2, node distance=1.7cm] (label_timess) {$\times$};

    \draw[->, thick] ([xshift=2mm]rect3.east) -- (label_right.west) node[midway, above] {\textbf{SVD}};
\end{tikzpicture} 
}
    \caption{A visualization of IMSAE. A sequence of projections is created using SVD based on the input representations (r.v. $\rv{X}$), the guarded attributes (r.v. $\rv{Z})$ and a language mask that dictates which languages to use.}
    \label{fig:alg-pic}

\end{figure}

\ignore{
\section{Problem Formulation and Notation}

\label{sec:problem}

Given a set of $N$ languages in $\mathcal{L}$ indexed by integers, for each language $l \in \mathcal{L}$, we define three random variables: $(\myvec{x}^{(i)}, \myvec{y}^{(i)}, \myvec{z}^{(i)})$ for $i \in [n_l]$. $\rv{X}_l \in \mathbb{R}^d$ representing input features generated by a multilingual model like m-BERT \cite{devlin-etal-2019-bert} or XLM-RoBERTa \cite{conneau-etal-2020-unsupervised}). $\rv{Y}_l \in \mathbb{R}$ representing target labels, $\rv{Z}_l \in \mathbb{R}^{d'}$ representing sensitive attributes with $d' \le d$. We assume that the samples drawn from these variables have a mean of zero. 
}

\section{Problem Formulation and Notation}

For an integer $n$, we let $[n] = \{ 1, \ldots, n\}$.
Let $\mathcal{L}$ be a set of languages indexed by integers.
Let $\mathcal{L}_s \subseteq \mathcal{L}$ be a subset
of source languages and $\mathcal{L}_t$ be a subset
of target languages. We do not require
$\mathcal{L}_s \cap \mathcal{L}_t = \emptyset$.

We assume a joint multilingual representation
space for the languages, where text from any language in $\mathcal{L}$ can be represented in a vector from that space in $\mathbb{R}^d$. We assume $d$-dimensional
random vectors $\rv{X}_{\ell}$ for any $\ell \in \mathcal{L}$.
These vectors vary over the representations. Recent work has demonstrated effective compression of words and definitions into shared multilingual spaces \cite{chen-etal-2024-cher}.

Our goal is to use representations for languages from $\mathcal{L}_s$
to erase information about a random vector $\rv{Z}$ from representations of languages in $\mathcal{L}_t$. The algorithm is inspired by the SAL algorithm of \newcite{shao-etal-2022-SAL}, and adds a structural component. The SAL algorithm erases protected attribute markings from neural representations by computing a cross-covariance matrix between the input representations and the protected attribute, and then projecting the input representations to the directions which least agree with the protected attribute.

\section{The IMSAE Algorithm}

Our algorithm, IMSAE, is based on the SAL algorithm presented by \newcite{shao-etal-2022-SAL}. Rather than relying on a single projection that is derived from the cross-covariance matrix between input representations and a guarded attribute and that removes information from monolingual input representations, the method creates a sequence of such projections, each corresponding to inputs from a predefined subset of languages.

Figure~\ref{fig:mainalg} shows the IMSAE algorithm. The algorithm uses $n_{\ell}$ samples from the representations of each language in $\mathcal{L}_s$, $\myvec{x}^{(\ell,i)}$ where $\ell \in \mathcal{L}_s$ and $i \in [n_{\ell}]$.
In addition, there are corresponding samples $\myvec{z}^{(\ell,i)}$.
The algorithm also receives as input a sequence of possibly overlapping subsets of $\mathcal{L}_s$, denoted $\mathcal{L}_1, \ldots, \mathcal{L}_m$. Each of these subsets determines one possible way in the sequence to jointly remove bias. The sequence defines a spectrum of how to group languages to erase information.

We explore the interplay between the different languages by grouping together the source languages in various ways. More specifically, we focus on three specific settings.

\begin{itemizesquish}{-0.3em}{0.5em}
    \item Monolingual or cross-lingual (we assume $|\mathcal{L}_t| = 1$): where $m=1$ and $|\mathcal{L}_1| = 1$. This means that we use one language (possibly different from the target language) to erase information.
    \item All subsets without target languages: where $m=2^{|\mathcal{L}_s\setminus \mathcal{L}_t|}-1$, and the $m$ subsets of $\mathcal{L}_s$ vary over all possible subsets of languages (except the empty subset), excluding any target languages.
    \item All subsets with the target languages: where $m=2^{|\mathcal{L}_s|}-1$, and we use all subsets of the source languages except for the empty set.
\end{itemizesquish}

Note that in the above, the order the subsets are in, which is important to consider in the execution of IMSAE, is left underspecified. More of this is discussed in \S\ref{section:global-local}.
We recover (and therefore generalize through a structural extension) the SAL algorithm of \newcite{shao-etal-2022-SAL} when $|\mathcal{L}|=1$, $m=1$ and $\mathcal{L}_s=\mathcal{L}_t=\mathcal{L}$.

\begin{figure}[t]
    \centering
    \includegraphics[width=\linewidth]{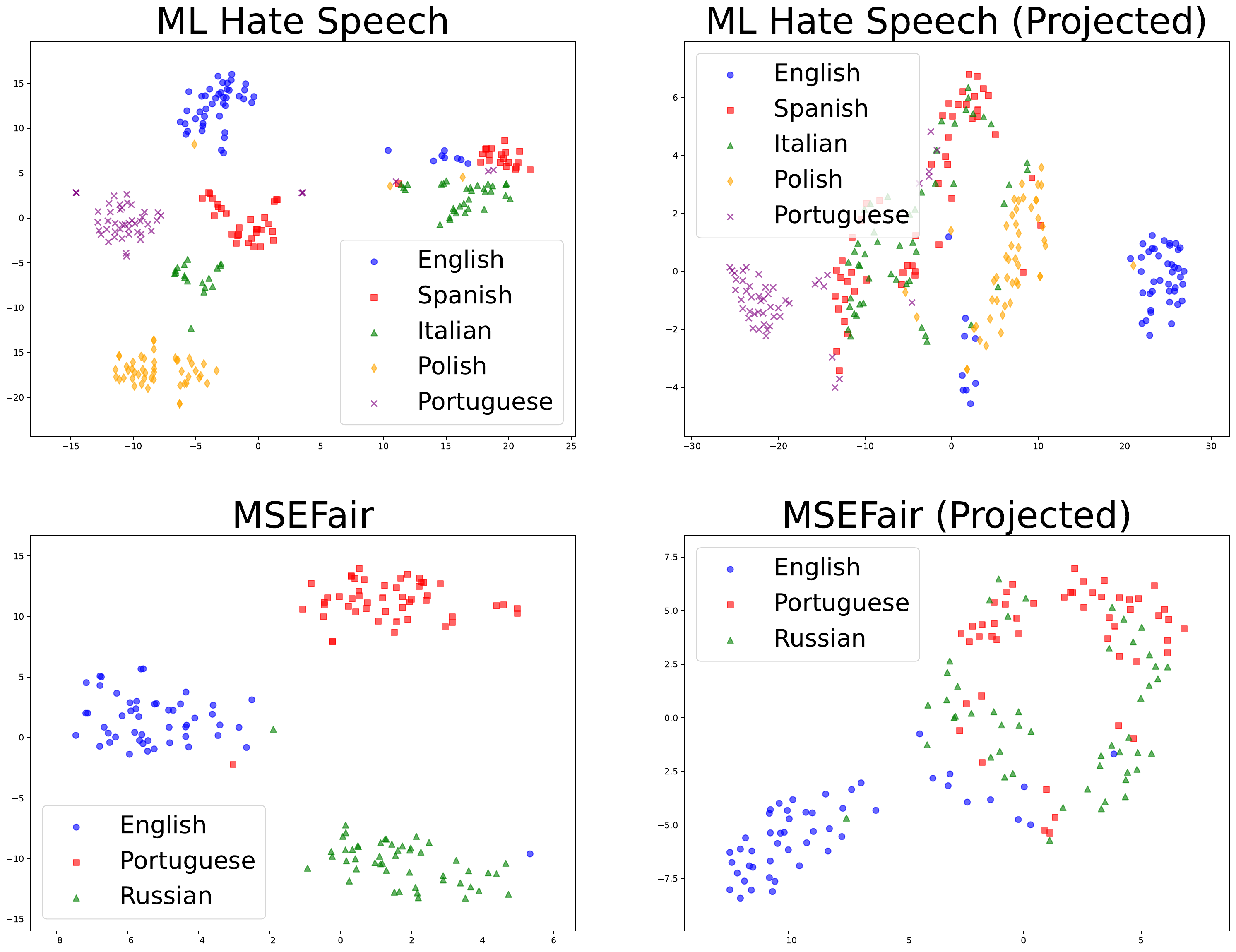}

    \caption{Visualization of Llama embeddings across the Multilingual Hate Speech and Multilingual SEFair datasets. ``Projected'' denotes the projection of the embeddings against the corresponding $\rv{Z}$ variable (using SVD). The projection shows that there is a natural mixture both originally and when projecting the representations against Z. Once we project X onto the direction of the protected attribute, the level of mixture increases, but it remains separable, validating the assumption about mixture of different languages. }
    \label{fig:dataset-lang}

\end{figure}

Our algorithm has a matrix formulation, as an iterative projection algorithm with an SVD on a masked version of the cross-covariance matrix. The subsets of $\mathcal{L}$ essentially select examples from the entire set of examples in the pool, as depicted in Figure~\ref{fig:alg-pic}. See also similar discussion by \newcite{osborne-etal-2016-encoding}.

In practice, we set $k=2$ for all experiments since binary protected attributes (e.g., male/female) yield a rank-2 cross-covariance matrix, eliminating the need for hyperparameter tuning. All evaluation uses identical LogisticRegression settings (max\_iter=1000) to ensure consistent comparison across methods.

\begin{figure}[t]
\framebox{\parbox{0.96\linewidth}{

{\bf Inputs:} Samples $\myvec{x}^{(\ell,i)}$, and $\myvec{z}^{(\ell,i)}$, $\ell \in \mathcal{L}_s$ and $i \in [n_\ell]$, $\mathcal{L}_1, \ldots, \mathcal{L}_m$ subsets of $\mathcal{L}_s$.

{\bf Algorithm:} (erase information based on the samples sequentially)

\medskip

Initialize $\mat{P}^{\ast}$ to be the identity matrix.

\medskip

Repeat the following for $j \in [m]$:

\begin{itemize}

\item Calculate $\mat{\Omega}$ as follows:
\begin{align}
\mat{\Omega} \leftarrow & \sum_{\ell \in \mathcal{L}_j} \sum_{i=1}^{n_\ell} \myvec{x}^{(\ell,i)} (\myvec{z}^{(\ell,i)})^{\top}, \\
\mat{\Omega} \leftarrow & \mat{P}^{\ast} \mat{\Omega}.
\end{align}

\item Calculate SVD on $\mat{\Omega}$ to calculate $(\mat{U}, \mat{\Sigma}, \mat{V})$ with bottom $k$ left singular vectors being $\mat{U}$.

\item Update $\mat{P}^{\ast} \leftarrow \mat{U} \mat{U}^{\top} \mat{P}^{\ast}$.

\end{itemize}

{\bf Return:} The erasure matrix $\mat{P}^\ast$.
}}
\caption{The IMSAE algorithm. IMSAE targets attribute-specific bias through cross-covariance between representations $\rv{X}$ and protected attributes $\rv{Z}$, using SVD to remove only attribute-correlated directions.}
\label{fig:mainalg}
\end{figure}

\subsection{Order Sensitivity in Sequential Projections}
\label{section:global-local}

The final erasure matrix $\mat{P}^{\ast}$ is a product of projection matrices, and such matrix multiplication is not commutative in general. Therefore, the order in which these projections are applied matters. We consider two orderings: (a) Global-then-Local: First apply a projection using all languages $\mathcal{L}_s$ to identify and remove shared bias directions, followed by language-specific projections for each $\ell \in \mathcal{L}_s$; (b) Local-then-Global: the reverse.

The global-first approach may capture broad bias patterns that are diluted when looking at languages individually, while the local-first approach may better preserve language-specific nuances. In our empirical analysis, we found that both orderings achieve similar debiasing performance across our evaluation tasks. This suggests that while the mathematical difference exists, the practical impact is limited--likely because the core bias directions are relatively stable regardless of the order of removal. Therefore, we report the Global-then-Local ordering in our main experiments.

\section{Justification and the Fully Joint Baseline with Its Drawback}
\label{section:fullyjoint}
The following can be skimmed through in a first read.
We proceed with providing an intuition for a justification of
the IMSAE algorithm, especially when a target language is missing from training. Central to our analysis is a two-fold assumption:

\begin{assumption}
 There exists $C$ \textbf{meta-linguistic} random vectors, $p(\overline{\rv{X}}_i)$ and a latent variable $\rv{H}$ such that for any  language $\ell \in \mathcal{L}$
 there exists constants  $\zeta_{\ell,h,1}, \ldots, \zeta_{\ell,h,C}$ for any $h$ value of $\rv{H}$ where:

 \begin{equation}
 \mathbb{E}[\rv{X}_{\ell} \mid h] = \sum_{i} \zeta_{\ell, h,i} \mathbb{E}[\overline{\rv{X}}_{i} \mid h],
 \end{equation}
\label{assum1} 

\noindent and $\rv{X}_{\ell}$ and $\rv{Z}$ are conditionally independent given $\rv{H}$.
\end{assumption}

For the condition on the latent variable $\rv{H}$, see \newcite{shao2023erasure}. The reference to meta-linguistic variables implies
that any representation of a specific language (at least in expectation) can be
represented as a combination of representations of some core prototypical meta-linguistic vectors. Figure~\ref{fig:dataset-lang} demonstrates how various languages may cluster in the embedding space of two datasets (Hate Speech and MSEFair, discussed in \S\ref{exp:hate-speech} and \S\ref{section:msefair}), with several centroids. The mixture of centroids is even more noticeable when considering the projection of $\rv{X}$ against $\rv{Z}$ (Projected). We could have a stronger mixture condition on the probability distribution, but it is sufficient to require the expectation condition for our needs. We expect $C < \min\{ |\mathcal{L}|, d \}$.

We also assume a set of coefficients over $\rv{X}$, the r.v. vector from all languages, such that $p(\rv{X} \mid h) = \sum_{h,\ell} \mu_{h,\ell} p(\rv{X}_\ell \mid h)$. This is a reasonable assumption, as all it implies is that for a specific subset $\mathcal{L}_j$, the resulting vector $\rv{X}$ on which we operate has a mixture distribution over the languages based on the relative frequency of each language in the data \cite{zhou-etal-2019-density,zhao-etal-2023-joint}. Such coefficient could be zero, for example, if a language $\ell$ does not appear in $\mathcal{L}_j$.\footnote{Empirically, each $\mathcal{L}_j$ defines a different distribution over $\rv{X}$ because of the different frequencies of the languages.}

The implication of Assumption~\ref{assum1}  together with the mixture of languages requirement is that for an $\mathcal{L}_j$, the covariance $\mathbf{E}[\rv{X}\rv{Z}^{\top}]$ between a vector $\rv{X}$ from that distribution and $\mathbf{Z}$ can be rewritten as:

\begin{align}
\sum_{\ell} \sum_h \sum_i \mu_{h,\ell} \zeta_{h,\ell,i} \mathbf{E}[\overline{\mathbf{X}}_{i} \mid h] \cdot \mathbf{E}[ \mathbf{Z}^{\top} \mid h].
\end{align}

Combining the indices $(h,\ell) = k$ together, this means that for any $\mathcal{L}_j$, there exist coefficients $\eta_{k,i}$ such that:

\begin{equation}
\mathbf{E}[\mathbf{X} \mathbf{Z}^{\top}] = \sum_{k,i} \eta_{k,i} \mathbf{E}[\overline{\mathbf{X}}_{i} \mid h] \mathbb{E}[ \mathbf{Z}^{\top} \mid h].
\end{equation}

Each specific subset of languages, as presented by the algorithm in Figure~\ref{fig:mainalg}, provides a different set of coefficients $\eta_{k,i}$. This is also true for the target language, even when missing from the data. Therefore, running the algorithm provides an erasure matrix that iteratively erases directions that correspond to different coefficients $\eta_{k,i}$. This means that the final matrix is more robust to the specific
coefficient of the left-out target language, which is also based on some combination of $\eta_{k,i}$, hopefully targeted by one of $\mathcal{L}_j$.\footnote{This leaves an idea for further exploration, where not only subsets of the languages are taken, but also subsets of different sizes of the data for these languages, yielding different coefficients.}

We note that while it may seem like we are identifying the intersection of the projections ranges, that is not the case, and intersecting them would require, for example, using Ben-Israel’s algorithm or using repeated projection \cite{ben2015projectors}. We did not experiment with these approaches, though it might be worth considering that in future work.

\paragraph{The FullyJoint Baseline}
A straightforward baseline to erase information from representations across multiple languages is to concatenate all the input representations from the different languages and their corresponding guarded attributes and feed them to an erasure algorithm. For example, running IMSAE with $m=1$, and $\mathcal{L}_1 = \mathcal{L}$. This one-shot reduction may be less effective than IMSAE in its full generality, which erases information iteratively on a per-language subset. We refer to this baseline as \emph{FullyJoint}. 

We turn to explain a specific case of the analysis above with FullyJoint, illustrating why $m > 1$ is needed. If $\rv{Z}$ is essentially a scalar (represented possibly as a one-hot vector or otherwise), the rank of any of the covariance matrices between $\rv{X}$ and $\rv{Z}$ is 1. With the FullyJoint baseline, $\mathcal{L}_1$ corresponds to a specific value of $\eta_{k,i}$ and a single erasure direction $\myvec{u}$. These quantities do not have to coincide with the values of using, for example, a singleton subset containing one language.
Some residual association between the input representations of each language and the protected attributes may remain, allowing a linear classifier to predict the protected attribute. Thus, by erasing the attribute-related direction iteratively through different subsets (unlike FullyJoint), for example, including singletons, we ensure a more robust erasure.

\section{Multilingual Stack Exchange Fairness Dataset}
\label{sec:MSEFair}
\label{section:msefair}
This section introduces MSEFair, a challenging dataset curated to support experimentation with non-English languages, which are often more difficult to transfer to.

\subsection{Motivation}
Previous debiasing research has focused on Anglo-centric text, with limited attention to non-Western contexts \cite{ramesh-etal-2023-fairness}. A key open question is whether it is possible to effectively debias across distant languages. While \citet{vashishtha-etal-2023-evaluating} extended DisCo \cite{webster2021measuring} to Indian languages through human translation, this approach fails to capture culture-specific bias \cite{neveol-etal-2022-french}. Moreover, existing studies have largely focused on specific tasks like sentiment analysis and profession prediction. To address these issues, we introduce the Multilingual Stack Exchange dataset. This dataset offers two key advantages: it contains verified protected attributes; and, it represents authentic Russian language use rather than translations, providing a more reliable testbed in non-Western contexts.

\subsection{Data Collection for MSEFair}
In addition to its English-language sites, the Stack Exchange platform also hosts localized versions of its most popular site, Stack Overflow in Russian and Portuguese. We curated posts (questions and answers) from the users of those websites. We use the user reputation as the protected attribute for predicting post helpfulness.
We classify users in the top 1\% of reputation as high-reputation users and those in the bottom 98\% as low-reputation users.\footnote{The reputation distribution in the Stack Exchange network is highly skewed, where users with a reputation of 1 are in the top 80\% of users in Stack Overflow, for example - \url{https://tinyurl.com/2uhn7u52}.} As the Stack Exchange platform provides user reputation, we consider it a reliable protected attribute. For the helpfulness prediction, we classify posts with four or more upvotes as helpful and posts with zero upvotes as not helpful. We aim to classify posts as helpful or not ($x$'s) regardless of their authors reputation ($z$'s). Another challenge posed by this dataset is the correlation between the protected attribute and the downstream task label, as upvotes are a major factor in determining user reputation on Stack Exchange platforms. This correlation makes it difficult to remove information about the protected attribute without discarding information essential for the downstream task. The dataset statistics are provided in Table \ref{tab:stats}.

\begin{table}[h!]
\centering
\begin{tabular}{lrrrr}
\hline
\textbf{Name} & \textbf{Train} & \textbf{Val} & \textbf{Test} & \textbf{Total} \\
\hline
English & 4,058 & 2,029 & 14,205 & 20,292 \\
Russian & 2,370 & 1,186 & 8,300 & 11,856 \\
Portuguese & 1,670 & 835 & 5,847 & 8,352 \\
\hline
\end{tabular}
\caption{Statistics for the MSEFair datasets.}
\label{tab:stats}
\end{table}

\section{Experiments}
We explore the effectiveness of IMSAE in two scenarios. First, when the target language is included in the training set, we demonstrate that IMSAE, with additional languages, yields better results compared to monolingual debiasing. Second, when the target language is not part of the training set, we show that using multiple source languages via IMSAE outperforms the typical approach of conducting cross-lingual debiasing with a single source language. 
We experiment with three tasks: profession prediction (\S\ref{exp:fair-profession}), hate speech recognition (\S\ref{exp:hate-speech}) and helpfulness prediction (\S\ref{exp:sefair}). For more information on the data sets split and statistics, see Appendix \ref{appendix:biasbios-dataset}. %

\paragraph{Evaluation Metrics}
Our ultimate goal is to reduce bias while ensuring high downstream task performance. We measured disparities in classifier performance across different protected groups to quantify bias in language models. For example, we compared the performance of male and female biographies in our profession prediction task. Specifically, we used the True Positive Rate Gap (TPR-Gap), which calculates the difference in true positive rates between demographic groups, conditioned on the true class. A lower TPR-Gap indicates greater fairness, as it suggests the model performs similarly for both gender groups when predicting professions. We use accuracy to measure the downstream task performance. 

\paragraph{Models}
We benchmark against eight open-weight models. Our chosen models represent varied architectures, parameter sizes, multilingual capabilities, performance levels, and bias tendencies: Multilingual BERT \cite{devlin-etal-2019-bert}, Llama 3 \cite{grattafiori2024Llama3herdmodels}, Llama 3.1 \cite{meta2024Llama31}, Llama 3.2 \cite{meta2024Llama32}, Mistral 7B \cite{jiang2023mistral7b} and Mistral Nemo \cite{mistral2024nemo}. Due to page constraints, we focus on three representative models in the main paper: mBERT, Llama-3.1, and Mistral-7B-Instruct-v0.3, with comprehensive results for all eight models provided in Appendix \ref{appendix:biasbios}. In our methodology, we used the final hidden state representation from the model's last layer as input for probing experiments on both the primary and bias detection tasks, enabling evaluation of bias before and after debiasing.

\subsection{Fair Profession Prediction}
\label{exp:fair-profession}

\paragraph{Task and Data}
We use the Multilingual BiasBios dataset \cite{zhao-etal-2020-gender}, an extension of the BiasBios dataset \cite{de2019bias} with French, Spanish, and German biographies. The dataset was constructed by extracting biographies from Common Crawl using the template ``NAME is an OCCUPATION-TITLE''. Each biography is annotated with gender and profession labels.

\subsubsection{Crosslingual Debiasing Results}

\begin{table}[t]
\centering
\resizebox{\columnwidth}{!}{
\begin{tabular}{lrr  rr  rr  rr}
\toprule
Target & \multicolumn{2}{c}{Baseline} & \multicolumn{2}{c}{SAL (EN)} & \multicolumn{2}{c}{INLP (EN)} & \multicolumn{2}{c}{SentenceDebias (EN)} \\
       & Main   & TPR-Gap   & \multicolumn{1}{c}{Main}    & \multicolumn{1}{c}{TPR-Gap}   & \multicolumn{1}{c}{Main}    & \multicolumn{1}{c}{TPR-Gap}   & \multicolumn{1}{c}{Main}    & \multicolumn{1}{c}{TPR-Gap}   \\
       \multicolumn{9}{l}{Multilingual BERT} \\
EN & 80.5 & 15.4 & \dab{0.1} 80.4 & \da{1.9} 13.5 & 80.5 & \da{0.2} 15.2 & \dab{0.2} 80.3 & \da{0.1} 15.3 \\
DE & 77.7 & 27.6 & \uag{0.1} 77.8 & \da{4.5} 23.1 & \uag{0.1} 77.8 & \da{2.2} 25.4 & \dab{0.1} 77.6 & \da{2.4} 25.2 \\
FR & 72.7 & 22.8 & \dab{0.1} 72.6 & \da{0.8} 22.0 & 72.7 & \da{0.5} 22.3 & \uag{0.1} 72.8 & \da{0.5} 22.3 \\
\midrule 
\multicolumn{9}{l}{Llama-3.1-8B} \\
EN & 81.1 & 13.7 & \dab{2.1} 79.0 & \da{0.7} 13.0 & \dab{0.7} 80.4 & \ua{0.2} 13.9 & \dab{0.5} 80.6 & \da{0.4} 13.3 \\
DE & 79.8 & 26.8 & \dab{0.3} 79.5 & \ua{0.2} 27.0 & 79.8 & 26.8 & \dab{0.2} 79.6 & \ua{0.2} 27.0 \\
FR & 72.5 & 25.0 & \uag{0.1} 72.6 & \ua{0.1} 25.1 & 72.5 & \ua{1.1} 26.1 & \dab{0.1} 72.4 & \ua{1.0} 26.0 \\
\midrule 
\multicolumn{9}{l}{Mistral-7B-Instruct-v0.3} \\
EN & 80.5 & 14.0 & \dab{2.7} 77.8 & \da{1.3} 12.7 & \dab{0.2} 80.3 & \da{0.2} 13.8 & \dab{0.3} 80.2 & \da{0.2} 13.8 \\
DE & 77.3 & 23.3 & \dab{0.2} 77.1 & \ua{0.3} 23.6 & \dab{0.1} 77.2 & 23.3 & \dab{0.2} 77.1 & \da{0.1} 23.2 \\
FR & 71.6 & 23.1 & \uag{0.2} 71.8 & \da{1.4} 21.7 & \dab{0.2} 71.4 & \da{0.4} 22.7 & \dab{0.2} 71.4 & \da{2.1} 21.0 \\

\bottomrule
\end{tabular}
}
\caption{Evaluation of post-hoc debiasing methods on the multilingual BiasBios dataset. The main task is profession prediction, while the TPR-Gap (True Positive Rate Gap) between males and females demonstrates the extrinsic bias in downstream tasks.}
\label{tab:cross-bias}
\end{table}

We evaluated three erasure approaches: null-space projection (INLP; \citealt{ravfogel-etal-2020-null}), SVD-based erasure (SAL; \citealt{shao-etal-2022-SAL}), and SentenceDebias \cite{liang-etal-2020-towards}. Table~\ref{tab:cross-bias} presents the results using English as the source language, while the full results show similar trends. While all methods maintain strong downstream task performance in mBERT, their effectiveness varies across model architectures. SAL demonstrates superior bias reduction in LLMs, with average results of 2.4\% and 0.6\% in mBERT and Mistral respectively. Based on these findings, we selected SAL as our reference and further report results on IMSAE as its structural variant in different settings.

\subsubsection{IMSAE Results}

\begin{table*}[t]
\centering
\resizebox{\textwidth}{!}{
\begin{tabular}{lrr  rr  rr  rr  rr  rr}
\toprule
Target & \multicolumn{2}{c}{Baseline} & \multicolumn{2}{c}{SAL (Monolingual)} & \multicolumn{2}{c}{SAL (Avg-Crosslingual)} & \multicolumn{2}{c}{\subsetswithout{IMSAE (FullyJoint)}} & \multicolumn{2}{c}{\subsetswithout{IMSAE (Subsets w/o)}} & \multicolumn{2}{c}{\subsetswith{IMSAE (Three-Subsets)}} \\
       & Main   & TPR-Gap   & Main    & TPR-Gap    & Main    & TPR-Gap    & Main    & TPR-Gap    & Main    & TPR-Gap    & Main    & TPR-Gap \\
\midrule

\multicolumn{13}{l}{mBERT-uncased} \\
EN & 80.5 & 15.4 & \dab{0.1} 80.4 & \da{1.9} 13.5 & 80.5 & \ua{0.4} 15.8 & \dab{0.1} 80.4 & \da{1.2} 14.2 & 80.5 & \ua{0.2} 15.6 & \dab{0.1} 80.4 & \da{1.8} 13.6 \\
DE & 77.7 & 27.6 & \dab{0.3} 77.4 & \da{0.3} 27.3 & \uag{0.1} 77.8 & \da{2.2} 25.4& \uag{0.2} 77.9 & \da{4.6} 23.0 & \uag{0.1} 77.8 & \da{1.9} 25.7 & \dab{0.1} 77.6 & \da{2.2} 25.4 \\
FR & 72.7 & 22.8 & \dab{0.5} 72.2 & \da{3.4} 19.4 & 72.7 & \da{0.7} 22.1 & \dab{0.2} 72.5 & \da{1.5} 21.3 & \uag{0.1} 72.8 & \da{0.6} 22.2 & \dab{0.5} 72.2 & \da{3.2} 19.6 \\
\midrule
\multicolumn{13}{l}{Llama-3.1-8B} \\
EN & 81.1 & 13.7 & \dab{2.1} 79.0 & \da{0.7} 13.0 & \dab{0.8} 80.3 & \da{0.1} 13.6 & \dab{2.0} 79.1 & \da{1.1} 12.6 & \dab{1.1} 80.0 & \da{0.5} 13.2 & \dab{2.1} 79.0 & \da{0.5} 13.2 \\
DE & 79.8 & 26.8 & \dab{0.3} 79.5 & \da{5.2} 21.6 & \dab{0.2} 79.6 & \ua{0.4} 27.2 & \dab{0.2} 79.6 & \ua{0.3} 27.1 & \dab{0.4} 79.4 & \ua{0.4} 27.2 & \dab{0.2} 79.6 & \da{4.6} 22.2 \\
FR & 72.5 & 25.0 & \dab{0.1} 72.4 & \da{5.7} 19.3 & \uag{0.1} 72.6 & \da{0.4} 24.6 & \dab{0.1} 72.4 & \da{0.4} 24.6 & 72.5 & \da{0.5} 24.5 & \dab{0.1} 72.4 & \da{3.7} 21.3 \\
\midrule
\multicolumn{13}{l}{Mistral-7B-Instruct-v0.3} \\
EN & 80.5 & 14.0 & \dab{2.7} 77.8 & \da{1.3} 12.7 & \dab{0.5} 80.0 & 14.0 & \dab{2.8} 77.7 & \da{0.8} 13.2 & \dab{0.7} 79.8 & \da{0.1} 13.9 & \dab{2.8} 77.7 & \da{1.3} 12.7 \\
DE & 77.3 & 23.3 & 77.3 & 23.3 & \dab{0.2} 77.1 & \ua{0.5} 23.8 & \dab{0.2} 77.1 & \ua{0.3} 23.6 & \dab{0.3} 77.0 & \da{0.8} 22.5 & \dab{0.3} 77.0 & \ua{0.4} 23.7 \\
FR & 71.6 & 23.1 & \uag{0.2} 71.8 & \da{4.9} 18.2 & \uag{0.1} 71.7 & \da{1.2} 21.9 & 71.6 & \da{1.2} 21.9 & \dab{0.1} 71.5 & \da{1.8} 21.3 & \uag{0.1} 71.7 & \da{5.1} 18.0 \\

\bottomrule
\end{tabular}
}
\caption{Evaluation of demographic bias mitigation on the multilingual BiasBios dataset. Main shows profession prediction accuracy, while TPR-GAP shows true positive rates between different demographic groups. Results compare Baseline, Monolingual (target language only), Average - Crosslingual, IMSAE on FullyJoint (\ref{section:fullyjoint}), IMSAE on Two-Subsets-Without (excluding target language) and IMSAE Three-Subsets (using all languages).}
\label{tab:biasbios}
\end{table*}

\paragraph{With Target Language} Consider Table~\ref{tab:biasbios}.
For two out of three LMs, incorporating information from additional languages using IMSAE (``Three-Subsets'') further reduces bias on average compared to relying solely on the target language (``Monolingual''). While the FullyJoint approach slightly outperforms IMSAE for mBERT, IMSAE significantly outperforms FullyJoint for both LLMs. Regarding downstream task performance, all methods perform well, with only a relatively small drop in accuracy.

\paragraph{Without Target Language}

We observe that two out of three LMs, IMSAE (``Subsets w/o'') outperforms the average cross-lingual debiasing method in terms of debiasing. Both approaches minimally affect main-task performance while effectively reducing bias. See also results in Appendix~\ref{app:multilingual-bb}.

\subsection{Hate Speech Recognition}
\label{exp:hate-speech}

The Multilingual Twitter Hate Speech corpus \cite{huang-etal-2020-multilingual} provides data across multiple demographic dimensions (gender, race, country, and age) and languages (English, Spanish, Italian, Polish, and Portuguese). The demographic attributes are derived directly from user profiles, making the demographic attributes reliable.

\paragraph{Task and Data}

For training, we used approximately 32,000, 1,900, 1,600, 6,800, and 800 samples from these respective languages. The test set sizes range from 20\% to 25\% of the corresponding training set sizes.  Some results are excluded due to severe class imbalance in certain subsets. Complete data statistics are provided in Tables \ref{appendix:hate-speech-size} and \ref{appendix:hate-speech-data-distribution} in the appendix. \citet{huang-etal-2020-multilingual} labeled the datasets by inferring the author attributes from user profiles across four demographic dimensions: gender (male/female), race (white/non-white), age (young/old), and country (US/non-US). The primary labels assigned to each tweet indicate whether it contains hate speech or not.

\subsubsection{Results}

\begin{table*}[t]
\centering
\resizebox{\textwidth}{!}{
\begin{tabular}{lrr  rr  rr  rr  rr  rr rr}
\toprule

Target & \multicolumn{2}{c}{Baseline} & \multicolumn{2}{c}{Monolingual}  & \multicolumn{2}{c}{SAL (Avg-Crosslingual)}  & \multicolumn{2}{c}{\subsetswithout{IMSAE (FullyJoint)}} & \multicolumn{2}{c}{\subsetswithout{IMSAE (Subsets w/o)}} & \multicolumn{2}{c}{\subsetswith{IMSAE (Five-Subsets)}}  \\

& Main & TPR-Gap & Main & TPR-Gap & Main & TPR-Gap & Main & TPR-Gap & Main & TPR-Gap & Main & TPR-Gap \\

\midrule

\multicolumn{13}{l}{\textbf{Race}} \\

EN & 86.8 & 4.1 & \dab{1.7} 85.1 & \da{1.3} 2.8 & 86.8 & \ua{0.1} 4.2 & \dab{0.3} 86.5 & \ua{0.8} 4.9 & \dab{0.1} 86.7 & \ua{1.9} 6.0 & \dab{2.0} 84.8 & \da{4.1} 0.0 \\
ES & 63.7 & 10.2 & \uag{0.4} 64.1 & \da{0.1} 10.1 & \uag{0.4} 64.1 & \da{0.2} 10.0 & \uag{0.4} 64.1 & \da{0.1} 10.1 & 63.7 & \da{9.2} 1.0 & \dab{0.3} 63.4 & \da{1.7} 8.5 \\
IT & -    & -    & -    & -    & -    & -    & -    & -    & -    & -  & - & - \\
PL & 91.3 & 6.2 & 91.3 & \da{1.2} 5.0 & 91.3 & \da{0.3} 5.9 & 91.3 & 6.2 & 91.3 & \da{4.7} 1.5 & \dab{0.2} 91.1 & \da{6.2} 0.0 \\
PT & 61.3 & 1.0 & \dab{0.6} 60.7 & \ua{0.1} 1.1 & \dab{0.7} 60.6 & \ua{1.0} 2.0 & 61.3 & \ua{1.4} 2.4 & \uag{0.7} 62.0 & \ua{5.0} 6.0 & \uag{1.3} 62.6 & \ua{11.3} 12.3 \\

\midrule
\multicolumn{11}{l}{\textbf{Gender}} \\

EN & 86.7 & 4.4 & 86.7 & \ua{0.7} 5.1 & 86.7 & 4.4 & 86.7 & 4.4 & 86.7 & \da{4.3} 0.1 & \dab{0.1} 86.6 & \da{4.3} 0.1 \\
ES & 63.7 & 3.6 & \uag{0.4} 64.1 & \da{0.2} 3.4 & \dab{0.1} 63.6 & 3.6 & \uag{0.9} 64.6 & \ua{0.1} 3.7 & 63.7 & \ua{2.3} 5.9 & \dab{0.3} 63.4 & \ua{0.7} 4.3 \\
IT & 68.4 & 2.1 & 68.4 & \da{0.6} 1.5 & \dab{0.3} 68.1 & \da{0.3} 1.8 & \dab{0.2} 68.2 & \da{1.7} 0.4 & \uag{0.3} 68.7 & \ua{1.0} 3.1 & \dab{0.5} 67.9 & \da{0.8} 1.3 \\
PL & 88.2 & 11.6 & \dab{0.2} 88.0 & \da{8.8} 2.8 & 88.2 & \da{0.1} 11.5 & 88.2 & 11.6 & \dab{0.1} 88.1 & \ua{0.4} 12.0 & \dab{0.2} 88.0 & \da{11.6} 0.0 \\
PT & 61.3 & 12.0 & \uag{0.7} 62.0 & \ua{1.4} 13.4 & 61.0 & 12.6 & \dab{1.2} 60.1 & \ua{1.8} 13.8 & \uag{1.3} 62.6 & \da{5.2} 6.8 & \uag{3.1} 64.4 & \ua{2.4} 14.4 \\

\midrule
\multicolumn{11}{l}{\textbf{Age}} \\

EN & 86.7 & 9.1 & \uag{0.3} 87.0 & \ua{0.4} 9.5 & \dab{0.2} 86.5 & \da{0.1} 9.0 & 86.7 & \ua{0.2} 9.3 & 86.7 & \da{9.0} 0.1 & \uag{0.2} 86.9 & \da{9.0} 0.1 \\
ES & 63.7 & 12.9 & \dab{0.3} 63.4 & \da{0.5} 12.4 & 63.7 & \ua{0.5} 13.4 & \dab{0.3} 63.4 & \da{0.4} 12.5 & \uag{0.2} 63.9 & \da{8.0} 4.9 & \uag{0.4} 64.1 & \da{6.3} 6.6 \\
IT & 68.2 & 3.6 & \dab{0.3} 67.9 & \ua{0.3} 3.9 & 68.2 & \ua{0.1} 3.7 & \dab{0.3} 67.9 & \ua{0.3} 3.9 & \dab{1.2} 67.0 & \da{2.8} 0.8 & \dab{0.5} 67.7 & \da{0.7} 2.9 \\
PL & 91.3 & 8.8 & \dab{0.9} 90.4 & \da{1.9} 6.9 & 91.3 & \da{0.7} 8.1 & \dab{0.1} 91.2 & \da{1.3} 7.5 & \dab{0.4} 90.9 & \da{5.0} 3.8 & \dab{0.9} 90.4 & \da{8.8} 0.0 \\
PT & 61.3 & 17.6 & \dab{0.6} 60.7 & \ua{1.5} 19.1 & \dab{0.7} 60.6 & \ua{0.4} 18.0 & 61.3 & \da{1.9} 15.7 & \uag{1.3} 62.6 & \ua{3.4} 21.0 & \uag{1.3} 62.6 & \da{3.8} 13.8 \\

\midrule
\multicolumn{11}{l}{\textbf{Country}} \\

EN & 82.3 & 6.7 & \dab{0.1} 82.2 & \da{0.8} 5.9 & 82.3 & 6.7 & 82.3 & 6.7 & \dab{0.1} 82.2 & \da{5.4} 1.3 & 82.3 & \da{6.6} 0.1 \\
ES & 65.1 & 5.1 & 65.1 & \da{0.2} 4.9 & \dab{0.1} 65.0 & \ua{0.3} 5.4 & \dab{0.4} 64.7 & \ua{1.8} 6.9 & \dab{0.6} 64.5 & \da{0.4} 4.7 & \dab{0.4} 64.7 & \da{3.0} 2.1 \\
IT & 71.0 & 1.7 & \uag{0.1} 71.1 & \da{0.4} 1.3 & \dab{0.1} 70.9 & \ua{0.4} 2.1 & \dab{0.2} 70.8 & \ua{0.4} 2.1 & 71.0 & \ua{1.6} 3.3 & \dab{0.5} 70.5 & \da{1.1} 0.6 \\
PL & -    & -    & -    & -    & -    & -    & -    & -    & -    & -  & - & - \\
PT & 64.5 & 5.1 & \dab{0.5} 64.0 & \da{4.3} 0.8 & \uag{0.4} 64.9 & \ua{0.4} 5.5 & \uag{2.0} 66.5 & \ua{4.4} 9.5 & \uag{2.0} 66.5 & \da{4.9} 0.2 & \dab{1.0} 63.5 & \da{4.2} 0.9 \\

\bottomrule

\end{tabular}
}
\caption{Demographic bias mitigation results on the Multilingual Hate Speech dataset using mBERT, comparing monolingual and multilingual debiasing approaches. Main: Hate speech prediction accuracy; TPR-Gap: True positive rate gap between demographic groups. We exclude results for Italian in race bias evaluation and Poland in country bias evaluation due to severely imbalanced class distributions in these subsets that could lead to unreliable bias measurements. Detailed dataset statistics can be found in Table~\ref{appendix:hate-speech-size} and Table~\ref{appendix:hate-speech-data-distribution} (appendix).}
\label{tab:hate-speech}
\end{table*}

\paragraph{With Target Language} Consider Table~\ref{tab:hate-speech}. IMSAE (``Five-Subsets'') demonstrates stronger bias mitigation compared to monolingual debiasing for gender and age debiasing. For race bias, we achieve reductions of 4.1\% for English and 6.2\% for Polish, compared to monolingual reductions of 1.3\% and 1.2\% respectively. Similarly significant improvements are seen for gender bias (11.6\% reduction by IMSAE on Polish vs 8.8\% monolingual) and age bias for all languages. Importantly, these improvements come with minimal impact on main task performance, with accuracy changes generally below or equal to 2\%.

\paragraph{Without Target Language}
Even when target language data is unavailable, IMSAE (``Subsets w/o'') effectively reduces bias across different attributes. Using only non-target languages, we achieve maximum bias reductions of 9.2\% for race, 5.2\% for gender, 9.0\% for age, 5.4\% for country. Gender debiasing proves particularly challenging in cross-lingual scenarios due to fundamental differences in gender representation across languages, especially when grammatical gender in some languages creates structural barriers to transfer. The main task performance remains stable, demonstrating IMSAE's ability to preserve useful features while removing bias. See also results in Appendix~\ref{appendix:hate-speech}.

\subsection{Multilingual Stack Exchange Fairness Benchmark}
\label{exp:sefair}
We use the MSEFair dataset, for which a detailed description is provided in \S\ref{sec:MSEFair}.

\subsubsection{Results} 

\begin{table*}[th!]
\centering
\resizebox{\textwidth}{!}{
\begin{tabular}{lrr  rr  rr  rr  rr  rr}
\toprule
Target & \multicolumn{2}{c}{Baseline} & \multicolumn{2}{c}{Monolingual} & \multicolumn{2}{c}{SAL (Avg-Crosslingual)} & \multicolumn{2}{c}{\subsetswithout{IMSAE (FullyJoint)}} & \multicolumn{2}{c}{\subsetswithout{IMSAE (Subsets w/o)}} & \multicolumn{2}{c}{\subsetswith{IMSAE (Three-Subsets)}}  \\

& Main & TPR-Gap & Main & TPR-Gap & Main & TPR-Gap & Main & TPR-Gap & Main & TPR-Gap & Main & TPR-Gap\\

\midrule 
\multicolumn{13}{l}{Multilingual BERT} \\
EN & 67.5 & 10.7 & \dab{4.3} 63.2 & \da{9.4} 1.3 & \dab{0.2} 67.3 & \ua{0.3} 11.0 & \dab{0.3} 67.2 & 10.7 & \dab{4.2} 63.3 & \da{9.3} 1.4 & \dab{4.4} 63.1 & \da{9.5} 1.2 \\
PT & 78.3 & 16.2 & \dab{19.6} 58.7 & \da{14.8} 1.4 & 78.3 & 16.2 & \dab{0.1} 78.2 & \da{0.8} 15.4 & \dab{19.7} 58.6 & \da{14.5} 1.7 & \dab{19.8} 58.5 & \da{15.6} 0.6 \\
RU & 70.0 & 18.0 & \dab{11.9} 58.1 & \da{15.1} 2.9 & \dab{0.3} 69.7 & \ua{0.1} 18.1 & \dab{0.2} 69.8 & \da{0.3} 17.7 & \dab{0.6} 69.4 & \da{0.1} 17.9 & \dab{12.3} 57.7 & \da{14.7} 3.3 \\
\midrule 
\multicolumn{13}{l}{Llama-3.1-8B} \\
EN & 68.4 & 11.2 & \dab{4.3} 64.1 & \da{6.0} 5.2 & 68.4 & 11.2 & \dab{0.1} 68.3 & \ua{0.2} 11.4 & \dab{4.3} 64.1 & \da{6.1} 5.1 & \dab{4.5} 63.9 & \da{5.9} 5.3 \\
PT & 82.6 & 22.4 & \dab{18.2} 64.4 & \da{13.6} 8.8 & \dab{0.2} 82.4 & \da{0.1} 22.3 & \dab{0.7} 81.9 & \da{0.6} 21.8 & \dab{18.1} 64.5 & \da{13.1} 9.3 & \dab{18.3} 64.3 & \da{13.6} 8.8 \\
RU & 72.1 & 16.4 & \dab{8.4} 63.7 & \da{10.4} 6.0 & \dab{0.1} 72.0 & \da{0.1} 16.3 & 72.1 & \ua{0.2} 16.6 & \dab{0.1} 72.0 & \da{0.3} 16.1 & \dab{8.7} 63.4 & \da{10.6} 5.8 \\
\midrule 
\multicolumn{13}{l}{Mistral-7B-Instruct-v0.3} \\
EN & 66.8 & 9.5 & \dab{3.7} 63.1 & \da{5.8} 3.7 & 66.8 & \da{0.1} 9.3 & \dab{0.1} 66.7 & \da{0.5} 9.0 & \dab{3.6} 63.2 & \da{6.3} 3.2 & \dab{3.8} 63.0 & \da{6.4} 3.1 \\
PT & 81.4 & 20.9 & \dab{18.2} 63.2 & \da{9.5} 11.4 & \dab{0.2} 81.2 & \ua{0.4} 21.3 & \dab{0.1} 81.3 & \da{0.5} 20.4 & \dab{18.1} 63.3 & \da{9.7} 11.2 & \dab{17.9} 63.5 & \da{8.2} 12.7 \\
RU & 70.5 & 14.7 & \dab{8.2} 62.3 & \da{9.3} 5.4 & \dab{0.1} 70.4 & \da{0.3} 14.4 & 70.5 & \da{0.3} 14.4 & \dab{0.1} 70.4 & \da{0.9} 13.8 & \dab{8.4} 62.1 & \da{9.4} 5.3 \\
\midrule

\end{tabular}

}

\caption{Reputation bias mitigation on the MSEFair dataset across English, Portuguese and Russian, comparing monolingual and multilingual debiasing approaches. Main: helpfulness prediction accuracy; TPR-Gap: True positive rate gap between demographic groups.}

\label{tab:SEFair}

\end{table*}

\paragraph{With Target Language} 
Consider Table~\ref{tab:SEFair}. Our method IMSAE, which uses all available languages (``Three-Subsets``), demonstrates superior bias mitigation on average across all language models compared to monolingual debiasing using SAL. However, both methods cause significant damage to model utility for Portuguese and Russian. We want to bring to the community's attention that Stack Exchange helpfulness is a challenging topic to work on, as small changes in embeddings can lead to huge drops in classification performance.

\paragraph{Without Target Language}

The performance of IMSAE's (``Subsets w/o'') varies with linguistic similarity, but outperforms SAL and FullyJoint (\S\ref{section:fullyjoint}). However, for Russian, cross-lingual debiasing shows limited effectiveness, with small TPR-Gap reduction. This limitation may stem from Russian's linguistic features affecting bias transfer (Cyrillic script, complex case system, different word order). Newer architectures (Llama and Mistral) show improved cross-lingual debiasing performance compared to mBERT, suggesting better cross-lingual representation alignment in these models.

\section{Related Work}

We focus in our related work on debiasing in the context of multilingual representations.
Debiasing multilingual representations is harder due to language-specific traits \cite{ramesh-etal-2023-fairness}, mismatches like grammatical vs. biological gender \cite{booij2010construction, veeman-etal-2020-cross}, biases introduced during language alignment \cite{zhao-etal-2020-gender}, and the partial overlap of gender components across languages \cite{gonen-etal-2022-analyzing}. Past work on multilingual debiasing has largely focused on cross-lingual transfer—applying debiasing from one language to another. \citet{zhou-etal-2019-examining} laid the groundwork by separating grammatical and semantic gender bias, enabling targeted debiasing via semantic shifts or alignment with English embeddings. \citet{zhao-etal-2020-gender} refined this by equalizing distances between target words and protected sets. \citet{liang-etal-2020-monolingual} proposed maximizing intra-group and minimizing inter-group distances across genders. \citet{reusens2023investigating} found SentenceDebias most effective on mBERT for cross-lingual debiasing. However, \citet{gonen-etal-2022-analyzing} showed that relying on a single source language misses language-specific bias, limiting effectiveness. Multilingual debiasing requires a comprehensive approach. \citet{vashishtha-etal-2023-evaluating} found limited debiasing transfer, particularly from English to languages without a Western context. Recent activation editing methods including Sparse Activation Editing \cite{zhao2025sae} and SEA \cite{qiu2024spectral} address alignment and bias at inference time, while IMSAE specifically targets joint demographic bias subspaces across multiple languages through iterative erasure; related inference-time representation editing has also been explored in text-to-image diffusion via prompt-embedding projection \cite{fu2025fairimagen}.

\section{Conclusion}

We have studied debiasing multilingual representations by identifying joint linear bias subspaces across languages. Our proposed method, IMSAE, iteratively identifies and removes bias patterns using different language subsets. The method leverages bias patterns from multiple languages, enabling effective zero-shot debiasing where target language data is unavailable but linguistically similar languages can be used.  Through  experiments across eight languages and five demographic attributes, we demonstrate IMSAE's effectiveness in reducing bias while preserving model utility in state-of-the-art language models. In addition, we have introduced the MSEFair dataset to support future multilingual fairness research further.

\ignore{
We assume a latent variable $\rv{H}$ that when provided,
$\mathbf{X}_{\ell}$ and $\mathbf{Z}$ are conditionally independent \cite{shao2023erasure}. Then, there are mixture coefficients $\xi_h$ and:

\begin{equation}
\mathbf{E}[\mathbf{X}_{\ell} \mathbf{Z}^{\top}] = \sum_h \xi_h \mathbf{E}[\mathbf{X}_{\ell} \mid h] \cdot \mathbf{E}[ \mathbf{Z}^{\top} \mid h]. \label{eq1}
\end{equation}

In addition, we assume that there are $C$ meta-linguistic random variables,
$\overline{\mathbf{X}}_1, \ldots, \overline{\mathbf{X}}_C$ such that for each language, the probability over $\mathbf{X}_{\ell}$ is uniquely determined by setting mixture coefficients $\zeta_{h,\ell,1}, \ldots, \zeta_{h,\ell,C}$ such that:

\begin{equation}
p(\mathbf{X}_{\ell} = x \mid h) = \sum_{i} \zeta_{h,\ell,i} p(\overline{\mathbf{X}}_i = x \mid h).   \label{eq2}
\end{equation}

This implies that each language vector representation distribution is determined by these meta-linguistic vectors as a combination of them whenever $\mathbf{H}$ is provided. This is how multiple languages are represented in the joint space.

The implication of Eq.~\refeq{eq1}--\refeq{eq2} is that given a specific distribution over the languages in $\mathcal{L}$, the covariance $\mathbf{E}[\rv{X}\rv{Z}^{\top}]$ between a vector $\rv{X}$ from that distribution and $\mathbf{Z}$ can be rewritten as:

\begin{align}
\sum_{\ell} \sum_h \sum_{i=1}^C \mu_{\ell} \xi_h \zeta_{h,\ell,i} \mathbf{E}[\overline{\mathbf{X}}_{i} \mid h] \cdot \mathbf{E}[ \mathbf{Z}^{\top} \mid h].
\end{align}
}

\section*{Limitations}
While IMSAE provides a promising approach to multilingual debiasing, it relies on the assumption of a shared embedding space \cite{wendler2024llama, fierro2025facts}, which may be an issue with highly divergent languages.

In addition, our evaluations mostly focus on European languages and a narrow range of demographic attributes, leaving open questions about IMSAE’s ability to generalize to typologically diverse languages.
This may compromise its performance for such languages and, ultimately, language communities with limited resources. 
Beyond demographic bias, fairness considerations span diverse AI applications including embodied systems \cite{li2024voice}.

Furthermore, our analysis focuses on the final representations of the LLMs. While prior work shows that information is distributed across layers \citep{zhao-etal-2024-layer}, we leave a layer-wise analysis to future work.


\section*{Ethical Considerations}
Our reliance on datasets that use binary demographic categories, such as male/female or white/non-white, risks reinforcing reductive stereotypes and marginalizing non-binary and intersectional identities. For a discussion,
see \newcite{dev-etal-2021-harms} and \newcite{cao-daume-iii-2020-toward}.
As we consider deploying IMSAE in real-world applications, it is essential to recognize that bias in language models is complex and context-dependent. Like any debiasing method, IMSAE may have unintended consequences, particularly across different languages and cultural settings. It should not be treated as a superficial fix to mask deeper systemic issues in AI systems. Instead, responsible deployment demands ongoing validation, transparency, and meaningful engagement with the communities affected by these technologies.

\section*{Acknowledgments}
We thank the reviewers for their helpful comments and feedback. SS and AK acknowledge the support of the UKRI Frontier grant
EP/Y031350/1. We gratefully acknowledge an Apple AI/ML scholarship awarded to YQ. ZZ is supported by the UKRI Centre for Doctoral Training in Natural Language Processing (EP/S022481/1). We are grateful for UKRI's support (SC) through the provision of compute resources at the University of Birmingham (Baskerville). We thank the Stack Overflow team
for making their data easily available to download.

\bibliography{anthology,custom, custom2}

@inproceedings{veeman-etal-2020-cross,
    title = "Cross-lingual Embeddings Reveal Universal and Lineage-Specific Patterns in Grammatical Gender Assignment",
    author = "Veeman, Hartger  and
      Allassonni{\`e}re-Tang, Marc  and
      Berdicevskis, Aleksandrs  and
      Basirat, Ali",
    booktitle = "Proceedings of the 24th Conference on Computational Natural Language Learning",
    month = nov,
    year = "2020",
    address = "Online",
    publisher = "Association for Computational Linguistics",
    url = "https://aclanthology.org/2020.conll-1.20",
    doi = "10.18653/v1/2020.conll-1.20",
    pages = "265--275",
}

@inproceedings{conneau-etal-2020-unsupervised,
    title = "Unsupervised Cross-lingual Representation Learning at Scale",
    author = "Conneau, Alexis  and
      Khandelwal, Kartikay  and
      Goyal, Naman  and
      Chaudhary, Vishrav  and
      Wenzek, Guillaume  and
      Guzm{\'a}n, Francisco  and
      Grave, Edouard  and
      Ott, Myle  and
      Zettlemoyer, Luke  and
      Stoyanov, Veselin",
    booktitle = "Proceedings of the 58th Annual Meeting of the Association for Computational Linguistics",
    month = jul,
    year = "2020",
    address = "Online",
    publisher = "Association for Computational Linguistics",
    url = "https://aclanthology.org/2020.acl-main.747",
    doi = "10.18653/v1/2020.acl-main.747",
    pages = "8440--8451",
}

@inproceedings{huang-etal-2020-multilingual,
    title = "Multilingual {T}witter Corpus and Baselines for Evaluating Demographic Bias in Hate Speech Recognition",
    author = "Huang, Xiaolei  and
      Xing, Linzi  and
      Dernoncourt, Franck  and
      Paul, Michael J.",
    booktitle = "Proceedings of the Twelfth Language Resources and Evaluation Conference",
    month = may,
    year = "2020",
    address = "Marseille, France",
    publisher = "European Language Resources Association",
    url = "https://aclanthology.org/2020.lrec-1.180",
    pages = "1440--1448",
    language = "English",
    ISBN = "979-10-95546-34-4",
}

@article{booij2010construction,
  title={Construction morphology},
  author={Booij, Geert},
  journal={Language and linguistics compass},
  volume={4},
  number={7},
  pages={543--555},
  year={2010},
  publisher={Wiley Online Library}
}

@article{hovy-etal-2021,
author = {Hovy, Dirk and Prabhumoye, Shrimai},
title = {Five sources of bias in natural language processing},
journal = {Language and Linguistics Compass},
volume = {15},
number = {8},
pages = {e12432},
doi = {https://doi.org/10.1111/lnc3.12432},
url = {https://compass.onlinelibrary.wiley.com/doi/abs/10.1111/lnc3.12432},
eprint = {https://compass.onlinelibrary.wiley.com/doi/pdf/10.1111/lnc3.12432},
year = {2021}
}

@inproceedings{talat-etal-2022-reap,
    title = "You reap what you sow: On the Challenges of Bias Evaluation Under Multilingual Settings",
    author = "Talat, Zeerak  and
      N{\'e}v{\'e}ol, Aur{\'e}lie  and
      Biderman, Stella  and
      Clinciu, Miruna  and
      Dey, Manan  and
      Longpre, Shayne  and
      Luccioni, Sasha  and
      Masoud, Maraim  and
      Mitchell, Margaret  and
      Radev, Dragomir  and
      Sharma, Shanya  and
      Subramonian, Arjun  and
      Tae, Jaesung  and
      Tan, Samson  and
      Tunuguntla, Deepak  and
      Van Der Wal, Oskar",
    editor = "Fan, Angela  and
      Ilic, Suzana  and
      Wolf, Thomas  and
      Gall{\'e}, Matthias",
    booktitle = "Proceedings of BigScience Episode {\#}5 -- Workshop on Challenges {\&} Perspectives in Creating Large Language Models",
    month = may,
    year = "2022",
    address = "virtual+Dublin",
    publisher = "Association for Computational Linguistics",
    url = "https://aclanthology.org/2022.bigscience-1.3",
    doi = "10.18653/v1/2022.bigscience-1.3",
    pages = "26--41",
}

@article{chu2024fairness,
  title={Fairness in Large Language Models: A Taxonomic Survey},
  author={Chu, Zhibo and Wang, Zichong and Zhang, Wenbin},
  journal={arXiv preprint arXiv:2404.01349},
  year={2024}
}

@article{osborne-etal-2016-encoding,
    title = "Encoding Prior Knowledge with Eigenword Embeddings",
    author = "Osborne, Dominique  and
      Narayan, Shashi  and
      Cohen, Shay B.",
    editor = "Lee, Lillian  and
      Johnson, Mark  and
      Toutanova, Kristina",
    journal = "Transactions of the Association for Computational Linguistics",
    volume = "4",
    year = "2016",
    address = "Cambridge, MA",
    publisher = "MIT Press",
    url = "https://aclanthology.org/Q16-1030/",
    doi = "10.1162/tacl_a_00108",
    pages = "417--430"
}

@inproceedings{zhao-etal-2023-joint,
    title = "A Joint Matrix Factorization Analysis of Multilingual Representations",
    author = "Zhao, Zheng  and
      Ziser, Yftah  and
      Webber, Bonnie  and
      Cohen, Shay",
    editor = "Bouamor, Houda  and
      Pino, Juan  and
      Bali, Kalika",
    booktitle = "Findings of the Association for Computational Linguistics: EMNLP 2023",
    month = dec,
    year = "2023",
    address = "Singapore",
    publisher = "Association for Computational Linguistics",
    url = "https://aclanthology.org/2023.findings-emnlp.851/",
    doi = "10.18653/v1/2023.findings-emnlp.851",
    pages = "12764--12783"
}

@inproceedings{cao-daume-iii-2020-toward,
    title = "Toward Gender-Inclusive Coreference Resolution",
    author = "Cao, Yang Trista  and
      Daum{\'e} III, Hal",
    editor = "Jurafsky, Dan  and
      Chai, Joyce  and
      Schluter, Natalie  and
      Tetreault, Joel",
    booktitle = "Proceedings of the 58th Annual Meeting of the Association for Computational Linguistics",
    month = jul,
    year = "2020",
    address = "Online",
    publisher = "Association for Computational Linguistics",
    url = "https://aclanthology.org/2020.acl-main.418/",
    doi = "10.18653/v1/2020.acl-main.418",
    pages = "4568--4595"
}

@article{ben2015projectors,
  title={Projectors on intersection of subspaces},
  author={Ben-Israel, Adi},
  journal={Contemporary Mathematics},
  volume={636},
  pages={41--50},
  year={2015},
  publisher={American Mathematical Society}
}

@inproceedings{fu2025fairimagen,
  title     = {FairImagen: Post-Processing for Bias Mitigation in Text-to-Image Models},
  author    = {Fu, Zihao and Brown, Ryan and Shao, Shun and Rawal, Kai and Delaney, Eoin D. and Russell, Chris},
  booktitle = {Proceedings of the 39th Conference on Neural Information Processing Systems (NeurIPS)},
  year      = {2025}
}

@inproceedings{de2019bias,
  title={Bias in bios: A case study of semantic representation bias in a high-stakes setting},
  author={De-Arteaga, Maria and Romanov, Alexey and Wallach, Hanna and Chayes, Jennifer and Borgs, Christian and Chouldechova, Alexandra and Geyik, Sahin and Kenthapadi, Krishnaram and Kalai, Adam Tauman},
  booktitle={proceedings of the Conference on Fairness, Accountability, and Transparency},
  pages={120--128},
  year={2019}
}

@misc{li2024voice,
      title={Beyond Voice Assistants: Exploring Advantages and Risks of an In-Car Social Robot in Real Driving Scenarios}, 
      author={Yuanchao Li and Lachlan Urquhart and Nihan Karatas and Shun Shao and Hiroshi Ishiguro and Xun Shen},
      year={2024},
      eprint={2402.11853},
      archivePrefix={arXiv},
      primaryClass={cs.HC},
      url={https://arxiv.org/abs/2402.11853}, 
}

@inproceedings{liang-etal-2020-monolingual,
    title = "Monolingual and Multilingual Reduction of Gender Bias in Contextualized Representations",
    author = {Liang, Sheng  and
      Dufter, Philipp  and
      Sch{\"u}tze, Hinrich},
    editor = "Scott, Donia  and
      Bel, Nuria  and
      Zong, Chengqing",
    booktitle = "Proceedings of the 28th International Conference on Computational Linguistics",
    month = dec,
    year = "2020",
    address = "Barcelona, Spain (Online)",
    publisher = "International Committee on Computational Linguistics",
    url = "https://aclanthology.org/2020.coling-main.446",
    doi = "10.18653/v1/2020.coling-main.446",
    pages = "5082--5093",
}

@inproceedings{shao-etal-2022-SAL,
    title = "Gold Doesn{'}t Always Glitter: Spectral Removal of Linear and Nonlinear Guarded Attribute Information",
    author = "Shao, Shun  and
      Ziser, Yftah  and
      Cohen, Shay B.",
    editor = "Vlachos, Andreas  and
      Augenstein, Isabelle",
    booktitle = "Proceedings of the 17th Conference of the European Chapter of the Association for Computational Linguistics",
    month = may,
    year = "2023",
    address = "Dubrovnik, Croatia",
    publisher = "Association for Computational Linguistics",
    url = "https://aclanthology.org/2023.eacl-main.118",
    doi = "10.18653/v1/2023.eacl-main.118",
    pages = "1611--1622",
}

@inproceedings{ravfogel-etal-2020-null,
    title = "Null It Out: Guarding Protected Attributes by Iterative Nullspace Projection",
    author = "Ravfogel, Shauli  and
      Elazar, Yanai  and
      Gonen, Hila  and
      Twiton, Michael  and
      Goldberg, Yoav",
    booktitle = "Proceedings of the 58th Annual Meeting of the Association for Computational Linguistics",
    month = jul,
    year = "2020",
    address = "Online",
    publisher = "Association for Computational Linguistics",
    url = "https://aclanthology.org/2020.acl-main.647",
    doi = "10.18653/v1/2020.acl-main.647",
    pages = "7237--7256",
}

@inproceedings{ramesh-etal-2023-fairness,
    title = "Fairness in Language Models Beyond {E}nglish: Gaps and Challenges",
    author = "Ramesh, Krithika  and
      Sitaram, Sunayana  and
      Choudhury, Monojit",
    editor = "Vlachos, Andreas  and
      Augenstein, Isabelle",
    booktitle = "Findings of the Association for Computational Linguistics: EACL 2023",
    month = may,
    year = "2023",
    address = "Dubrovnik, Croatia",
    publisher = "Association for Computational Linguistics",
    url = "https://aclanthology.org/2023.findings-eacl.157",
    doi = "10.18653/v1/2023.findings-eacl.157",
    pages = "2106--2119",
}

@inproceedings{dev-etal-2021-harms,
    title = "Harms of Gender Exclusivity and Challenges in Non-Binary Representation in Language Technologies",
    author = "Dev, Sunipa  and
      Monajatipoor, Masoud  and
      Ovalle, Anaelia  and
      Subramonian, Arjun  and
      Phillips, Jeff  and
      Chang, Kai-Wei",
    editor = "Moens, Marie-Francine  and
      Huang, Xuanjing  and
      Specia, Lucia  and
      Yih, Scott Wen-tau",
    booktitle = "Proceedings of the 2021 Conference on Empirical Methods in Natural Language Processing",
    month = nov,
    year = "2021",
    address = "Online and Punta Cana, Dominican Republic",
    publisher = "Association for Computational Linguistics",
    url = "https://aclanthology.org/2021.emnlp-main.150",
    doi = "10.18653/v1/2021.emnlp-main.150",
    pages = "1968--1994",
}

@inproceedings{chen-etal-2024-cher,
    title = "{C}her at {KSAA}-{CAD} 2024: Compressing Words and Definitions into the Same Space for {A}rabic Reverse Dictionary",
    author = "Chen, Pinzhen  and
      Zhao, Zheng  and
      Shao, Shun",
    editor = "Habash, Nizar  and
      Bouamor, Houda  and
      Eskander, Ramy  and
      Tomeh, Nadi  and
      Abu Farha, Ibrahim  and
      Abdelali, Ahmed  and
      Touileb, Samia  and
      Hamed, Injy  and
      Onaizan, Yaser  and
      Alhafni, Bashar  and
      Antoun, Wissam  and
      Khalifa, Salam  and
      Haddad, Hatem  and
      Zitouni, Imed  and
      AlKhamissi, Badr  and
      Almatham, Rawan  and
      Mrini, Khalil",
    booktitle = "Proceedings of the Second Arabic Natural Language Processing Conference",
    month = aug,
    year = "2024",
    address = "Bangkok, Thailand",
    publisher = "Association for Computational Linguistics",
    url = "https://aclanthology.org/2024.arabicnlp-1.75/",
    doi = "10.18653/v1/2024.arabicnlp-1.75",
    pages = "686--691"
}

@inproceedings{zhao-etal-2020-gender,
    title = "Gender Bias in Multilingual Embeddings and Cross-Lingual Transfer",
    author = "Zhao, Jieyu  and
      Mukherjee, Subhabrata  and
      Hosseini, Saghar  and
      Chang, Kai-Wei  and
      Hassan Awadallah, Ahmed",
    editor = "Jurafsky, Dan  and
      Chai, Joyce  and
      Schluter, Natalie  and
      Tetreault, Joel",
    booktitle = "Proceedings of the 58th Annual Meeting of the Association for Computational Linguistics",
    month = jul,
    year = "2020",
    address = "Online",
    publisher = "Association for Computational Linguistics",
    url = "https://aclanthology.org/2020.acl-main.260",
    doi = "10.18653/v1/2020.acl-main.260",
    pages = "2896--2907",
}

@inproceedings{gonen-etal-2022-analyzing,
    title = "Analyzing Gender Representation in Multilingual Models",
    author = "Gonen, Hila  and
      Ravfogel, Shauli  and
      Goldberg, Yoav",
    booktitle = "Proceedings of the 7th Workshop on Representation Learning for NLP",
    month = may,
    year = "2022",
    address = "Dublin, Ireland",
    publisher = "Association for Computational Linguistics",
    url = "https://aclanthology.org/2022.repl4nlp-1.8",
    doi = "10.18653/v1/2022.repl4nlp-1.8",
    pages = "67--77",
}

@inproceedings{reusens2023investigating,
    title = "Investigating Bias in Multilingual Language Models: Cross-Lingual Transfer of Debiasing Techniques",
    author = "Reusens, Manon  and
      Borchert, Philipp  and
      Mieskes, Margot  and
      De Weerdt, Jochen  and
      Baesens, Bart",
    editor = "Bouamor, Houda  and
      Pino, Juan  and
      Bali, Kalika",
    booktitle = "Proceedings of the 2023 Conference on Empirical Methods in Natural Language Processing",
    month = dec,
    year = "2023",
    address = "Singapore",
    publisher = "Association for Computational Linguistics",
    url = "https://aclanthology.org/2023.emnlp-main.175",
    doi = "10.18653/v1/2023.emnlp-main.175",
    pages = "2887--2896",
}

@inproceedings{vashishtha-etal-2023-evaluating,
    title = "On Evaluating and Mitigating Gender Biases in Multilingual Settings",
    author = "Vashishtha, Aniket  and
      Ahuja, Kabir  and
      Sitaram, Sunayana",
    editor = "Rogers, Anna  and
      Boyd-Graber, Jordan  and
      Okazaki, Naoaki",
    booktitle = "Findings of the Association for Computational Linguistics: ACL 2023",
    month = jul,
    year = "2023",
    address = "Toronto, Canada",
    publisher = "Association for Computational Linguistics",
    url = "https://aclanthology.org/2023.findings-acl.21",
    doi = "10.18653/v1/2023.findings-acl.21",
    pages = "307--318",
}

@misc{wendler2024llama,
      title={Do Llamas Work in English? On the Latent Language of Multilingual Transformers}, 
      author={Chris Wendler and Veniamin Veselovsky and Giovanni Monea and Robert West},
      year={2024},
      eprint={2402.10588},
      archivePrefix={arXiv},
      primaryClass={cs.CL},
      url={https://arxiv.org/abs/2402.10588}, 
}

@misc{fierro2025facts,
      title={How Do Multilingual Language Models Remember Facts?}, 
      author={Constanza Fierro and Negar Foroutan and Desmond Elliott and Anders Søgaard},
      year={2025},
      eprint={2410.14387},
      archivePrefix={arXiv},
      primaryClass={cs.CL},
      url={https://arxiv.org/abs/2410.14387}, 
}

@inproceedings{liang-etal-2020-towards,
    title = "Towards Debiasing Sentence Representations",
    author = "Liang, Paul Pu  and
      Li, Irene Mengze  and
      Zheng, Emily  and
      Lim, Yao Chong  and
      Salakhutdinov, Ruslan  and
      Morency, Louis-Philippe",
    editor = "Jurafsky, Dan  and
      Chai, Joyce  and
      Schluter, Natalie  and
      Tetreault, Joel",
    booktitle = "Proceedings of the 58th Annual Meeting of the Association for Computational Linguistics",
    month = jul,
    year = "2020",
    address = "Online",
    publisher = "Association for Computational Linguistics",
    url = "https://aclanthology.org/2020.acl-main.488",
    doi = "10.18653/v1/2020.acl-main.488",
    pages = "5502--5515",
}

@inproceedings{neveol-etal-2022-french,
    title = "{F}rench {C}row{S}-Pairs: Extending a challenge dataset for measuring social bias in masked language models to a language other than {E}nglish",
    author = {N{\'e}v{\'e}ol, Aur{\'e}lie  and
      Dupont, Yoann  and
      Bezan{\c{c}}on, Julien  and
      Fort, Kar{\"e}n},
    editor = "Muresan, Smaranda  and
      Nakov, Preslav  and
      Villavicencio, Aline",
    booktitle = "Proceedings of the 60th Annual Meeting of the Association for Computational Linguistics (Volume 1: Long Papers)",
    month = may,
    year = "2022",
    address = "Dublin, Ireland",
    publisher = "Association for Computational Linguistics",
    url = "https://aclanthology.org/2022.acl-long.583",
    doi = "10.18653/v1/2022.acl-long.583",
    pages = "8521--8531",
}

@inproceedings{orgad-belinkov-2023-blind,
    title = "{BLIND}: Bias Removal With No Demographics",
    author = "Orgad, Hadas  and
      Belinkov, Yonatan",
    editor = "Rogers, Anna  and
      Boyd-Graber, Jordan  and
      Okazaki, Naoaki",
    booktitle = "Proceedings of the 61st Annual Meeting of the Association for Computational Linguistics (Volume 1: Long Papers)",
    month = jul,
    year = "2023",
    address = "Toronto, Canada",
    publisher = "Association for Computational Linguistics",
    url = "https://aclanthology.org/2023.acl-long.490",
    doi = "10.18653/v1/2023.acl-long.490",
    pages = "8801--8821",
}

@article{shao2023erasure,
    author = {Shao, Shun and Ziser, Yftah and Cohen, Shay B.},
    title = "{Erasure of Unaligned Attributes from Neural Representations}",
    journal = {Transactions of the Association for Computational Linguistics},
    volume = {11},
    pages = {488-510},
    year = {2023},
    month = {05},
    issn = {2307-387X},
    doi = {10.1162/tacl_a_00558},
    url = {https://doi.org/10.1162/tacl\_a\_00558},
    eprint = {https://direct.mit.edu/tacl/article-pdf/doi/10.1162/tacl\_a\_00558/2110602/tacl\_a\_00558.pdf},
}

@misc{webster2021measuring,
      title={Measuring and Reducing Gendered Correlations in Pre-trained Models}, 
      author={Kellie Webster and Xuezhi Wang and Ian Tenney and Alex Beutel and Emily Pitler and Ellie Pavlick and Jilin Chen and Ed Chi and Slav Petrov},
      year={2021},
      eprint={2010.06032},
      archivePrefix={arXiv},
      primaryClass={cs.CL}
}

@article{qiu2024spectral,
  title={Spectral editing of activations for large language model alignment},
  author={Qiu, Yifu and Zhao, Zheng and Ziser, Yftah and Korhonen, Anna and Ponti, Edoardo Maria and Cohen, Shay},
  journal={Advances in Neural Information Processing Systems},
  volume={37},
  pages={56958--56987},
  year={2024}
}

@inproceedings{zhou-etal-2019-density,
    title = "Density Matching for Bilingual Word Embedding",
    author = "Zhou, Chunting  and
      Ma, Xuezhe  and
      Wang, Di  and
      Neubig, Graham",
    editor = "Burstein, Jill  and
      Doran, Christy  and
      Solorio, Thamar",
    booktitle = "Proceedings of the 2019 Conference of the North {A}merican Chapter of the Association for Computational Linguistics: Human Language Technologies, Volume 1 (Long and Short Papers)",
    month = jun,
    year = "2019",
    address = "Minneapolis, Minnesota",
    publisher = "Association for Computational Linguistics",
    url = "https://aclanthology.org/N19-1161/",
    doi = "10.18653/v1/N19-1161",
    pages = "1588--1598"
}

@inproceedings{zhou-etal-2019-examining,
    title = "Examining Gender Bias in Languages with Grammatical Gender",
    author = "Zhou, Pei  and
      Shi, Weijia  and
      Zhao, Jieyu  and
      Huang, Kuan-Hao  and
      Chen, Muhao  and
      Cotterell, Ryan  and
      Chang, Kai-Wei",
    editor = "Inui, Kentaro  and
      Jiang, Jing  and
      Ng, Vincent  and
      Wan, Xiaojun",
    booktitle = "Proceedings of the 2019 Conference on Empirical Methods in Natural Language Processing and the 9th International Joint Conference on Natural Language Processing (EMNLP-IJCNLP)",
    month = nov,
    year = "2019",
    address = "Hong Kong, China",
    publisher = "Association for Computational Linguistics",
    url = "https://aclanthology.org/D19-1531",
    doi = "10.18653/v1/D19-1531",
    pages = "5276--5284",
}

@inproceedings{lauscher-etal-2021-sustainable-modular,
    title = "Sustainable Modular Debiasing of Language Models",
    author = "Lauscher, Anne  and
      Lueken, Tobias  and
      Glava{\v{s}}, Goran",
    editor = "Moens, Marie-Francine  and
      Huang, Xuanjing  and
      Specia, Lucia  and
      Yih, Scott Wen-tau",
    booktitle = "Findings of the Association for Computational Linguistics: EMNLP 2021",
    month = nov,
    year = "2021",
    address = "Punta Cana, Dominican Republic",
    publisher = "Association for Computational Linguistics",
    url = "https://aclanthology.org/2021.findings-emnlp.411",
    doi = "10.18653/v1/2021.findings-emnlp.411",
    pages = "4782--4797",
}

@inproceedings{devlin-etal-2019-bert,
    title = "{BERT}: Pre-training of Deep Bidirectional Transformers for Language Understanding",
    author = "Devlin, Jacob  and
      Chang, Ming-Wei  and
      Lee, Kenton  and
      Toutanova, Kristina",
    editor = "Burstein, Jill  and
      Doran, Christy  and
      Solorio, Thamar",
    booktitle = "Proceedings of the 2019 Conference of the North {A}merican Chapter of the Association for Computational Linguistics: Human Language Technologies, Volume 1 (Long and Short Papers)",
    month = jun,
    year = "2019",
    address = "Minneapolis, Minnesota",
    publisher = "Association for Computational Linguistics",
    url = "https://aclanthology.org/N19-1423",
    doi = "10.18653/v1/N19-1423",
    pages = "4171--4186",
}

@misc{grattafiori2024llama3herdmodels,
      title={The {Llama} 3 Herd of Models}, 
      author={Aaron Grattafiori and others},
      year={2024},
      eprint={2407.21783},
      archivePrefix={arXiv},
      primaryClass={cs.AI},
      url={https://arxiv.org/abs/2407.21783}, 
}

@article{meta2024llama31,
  title={Introducing {Llama} 3.1: Our most capable models to date},
  author={Meta, AI},
  journal={Meta AI Blog},
  volume={12},
  year={2024}
}

@article{meta2024llama32,
  title={Llama 3.2: Revolutionizing edge AI and vision with open, customizable models},
  author={Meta, AI},
  journal={Meta AI Blog. Retrieved December},
  volume={20},
  pages={2024},
  year={2024}
}

@misc{jiang2023mistral7b,
      title={Mistral 7B}, 
      author={Albert Q. Jiang and Alexandre Sablayrolles and Arthur Mensch and Chris Bamford and Devendra Singh Chaplot and Diego de las Casas and Florian Bressand and Gianna Lengyel and Guillaume Lample and Lucile Saulnier and Lélio Renard Lavaud and Marie-Anne Lachaux and Pierre Stock and Teven Le Scao and Thibaut Lavril and Thomas Wang and Timothée Lacroix and William El Sayed},
      year={2023},
      eprint={2310.06825},
      archivePrefix={arXiv},
      primaryClass={cs.CL},
      url={https://arxiv.org/abs/2310.06825}, 
}

@misc{mistral2024nemo,
  author = {{Mistral AI}},
  title = {Mistral {NeMo}},
  year = {2024},
  url = {https://mistral.ai/news/mistral-nemo/},
  note = {Accessed: January 14, 2025}
}

@misc{zhao2025sae,
      title={Sparse Activation Editing for Reliable Instruction Following in Narratives}, 
      author={Runcong Zhao and Chengyu Cao and Qinglin Zhu and Xiucheng Lv and Shun Shao and Lin Gui and Ruifeng Xu and Yulan He},
      year={2025},
      eprint={2505.16505},
      archivePrefix={arXiv},
      primaryClass={cs.CL},
      url={https://arxiv.org/abs/2505.16505}, 
}

@inproceedings{zhao-etal-2024-layer,
    title = "Layer by Layer: Uncovering Where Multi-Task Learning Happens in Instruction-Tuned Large Language Models",
    author = "Zhao, Zheng  and
      Ziser, Yftah  and
      Cohen, Shay B",
    editor = "Al-Onaizan, Yaser  and
      Bansal, Mohit  and
      Chen, Yun-Nung",
    booktitle = "Proceedings of the 2024 Conference on Empirical Methods in Natural Language Processing",
    month = nov,
    year = "2024",
    address = "Miami, Florida, USA",
    publisher = "Association for Computational Linguistics",
    url = "https://aclanthology.org/2024.emnlp-main.847/",
    doi = "10.18653/v1/2024.emnlp-main.847",
    pages = "15195--15214"
}
\bibliographystyle{acl_natbib}

\clearpage

\appendix

\section*{Appendices}

We include below further results that can complement and complete the results in the main part of the paper. The appendix can be skimmed on a first read, and is required only for a more in-depth analysis of IMSAE for those who are interested.

\section{Multilingual BiasBios Details}
\label{appendix:biasbios}
\label{app:multilingual-bb}

This appendix provides comprehensive information about the Multilingual BiasBios dataset used in our profession prediction experiments. Table \ref{appendix:dataset-stats-biasbios} includes detailed statistics on data distribution across languages, demographic attributes, and experimental configurations. Table \ref{appendix:biasbios-fullresults} shows the complete results for eight different LLMs debiased by SAL, IMSAE ("Three-Subsets"), and IMSAE ("Two-Subsets-Without"). Table \ref{appendix:biasbios-crosslingual} compares different post-hoc debiasing methods (SAL, INLP, and SentenceDebias) in crosslingual settings, showing their relative effectiveness when applied across language boundaries.

\label{appendix:biasbios-dataset}
\begin{table*}
\centering
\scalebox{1.0}{
\begin{tabular}{lrrrr}
\toprule
Language & Train Size & Test Size & \# Professions & Gender Labels \\
\midrule
English & 295,044 & 98,379 & 28 & Binary \\
Spanish & 54,179 & 18,090 & 72 & Binary \\
French & 49,373 & 16,478 & 27 & Binary \\
\bottomrule
\end{tabular}
}
\caption{Dataset statistics for multilingual BiasBios. Each sample contains a biography text paired with profession and gender labels. The main task is profession prediction, while gender information is used for bias evaluation through TPR-Gap.}
\label{appendix:dataset-stats-biasbios}
\end{table*}

\begin{table*}[!b]
\centering
\resizebox{\textwidth}{!}{
\begin{tabular}{lrr  rr  rr  rr  rr  rr rr}
\toprule
Target & \multicolumn{2}{c}{Baseline} & \multicolumn{2}{c}{SAL (EN)} & \multicolumn{2}{c}{SAL (DE)} & \multicolumn{2}{c}{SAL (FR)} & \multicolumn{2}{c}{\subsetswithout{IMSAE (FullyJoint)}} & \multicolumn{2}{c}{\subsetswithout{IMSAE (Subsets w/o)}} & \multicolumn{2}{c}{\subsetswith{IMSAE (Three-Subsets)}} \\

& Main & TPR-Gap & Main & TPR-Gap & Main & TPR-Gap & Main & TPR-Gap & Main & TPR-Gap & Main & TPR-Gap & Main & TPR-Gap \\

\midrule
\multicolumn{15}{l}{mBERT-uncased} \\
EN & 80.5 & 15.4 & \dab{0.1} 80.4 & \da{1.9} 13.5 & 80.5 & \ua{0.4} 15.8 & 80.5 & \ua{0.3} 15.7 & \dab{0.1} 80.4 & \da{1.2} 14.2 & 80.5 & \ua{0.2} 15.6 & \dab{0.1} 80.4 & \da{1.8} 13.6 \\
DE & 77.7 & 27.6 & \uag{0.1} 77.8 & \da{4.5} 23.1 & \dab{0.3} 77.4 & \da{0.3} 27.3 & \uag{0.1} 77.8 & 27.6 & \uag{0.2} 77.9 & \da{4.6} 23.0 & \uag{0.1} 77.8 & \da{1.9} 25.7 & \dab{0.1} 77.6 & \da{2.2} 25.4 \\
FR & 72.7 & 22.8 & \dab{0.1} 72.6 & \da{0.8} 22.0 & 72.7 & \da{0.7} 22.1 & \dab{0.5} 72.2 & \da{3.4} 19.4 & \dab{0.2} 72.5 & \da{1.5} 21.3 & \uag{0.1} 72.8 & \da{0.6} 22.2 & \dab{0.5} 72.2 & \da{3.2} 19.6 \\
\midrule
\multicolumn{15}{l}{Llama3-8B} \\
EN & 81.2 & 13.3 & \dab{2.2} 79.0 & \da{0.8} 12.5 & \dab{1.1} 80.1 & \da{0.6} 12.7 & \dab{0.5} 80.7 & \da{0.5} 12.8 & \dab{2.1} 79.1 & 13.3 & \dab{1.1} 80.1 & \da{0.2} 13.1 & \dab{2.2} 79.0 & \da{0.4} 12.9 \\
DE & 79.0 & 26.3 & \dab{0.3} 78.7 & \da{0.8} 25.5 & \dab{0.3} 78.7 & \ua{0.1} 26.4 & 79.0 & \da{0.9} 25.4 & \dab{0.3} 78.7 & \da{0.8} 25.5 & \dab{0.3} 78.7 & \da{0.8} 25.5 & \dab{0.4} 78.6 & \ua{0.2} 26.5 \\
FR & 72.8 & 26.1 & \uag{0.1} 72.9 & \ua{0.2} 26.3 & 72.8 & \ua{1.2} 27.3 & 72.8 & \da{4.5} 21.6 & 72.8 & \ua{1.1} 27.2 & 72.8 & 26.1 & \dab{0.1} 72.7 & \da{4.6} 21.5 \\
\midrule
\multicolumn{15}{l}{Llama-3.1-8B} \\
EN & 81.1 & 13.7 & \dab{2.1} 79.0 & \da{0.7} 13.0 & \dab{0.9} 80.2 & \da{0.1} 13.6 & \dab{0.8} 80.3 & \da{0.2} 13.5 & \dab{2.0} 79.1 & \da{1.1} 12.6 & \dab{1.1} 80.0 & \da{0.5} 13.2 & \dab{2.1} 79.0 & \da{0.5} 13.2 \\
DE & 79.8 & 26.8 & \dab{0.3} 79.5 & \ua{0.2} 27.0 & \dab{0.3} 79.5 & \da{5.2} 21.6 & \dab{0.2} 79.6 & \ua{0.5} 27.3 & \dab{0.2} 79.6 & \ua{0.3} 27.1 & \dab{0.4} 79.4 & \ua{0.4} 27.2 & \dab{0.2} 79.6 & \da{4.6} 22.2 \\
FR & 72.5 & 25.0 & \uag{0.1} 72.6 & \ua{0.1} 25.1 & 72.5 & \da{0.9} 24.1 & \dab{0.1} 72.4 & \da{5.7} 19.3 & \dab{0.1} 72.4 & \da{0.4} 24.6 & 72.5 & \da{0.5} 24.5 & \dab{0.1} 72.4 & \da{3.7} 21.3 \\
\midrule
\multicolumn{15}{l}{Llama-3.2-3B} \\
EN & 79.9 & 13.0 & \dab{1.1} 78.8 & \da{1.4} 11.6 & \dab{0.2} 79.7 & \da{0.5} 12.5 & \dab{0.2} 79.7 & \da{0.6} 12.4 & \dab{1.1} 78.8 & \da{1.0} 12.0 & \dab{0.5} 79.4 & \da{0.6} 12.4 & \dab{1.1} 78.8 & \da{1.5} 11.5 \\
DE & 78.2 & 27.9 & \dab{0.1} 78.1 & 27.9 & \dab{0.3} 77.9 & \da{0.6} 27.3 & 78.2 & 27.9 & \dab{0.1} 78.1 & 27.9 & \dab{0.2} 78.0 & \ua{0.2} 28.1 & \dab{0.4} 77.8 & \da{0.5} 27.4 \\
FR & 71.2 & 16.3 & \dab{0.2} 71.0 & \ua{1.1} 17.4 & 71.2 & \da{1.4} 14.9 & \dab{0.2} 71.0 & \da{0.9} 15.4 & \dab{0.2} 71.0 & \ua{0.3} 16.6 & 71.2 & \da{1.0} 15.3 & \dab{0.2} 71.0 & \da{1.5} 14.8 \\
\midrule
\multicolumn{15}{l}{Mistral-7B-Instruct-v0.3} \\
EN & 80.5 & 14.0 & \dab{2.7} 77.8 & \da{1.3} 12.7 & \dab{0.4} 80.1 & 14.0 & \dab{0.6} 79.9 & \da{0.1} 13.9 & \dab{2.8} 77.7 & \da{0.8} 13.2 & \dab{0.7} 79.8 & \da{0.1} 13.9 & \dab{2.8} 77.7 & \da{1.3} 12.7 \\
DE & 77.3 & 23.3 & \dab{0.2} 77.1 & \ua{0.3} 23.6 & 77.3 & 23.3 & \dab{0.2} 77.1 & \ua{0.6} 23.9 & \dab{0.2} 77.1 & \ua{0.3} 23.6 & \dab{0.3} 77.0 & \da{0.8} 22.5 & \dab{0.3} 77.0 & \ua{0.4} 23.7 \\
FR & 71.6 & 23.1 & \uag{0.2} 71.8 & \da{1.4} 21.7 & 71.6 & \da{1.0} 22.1 & \uag{0.2} 71.8 & \da{4.9} 18.2 & 71.6 & \da{1.2} 21.9 & \dab{0.1} 71.5 & \da{1.8} 21.3 & \uag{0.1} 71.7 & \da{5.1} 18.0 \\
\midrule
\multicolumn{15}{l}{Mistral-7B-v0.3} \\
EN & 80.9 & 13.9 & \dab{2.9} 78.0 & \da{1.2} 12.7 & \dab{0.9} 80.0 & \ua{0.1} 14.0 & \dab{1.2} 79.7 & 13.9 & \dab{2.9} 78.0 & \da{0.7} 13.2 & \dab{0.7} 80.2 & \da{0.1} 13.8 & \dab{3.0} 77.9 & \da{0.8} 13.1 \\
DE & 78.4 & 27.3 & \uag{0.1} 78.5 & \ua{0.2} 27.5 & \dab{0.7} 77.7 & \da{1.1} 26.2 & \dab{0.1} 78.3 & \ua{0.2} 27.5 & \uag{0.2} 78.6 & \ua{0.2} 27.5 & \dab{0.1} 78.3 & 27.3 & \dab{0.2} 78.2 & \da{1.2} 26.1 \\
FR & 72.2 & 23.7 & \uag{0.1} 72.3 & \da{2.8} 20.9 & \uag{0.1} 72.3 & \da{0.7} 23.0 & \dab{0.1} 72.1 & \da{4.4} 19.3 & 72.2 & \da{0.5} 23.2 & 72.2 & \da{1.1} 22.6 & \dab{0.1} 72.1 & \da{4.4} 19.3 \\
\midrule
\multicolumn{15}{l}{Mistral-Nemo-Base-2407} \\
EN & 82.2 & 13.1 & \dab{3.0} 79.2 & \da{0.2} 12.9 & \dab{0.7} 81.5 & \da{0.2} 12.9 & \dab{0.2} 82.0 & \da{0.4} 12.7 & \dab{3.2} 79.0 & \da{0.1} 13.0 & \dab{0.8} 81.4 & \ua{0.1} 13.2 & \dab{3.8} 78.4 & \da{0.6} 12.5 \\
DE & 79.4 & 31.0 & \uag{0.2} 79.6 & 31.0 & \uag{0.1} 79.5 & \da{1.0} 30.0 & \uag{0.2} 79.6 & \ua{0.1} 31.1 & \uag{0.2} 79.6 & 31.0 & \uag{0.1} 79.5 & \da{0.1} 30.9 & \uag{0.2} 79.6 & \da{1.0} 30.0 \\
FR & 73.9 & 22.0 & \uag{0.2} 74.1 & \ua{0.9} 22.9 & \uag{0.4} 74.3 & \ua{1.3} 23.3 & \uag{1.0} 74.9 & \da{0.3} 21.7 & \uag{0.3} 74.2 & \ua{0.7} 22.7 & \uag{0.2} 74.1 & \da{1.8} 20.2 & \uag{1.0} 74.9 & \ua{1.0} 23.0 \\
\midrule
\multicolumn{15}{l}{Mistral-Nemo-Instruct-2407} \\
EN & 81.1 & 12.5 & \dab{2.2} 78.9 & \da{0.6} 11.9 & \uag{0.3} 81.4 & \da{0.1} 12.4 & \uag{0.3} 81.4 & \da{0.4} 12.1 & \dab{2.5} 78.6 & \da{1.2} 11.3 & 81.1 & 12.5 & \dab{2.6} 78.5 & \da{0.9} 11.6 \\
DE & 78.4 & 26.6 & 78.4 & \da{0.2} 26.4 & \uag{0.4} 78.8 & \da{1.1} 25.5 & \dab{0.1} 78.3 & \da{0.6} 26.0 & \uag{0.1} 78.5 & \da{0.2} 26.4 & \uag{0.1} 78.5 & \da{0.3} 26.3 & \uag{0.6} 79.0 & \da{1.3} 25.3 \\
FR & 72.8 & 21.8 & \dab{0.2} 72.6 & \da{1.9} 19.9 & \dab{0.1} 72.7 & \da{1.9} 19.9 & \uag{0.6} 73.4 & \ua{1.1} 22.9 & \uag{0.1} 72.9 & \da{2.4} 19.4 & \dab{0.1} 72.7 & \da{2.8} 19.0 & \uag{0.4} 73.2 & \ua{0.7} 22.5 \\
\bottomrule

\end{tabular}
}

\caption{Gender debiasing performance evaluation on BiasBios across eight language models. We report both main task accuracy and TPR-Gap reduction for each model architecture and debiasing approach. IMSAE (Three-Subsets) demonstrates superior cross-lingual performance compared to monolingual methods, particularly for newer LLM architectures.}
\label{appendix:biasbios-fullresults}
\end{table*}

\clearpage

\begin{table*}[t]
\centering
\resizebox{\textwidth}{!}{

\begin{tabular}{l rr rr rr  rr rr rr  rr rr rr  rr}
\toprule
Target & \multicolumn{2}{c}{Baseline} & \multicolumn{2}{c}{SAL (EN)} & \multicolumn{2}{c}{SAL (DE)} & \multicolumn{2}{c}{SAL (FR)} & \multicolumn{2}{c}{ INLP (EN)} & \multicolumn{2}{c}{INLP (DE)} & \multicolumn{2}{c}{INLP (FR)} & \multicolumn{2}{c}{SentenceDebias (EN)} & \multicolumn{2}{c}{SentenceDebias (DE)} & \multicolumn{2}{c}{SentenceDebias (FR)} \\

& Main & Ext & Main & Ext & Main & Ext & Main & Ext & Main & Ext & Main & Ext & Main & Ext & Main & Ext & Main & Ext & Main & Ext \\

\midrule
\multicolumn{21}{l}{mBERT-uncased} \\
EN & 80.5 & 15.4 & \dab{0.1} 80.4 & \da{1.9} 13.5 & 80.5 & \ua{0.4} 15.8 & 80.5 & \ua{0.3} 15.7 & 80.5 & \da{0.2} 15.2 & 80.5 & \da{0.1} 15.3 & 80.5 & \da{0.5} 14.9 & \dab{0.2} 80.3 & \da{0.1} 15.3 & 80.5 & \da{0.1} 15.3 & 80.5 & \da{0.3} 15.1 \\
DE & 77.7 & 27.6 & \uag{0.1} 77.8 & \da{4.5} 23.1 & \dab{0.3} 77.4 & \da{0.3} 27.3 & \uag{0.1} 77.8 & 27.6 & \uag{0.1} 77.8 & \da{2.2} 25.4 & \uag{0.1} 77.8 & \da{1.5} 26.1 & 77.7 & 27.6 & \dab{0.1} 77.6 & \da{2.4} 25.2 & \dab{1.1} 76.6 & \da{0.5} 27.1 & \uag{0.2} 77.9 & \da{2.1} 25.5 \\
FR & 72.7 & 22.8 & \dab{0.1} 72.6 & \da{0.8} 22.0 & 72.7 & \da{0.7} 22.1 & \dab{0.5} 72.2 & \da{3.4} 19.4 & 72.7 & \da{0.5} 22.3 & \uag{0.1} 72.8 & \da{0.7} 22.1 & \dab{0.2} 72.5 & \da{1.1} 21.7 & \uag{0.1} 72.8 & \da{0.5} 22.3 & \dab{0.2} 72.5 & \da{0.1} 22.7 & \dab{0.2} 72.5 & \da{1.3} 21.5 \\
\midrule 
\multicolumn{21}{l}{Llama3-8B} \\
EN & 81.2 & 13.3 & \dab{2.2} 79.0 & \da{0.8} 12.5 & \dab{1.1} 80.1 & \da{0.6} 12.7 & \dab{0.5} 80.7 & \da{0.5} 12.8 & \dab{0.3} 80.9 & \da{0.6} 12.7 & \dab{0.3} 80.9 & \da{0.5} 12.8 & \dab{0.6} 80.6 & \da{0.5} 12.8 & \dab{0.4} 80.8 & \da{0.7} 12.6 & \dab{0.7} 80.5 & \da{0.5} 12.8 & \dab{0.5} 80.7 & \da{0.6} 12.7 \\
DE & 79.0 & 26.3 & \dab{0.3} 78.7 & \da{0.8} 25.5 & \dab{0.3} 78.7 & \ua{0.1} 26.4 & 79.0 & \da{0.9} 25.4 & 79.0 & 26.3 & \dab{0.2} 78.8 & \ua{0.1} 26.4 & \uag{0.1} 79.1 & 26.3 & \dab{0.1} 78.9 & \ua{0.1} 26.4 & \dab{0.9} 78.1 & \ua{0.7} 27.0 & \uag{0.1} 79.1 & \ua{1.3} 27.6 \\
FR & 72.8 & 26.1 & \uag{0.1} 72.9 & \ua{0.2} 26.3 & 72.8 & \ua{1.2} 27.3 & 72.8 & \da{4.5} 21.6 & \dab{0.1} 72.7 & \ua{1.0} 27.1 & \dab{0.1} 72.7 & \ua{1.6} 27.7 & \dab{0.1} 72.7 & \da{0.8} 25.3 & \uag{0.1} 72.9 & \da{0.7} 25.4 & 72.8 & \ua{0.3} 26.4 & 72.8 & \ua{0.8} 26.9 \\
\midrule 
\multicolumn{21}{l}{Llama-3.1-8B} \\
EN & 81.1 & 13.7 & \dab{2.1} 79.0 & \da{0.7} 13.0 & \dab{0.9} 80.2 & \da{0.1} 13.6 & \dab{0.8} 80.3 & \da{0.2} 13.5 & \dab{0.7} 80.4 & \ua{0.2} 13.9 & \dab{0.5} 80.6 & \da{0.2} 13.5 & \dab{0.7} 80.4 & 13.7 & \dab{0.5} 80.6 & \da{0.4} 13.3 & \dab{0.3} 80.8 & \ua{0.1} 13.8 & \dab{0.9} 80.2 & 13.7 \\
DE & 79.8 & 26.8 & \dab{0.3} 79.5 & \ua{0.2} 27.0 & \dab{0.3} 79.5 & \da{5.2} 21.6 & \dab{0.2} 79.6 & \ua{0.5} 27.3 & 79.8 & 26.8 & \uag{0.1} 79.9 & 26.8 & 79.8 & 26.8 & \dab{0.2} 79.6 & \ua{0.2} 27.0 & \dab{0.8} 79.0 & \da{2.7} 24.1 & 79.8 & \da{0.1} 26.7 \\
FR & 72.5 & 25.0 & \uag{0.1} 72.6 & \ua{0.1} 25.1 & 72.5 & \da{0.9} 24.1 & \dab{0.1} 72.4 & \da{5.7} 19.3 & 72.5 & \ua{1.1} 26.1 & 72.5 & \ua{2.0} 27.0 & \uag{0.1} 72.6 & \ua{0.7} 25.7 & \dab{0.1} 72.4 & \ua{1.0} 26.0 & \dab{0.1} 72.4 & \ua{0.2} 25.2 & \uag{0.2} 72.7 & \ua{0.2} 25.2 \\
\midrule 
\multicolumn{21}{l}{Llama-3.2-3B} \\
EN & 79.9 & 13.0 & \dab{1.1} 78.8 & \da{1.4} 11.6 & \dab{0.2} 79.7 & \da{0.5} 12.5 & \dab{0.2} 79.7 & \da{0.6} 12.4 & 79.9 & \ua{0.2} 13.2 & \uag{0.2} 80.1 & 13.0 & 79.9 & \da{0.1} 12.9 & \dab{0.1} 79.8 & 13.0 & \uag{0.1} 80.0 & \ua{0.2} 13.2 & \uag{0.6} 80.5 & \ua{0.4} 13.4 \\
DE & 78.2 & 27.9 & \dab{0.1} 78.1 & 27.9 & \dab{0.3} 77.9 & \da{0.6} 27.3 & 78.2 & 27.9 & 78.2 & 27.9 & 78.2 & 27.9 & 78.2 & 27.9 & 78.2 & 27.9 & \dab{1.0} 77.2 & \da{1.6} 26.3 & 78.2 & 27.9 \\
FR & 71.2 & 16.3 & \dab{0.2} 71.0 & \ua{1.1} 17.4 & 71.2 & \da{1.4} 14.9 & \dab{0.2} 71.0 & \da{0.9} 15.4 & \dab{0.2} 71.0 & \ua{0.7} 17.0 & \dab{0.1} 71.1 & \ua{1.1} 17.4 & \dab{0.2} 71.0 & \ua{1.3} 17.6 & 71.2 & \ua{2.1} 18.4 & \dab{0.1} 71.1 & \da{0.6} 15.7 & \dab{0.5} 70.7 & \da{1.2} 15.1 \\
\midrule 
\multicolumn{21}{l}{Mistral-7B-Instruct-v0.3} \\
EN & 80.5 & 14.0 & \dab{2.7} 77.8 & \da{1.3} 12.7 & \dab{0.4} 80.1 & 14.0 & \dab{0.6} 79.9 & \da{0.1} 13.9 & \dab{0.2} 80.3 & \da{0.2} 13.8 & \dab{0.2} 80.3 & \ua{0.3} 14.3 & \uag{0.1} 80.6 & \da{0.1} 13.9 & \dab{0.3} 80.2 & \da{0.2} 13.8 & \dab{0.5} 80.0 & 14.0 & \dab{0.4} 80.1 & \da{0.2} 13.8 \\
DE & 77.3 & 23.3 & \dab{0.2} 77.1 & \ua{0.3} 23.6 & 77.3 & 23.3 & \dab{0.2} 77.1 & \ua{0.6} 23.9 & \dab{0.1} 77.2 & 23.3 & 77.3 & 23.3 & \dab{0.1} 77.2 & 23.3 & \dab{0.2} 77.1 & \da{0.1} 23.2 & \dab{1.1} 76.2 & \da{0.4} 22.9 & 77.3 & 23.3 \\
FR & 71.6 & 23.1 & \uag{0.2} 71.8 & \da{1.4} 21.7 & 71.6 & \da{1.0} 22.1 & \uag{0.2} 71.8 & \da{4.9} 18.2 & \dab{0.2} 71.4 & \da{0.4} 22.7 & \uag{0.1} 71.7 & \da{0.3} 22.8 & 71.6 & \da{1.2} 21.9 & \dab{0.2} 71.4 & \da{2.1} 21.0 & 71.6 & \da{1.6} 21.5 & \dab{0.3} 71.3 & \da{2.7} 20.4 \\
\midrule 
\multicolumn{21}{l}{Mistral-7B-v0.3} \\
EN & 80.9 & 13.9 & \dab{2.9} 78.0 & \da{1.2} 12.7 & \dab{0.9} 80.0 & \ua{0.1} 14.0 & \dab{1.2} 79.7 & 13.9 & NaN & NaN & \dab{0.3} 80.6 & 13.9 & \dab{0.4} 80.5 & \da{0.1} 13.8 & \dab{0.3} 80.6 & \da{0.1} 13.8 & \uag{0.5} 81.4 & \ua{0.7} 14.6 & \dab{0.3} 80.6 & \ua{0.3} 14.2 \\
DE & 78.4 & 27.3 & \uag{0.1} 78.5 & \ua{0.2} 27.5 & \dab{0.7} 77.7 & \da{1.1} 26.2 & \dab{0.1} 78.3 & \ua{0.2} 27.5 & 78.4 & 27.3 & \dab{0.1} 78.3 & \ua{0.4} 27.7 & 78.4 & 27.3 & \uag{0.1} 78.5 & \da{0.1} 27.2 & \dab{0.3} 78.1 & \ua{1.0} 28.3 & 78.4 & 27.3 \\
FR & 72.2 & 23.7 & \uag{0.1} 72.3 & \da{2.8} 20.9 & \uag{0.1} 72.3 & \da{0.7} 23.0 & \dab{0.1} 72.1 & \da{4.4} 19.3 & \dab{0.1} 72.1 & \ua{1.2} 24.9 & 72.2 & \da{0.5} 23.2 & 72.2 & \da{0.1} 23.6 & 72.2 & \da{0.5} 23.2 & \dab{0.1} 72.1 & \da{0.2} 23.5 & \dab{0.3} 71.9 & \da{4.6} 19.1 \\
\midrule 
\multicolumn{21}{l}{Mistral-Nemo-Base-2407} \\
EN & 82.2 & 13.1 & \dab{3.0} 79.2 & \da{0.2} 12.9 & \dab{0.7} 81.5 & \da{0.2} 12.9 & \dab{0.2} 82.0 & \da{0.4} 12.7 & \dab{0.2} 82.0 & \da{0.7} 12.4 & \dab{0.5} 81.7 & \da{0.7} 12.4 & \uag{0.1} 82.3 & \da{0.2} 12.9 & \dab{0.4} 81.8 & \da{0.3} 12.8 & \dab{0.5} 81.7 & \ua{0.5} 13.6 & \dab{0.4} 81.8 & \da{0.4} 12.7 \\
DE & 79.4 & 31.0 & \uag{0.2} 79.6 & 31.0 & \uag{0.1} 79.5 & \da{1.0} 30.0 & \uag{0.2} 79.6 & \ua{0.1} 31.1 & \uag{0.1} 79.5 & \ua{0.1} 31.1 & 79.4 & \ua{0.1} 31.1 & 79.4 & 31.0 & \uag{0.1} 79.5 & 31.0 & \dab{0.1} 79.3 & \ua{0.6} 31.6 & 79.4 & \ua{0.1} 31.1 \\
FR & 73.9 & 22.0 & \uag{0.2} 74.1 & \ua{0.9} 22.9 & \uag{0.4} 74.3 & \ua{1.3} 23.3 & \uag{1.0} 74.9 & \da{0.3} 21.7 & \uag{0.2} 74.1 & \ua{1.0} 23.0 & \dab{0.3} 73.6 & \da{0.6} 21.4 & \dab{0.1} 73.8 & \ua{0.9} 22.9 & \dab{0.2} 73.7 & \da{1.1} 20.9 & \dab{0.1} 73.8 & \da{0.5} 21.5 & \uag{0.1} 74.0 & \da{0.1} 21.9 \\
\midrule 
\multicolumn{21}{l}{Mistral-Nemo-Instruct-2407} \\
EN & 81.1 & 12.5 & \dab{2.2} 78.9 & \da{0.6} 11.9 & \uag{0.3} 81.4 & \da{0.1} 12.4 & \uag{0.3} 81.4 & \da{0.4} 12.1 & \uag{0.2} 81.3 & \ua{0.5} 13.0 & \uag{0.4} 81.5 & \ua{0.5} 13.0 & \uag{0.5} 81.6 & \ua{0.2} 12.7 & \uag{0.3} 81.4 & \ua{0.1} 12.6 & \dab{0.4} 80.7 & \ua{0.3} 12.8 & \uag{0.1} 81.2 & \ua{0.2} 12.7 \\
DE & 78.4 & 26.6 & 78.4 & \da{0.2} 26.4 & \uag{0.4} 78.8 & \da{1.1} 25.5 & \dab{0.1} 78.3 & \da{0.6} 26.0 & 78.4 & 26.6 & \dab{0.3} 78.1 & \ua{2.7} 29.3 & \dab{0.1} 78.3 & \ua{4.1} 30.7 & \dab{0.4} 78.0 & \ua{3.1} 29.7 & \uag{0.3} 78.7 & \ua{5.8} 32.4 & \uag{0.1} 78.5 & \ua{0.2} 26.8 \\
FR & 72.8 & 21.8 & \dab{0.2} 72.6 & \da{1.9} 19.9 & \dab{0.1} 72.7 & \da{1.9} 19.9 & \uag{0.6} 73.4 & \ua{1.1} 22.9 & \dab{0.1} 72.7 & \da{1.3} 20.5 & \dab{0.1} 72.7 & \da{1.0} 20.8 & \dab{0.1} 72.7 & \da{0.6} 21.2 & \dab{0.5} 72.3 & \da{3.0} 18.8 & \dab{0.2} 72.6 & \da{2.1} 19.7 & \dab{0.1} 72.7 & \da{1.5} 20.3 \\
\bottomrule

\end{tabular}
}

\caption{Comparative analysis of crosslingual gender debiasing approaches on BiasBios, showing performance across different source-target language pairs. This analysis reveals how IMSAE leverages multilingual representations more effectively than single-source transfer methods.}
\label{appendix:biasbios-crosslingual}
\end{table*}

\clearpage

\section{Multilingual Hate Speech Details}

This appendix presents detailed information about our experiments on the Multilingual Twitter Hate Speech corpus. The detailed statistics for the Multilingual Hate Speech Dataset are presented in Table \ref{appendix:hate-speech-size} (training and test set sizes) and Table \ref{appendix:hate-speech-data-distribution} (class distribution across subsets). Comprehensive debiasing results comparing SAL and IMSAE across five languages and eight language models are provided for each demographic attribute: Age bias (Table \ref{Appendix:imsae-hate-speech-age}), Country bias (Table \ref{Appendix:imsae-hate-speech-country}), Gender bias (Table \ref{Appendix:imsae-hate-speech-gender}) and Race bias (Table \ref{Appendix:imsae-hate-speech-race}).

\subsection{Dataset Summary}

This section provides the detailed statistics of the Multilingual Hate Speech dataset, including the number of samples and distribution across different demographic attributes.

\begin{table*}[!b]
\centering
    \begin{tabular}{l rr rr}
        \hline
        \textbf{Language} & \textbf{Gender} & \textbf{Race} & \textbf{Age} & \textbf{Country} \\
        \hline
        English (en) & 31691/8746 & 31408/8646 & 31691/8746 & 36159/7373 \\
        \hline
        Spanish (es) & 1900/410 & 1900/410 & 1900/410 & 1956/439 \\
        \hline
        Italian (it) & 1605/418 & 1598/418 & 1605/418 & 2388/644 \\
        \hline
        Polish (pl) & 6806/1446 & 5649/1235 & 6806/1446 & 2155/471 \\
        \hline
        Portuguese (pt) & 816/163 & 816/163 & 816/163 & 757/197 \\
        \hline
    \end{tabular}
\caption{Training and test set sizes for different languages and demographic attributes in the Multilingual Hate Speech dataset. The diverse distribution across languages enables robust evaluation of cross-lingual transfer capabilities.}
\label{appendix:hate-speech-size}
\end{table*}

\begin{table*}[!b]
\centering
\small
\begin{tabular}{ll|rr|rr|rr|rr}
\toprule
& & \multicolumn{2}{c}{\textbf{Train}} & \multicolumn{2}{c}{\textbf{Dev}} & \multicolumn{2}{c}{\textbf{Test}} & \multicolumn{2}{c}{\textbf{Total}} \\
\textbf{Lang} & \textbf{Bias} & \textbf{C0} & \textbf{C1} & \textbf{C0} & \textbf{C1} & \textbf{C0} & \textbf{C1} & \textbf{C0} & \textbf{C1} \\
\midrule
\multirow{4}{*}{EN} 
& Gender & 13,017 & 18,674 & 3,640 & 3,134 & 3,791 & 4,955 & 20,448 & 26,763 \\
& Age & 13,467 & 16,635 & 3,488 & 2,780 & 2,922 & 5,465 & 19,877 & 24,880 \\
& Race & 19,475 & 11,933 & 3,578 & 3,123 & 3,844 & 4,802 & 26,897 & 19,858 \\
& Country & 13,719 & 22,440 & 3,400 & 5,021 & 2,393 & 4,980 & 19,512 & 32,441 \\
\midrule
\multirow{4}{*}{IT} 
& Gender & 1,091 & 514 & 256 & 103 & 290 & 128 & 1,637 & 745 \\
& Age & 791 & 809 & 178 & 180 & 229 & 189 & 1,198 & 1,178 \\
& Race & 1,568 & 30 & 354 & 3 & 413 & 5 & 2,335 & 38 \\
& Country & 610 & 1,778 & 107 & 345 & 79 & 565 & 796 & 2,688 \\
\midrule
\multirow{4}{*}{PL} 
& Gender & 3,552 & 3,254 & 787 & 674 & 716 & 730 & 5,055 & 4,658 \\
& Age & 1,300 & 4,349 & 281 & 918 & 276 & 959 & 1,857 & 6,226 \\
& Race & 4,030 & 1,619 & 860 & 339 & 879 & 356 & 5,769 & 2,314 \\
& Country & 15 & 2,140 & 3 & 486 & 3 & 468 & 21 & 3,094 \\
\midrule
\multirow{4}{*}{PT} 
& Gender & 682 & 134 & 76 & 74 & 60 & 103 & 818 & 311 \\
& Age & 737 & 79 & 92 & 58 & 68 & 95 & 897 & 232 \\
& Race & 634 & 182 & 84 & 66 & 86 & 77 & 804 & 325 \\
& Country & 341 & 416 & 88 & 110 & 66 & 131 & 495 & 657 \\
\midrule
\multirow{4}{*}{ES} 
& Gender & 997 & 903 & 210 & 197 & 251 & 159 & 1,458 & 1,259 \\
& Age & 923 & 977 & 197 & 210 & 239 & 171 & 1,359 & 1,358 \\
& Race & 1,027 & 873 & 228 & 179 & 236 & 174 & 1,491 & 1,226 \\
& Country & 1,334 & 622 & 277 & 159 & 236 & 203 & 1,847 & 984 \\
\bottomrule
\end{tabular}
\caption{Distribution of samples across different languages and protected attributes in the Multilingual Hate Speech dataset. C0 and C1 represent the two classes for each protected attribute. Note the inherent class imbalance in certain language-attribute combinations.}
\label{appendix:hate-speech-data-distribution}
\end{table*}

\clearpage

\subsection{IMSAE Debiasing Results}
\label{appendix:hate-speech}

This section presents the complete results of applying IMSAE for bias mitigation across different demographic attributes in the Multilingual Hate Speech dataset. Our evaluation compares both standard and zero-shot settings by: Age bias (Table \ref{Appendix:imsae-hate-speech-age}), Country bias (Table \ref{Appendix:imsae-hate-speech-country}), Gender bias (Table \ref{Appendix:imsae-hate-speech-gender}) and Race bias (Table \ref{Appendix:imsae-hate-speech-race}).

\begin{table*}[b]
\centering
\resizebox{\textwidth}{!}{
\begin{tabular}{lrr rr rr rr rr rr rr rr rr}
\toprule
Target & \multicolumn{2}{c}{Baseline} & \multicolumn{2}{c}{SAL (EN)} & \multicolumn{2}{c}{SAL (ES)} & \multicolumn{2}{c}{SAL (IT)} & \multicolumn{2}{c}{SAL (PL)} & \multicolumn{2}{c}{SAL (PT)} & \multicolumn{2}{c}{\subsetswithout{IMSAE (FullyJoint)}} & \multicolumn{2}{c}{\subsetswithout{IMSAE (Subsets w/o)}} & \multicolumn{2}{c}{\subsetswith{IMSAE (Five-Subsets)}}  \\
       & Main  & TPR-Gap  & Main  & TPR-Gap  & Main  & TPR-Gap  & Main  & TPR-Gap  & Main  & TPR-Gap  & Main  & TPR-Gap  & Main  & TPR-Gap  & Main  & TPR-Gap  & Main  & TPR-Gap \\
\midrule 
\multicolumn{17}{l}{mBERT-uncased} \\
EN & 86.7 & 9.1 & \uag{0.3} 87.0 & \ua{0.4} 9.5 & \dab{0.5} 86.2 & \da{0.3} 8.8 & 86.7 & \ua{0.1} 9.2 & \dab{0.5} 86.2 & \da{0.1} 9.0 & 86.7 & \da{0.1} 9.0 & 86.7 & \ua{0.2} 9.3 & 86.7 & \da{9.0} 0.1 & \uag{0.2} 86.9 & \da{9.0} 0.1 \\
ES & 63.7 & 12.9 & \uag{0.4} 64.1 & \ua{2.1} 15.0 & \dab{0.3} 63.4 & \da{0.5} 12.4 & \uag{0.2} 63.9 & \da{0.3} 12.6 & \uag{0.2} 63.9 & \ua{0.3} 13.2 & \dab{1.0} 62.7 & \ua{0.3} 13.2 & \dab{0.3} 63.4 & \da{0.4} 12.5 & \uag{0.2} 63.9 & \da{8.0} 4.9 & \uag{0.4} 64.1 & \da{6.3} 6.6 \\
IT & 68.2 & 3.6 & \dab{0.3} 67.9 & \ua{0.3} 3.9 & 68.2 & 3.6 & \dab{0.3} 67.9 & \ua{0.3} 3.9 & 68.2 & \da{0.3} 3.3 & \uag{0.2} 68.4 & \ua{0.4} 4.0 & \dab{0.3} 67.9 & \ua{0.3} 3.9 & \dab{1.2} 67.0 & \da{2.8} 0.8 & \dab{0.5} 67.7 & \da{0.7} 2.9 \\
PL & 91.3 & 8.8 & 91.3 & \da{0.7} 8.1 & \dab{0.1} 91.2 & \da{1.3} 7.5 & 91.3 & \da{0.7} 8.1 & \dab{0.9} 90.4 & \da{1.9} 6.9 & 91.3 & \da{0.7} 8.1 & \dab{0.1} 91.2 & \da{1.3} 7.5 & \dab{0.4} 90.9 & \da{5.0} 3.8 & \dab{0.9} 90.4 & \da{8.8} 0.0 \\
PT & 61.3 & 17.6 & \dab{1.2} 60.1 & \ua{1.2} 18.8 & \dab{1.2} 60.1 & \da{0.6} 17.0 & \dab{1.2} 60.1 & \ua{1.2} 18.8 & 61.3 & 17.6 & \dab{0.6} 60.7 & \ua{1.5} 19.1 & 61.3 & \da{1.9} 15.7 & \uag{1.3} 62.6 & \ua{3.4} 21.0 & \uag{1.3} 62.6 & \da{3.8} 13.8 \\
\midrule 
\multicolumn{17}{l}{Llama3-8B} \\
EN & 79.6 & 7.8 & \uag{0.8} 80.4 & \ua{0.6} 8.4 & 79.6 & \da{0.4} 7.4 & \uag{0.1} 79.7 & \da{0.1} 7.7 & \uag{0.1} 79.7 & \da{0.2} 7.6 & \uag{0.3} 79.9 & \da{0.3} 7.5 & \uag{0.1} 79.7 & 7.8 & \uag{0.1} 79.7 & \da{7.3} 0.5 & \uag{1.0} 80.6 & \da{0.4} 7.4 \\
ES & 70.7 & 11.7 & \uag{0.3} 71.0 & \ua{0.6} 12.3 & \dab{0.2} 70.5 & \da{0.6} 11.1 & \uag{0.3} 71.0 & \ua{0.6} 12.3 & 70.7 & 11.7 & \uag{0.5} 71.2 & \ua{0.5} 12.2 & 70.7 & 11.7 & 70.7 & \da{1.1} 10.6 & \dab{0.2} 70.5 & \ua{1.2} 12.9 \\
IT & 69.9 & 8.0 & \dab{0.3} 69.6 & \ua{0.3} 8.3 & 69.9 & 8.0 & \dab{0.5} 69.4 & \da{0.7} 7.3 & \uag{0.2} 70.1 & \da{0.9} 7.1 & \dab{0.3} 69.6 & \da{0.3} 7.7 & \dab{0.5} 69.4 & \ua{1.2} 9.2 & \dab{0.3} 69.6 & \da{0.1} 7.9 & \uag{0.2} 70.1 & \da{3.8} 4.2 \\
PL & 91.0 & 17.2 & 91.0 & \da{0.7} 16.5 & \dab{0.1} 90.9 & \da{0.6} 16.6 & 91.0 & \da{0.1} 17.1 & \dab{1.2} 89.8 & \da{7.8} 9.4 & 91.0 & \da{0.8} 16.4 & \dab{0.1} 90.9 & \da{0.7} 16.5 & 91.0 & \ua{3.7} 20.9 & \dab{1.3} 89.7 & \da{9.1} 8.1 \\
PT & 57.1 & 2.1 & \dab{0.7} 56.4 & \ua{0.9} 3.0 & \uag{1.2} 58.3 & \ua{1.9} 4.0 & \dab{0.7} 56.4 & \ua{0.9} 3.0 & 57.1 & 2.1 & 57.1 & 2.1 & \dab{1.3} 55.8 & \ua{1.4} 3.5 & \dab{1.3} 55.8 & \ua{3.2} 5.3 & \dab{0.7} 56.4 & \ua{8.3} 10.4 \\
\midrule 
\multicolumn{17}{l}{Llama-3.1-8B} \\
EN & 79.7 & 6.5 & \uag{0.7} 80.4 & \ua{0.7} 7.2 & 79.7 & \da{0.1} 6.4 & 79.7 & \ua{0.2} 6.7 & \uag{0.1} 79.8 & \ua{0.1} 6.6 & 79.7 & \da{0.2} 6.3 & \uag{0.1} 79.8 & \da{0.3} 6.2 & \uag{0.1} 79.8 & \da{2.1} 4.4 & \uag{1.0} 80.7 & \da{4.1} 2.4 \\
ES & 72.9 & 8.6 & \dab{0.2} 72.7 & \ua{0.1} 8.7 & \dab{0.2} 72.7 & \da{1.0} 7.6 & 72.9 & \da{1.1} 7.5 & \dab{0.5} 72.4 & \da{0.5} 8.1 & 72.9 & \da{2.2} 6.4 & 72.9 & \ua{0.7} 9.3 & \dab{0.9} 72.0 & \ua{2.6} 11.2 & \dab{0.5} 72.4 & \da{2.7} 5.9 \\
IT & 66.5 & 9.5 & \uag{0.2} 66.7 & \da{1.2} 8.3 & \uag{0.5} 67.0 & \da{1.0} 8.5 & \uag{0.7} 67.2 & \da{0.5} 9.0 & \uag{0.5} 67.0 & \da{1.0} 8.5 & \uag{0.5} 67.0 & \da{0.1} 9.4 & \uag{0.2} 66.7 & \da{1.2} 8.3 & \uag{0.7} 67.2 & \da{0.9} 8.6 & 66.5 & \ua{2.2} 11.7 \\
PL & 90.4 & 15.2 & \dab{0.1} 90.3 & \da{0.6} 14.6 & 90.4 & 15.2 & 90.4 & 15.2 & \dab{0.4} 90.0 & \da{6.9} 8.3 & \dab{0.1} 90.3 & \da{0.6} 14.6 & \dab{0.1} 90.3 & \da{0.6} 14.6 & 90.4 & \ua{2.1} 17.3 & \dab{0.7} 89.7 & \da{11.4} 3.8 \\
PT & 59.5 & 2.8 & \uag{0.6} 60.1 & \ua{0.5} 3.3 & \uag{0.6} 60.1 & \da{1.4} 1.4 & 59.5 & 2.8 & 59.5 & 2.8 & \uag{0.6} 60.1 & \ua{0.5} 3.3 & 59.5 & \ua{1.5} 4.3 & \uag{0.6} 60.1 & \ua{1.8} 4.6 & \dab{1.8} 57.7 & \ua{1.7} 4.5 \\
\midrule 
\multicolumn{17}{l}{Llama-3.2-3B} \\
EN & 79.7 & 6.2 & \uag{0.6} 80.3 & \ua{0.8} 7.0 & \uag{0.1} 79.8 & \ua{0.3} 6.5 & 79.7 & 6.2 & \uag{0.1} 79.8 & \ua{0.1} 6.3 & \uag{0.1} 79.8 & \ua{0.3} 6.5 & 79.7 & \ua{0.4} 6.6 & 79.7 & \da{6.0} 0.2 & \uag{0.7} 80.4 & \da{6.0} 0.2 \\
ES & 68.3 & 12.1 & \dab{0.3} 68.0 & \da{1.2} 10.9 & 68.3 & \ua{1.0} 13.1 & 68.3 & 12.1 & \uag{0.2} 68.5 & \ua{0.9} 13.0 & \dab{0.5} 67.8 & \ua{0.9} 13.0 & 68.3 & 12.1 & 68.3 & \da{0.5} 11.6 & \uag{0.2} 68.5 & \da{2.3} 9.8 \\
IT & 67.7 & 5.0 & 67.7 & \ua{0.3} 5.3 & \uag{0.5} 68.2 & \ua{2.0} 7.0 & \dab{0.2} 67.5 & \ua{0.7} 5.7 & \uag{0.2} 67.9 & \ua{1.8} 6.8 & \uag{0.2} 67.9 & \ua{1.5} 6.5 & \uag{0.2} 67.9 & \ua{0.5} 5.5 & 67.7 & \ua{3.5} 8.5 & \dab{0.2} 67.5 & \ua{0.8} 5.8 \\
PL & 90.9 & 17.1 & \dab{0.2} 90.7 & \ua{0.1} 17.2 & \dab{0.2} 90.7 & \da{0.6} 16.5 & \dab{0.3} 90.6 & \da{0.6} 16.5 & \dab{0.9} 90.0 & \da{1.4} 15.7 & 90.9 & 17.1 & \dab{0.2} 90.7 & \ua{0.1} 17.2 & \dab{0.3} 90.6 & \da{10.7} 6.4 & \dab{0.9} 90.0 & \ua{7.4} 24.5 \\
PT & 56.4 & 8.9 & \uag{0.7} 57.1 & \ua{2.8} 11.7 & \uag{0.7} 57.1 & \ua{0.8} 9.7 & \uag{0.7} 57.1 & \ua{0.8} 9.7 & \dab{1.2} 55.2 & \ua{1.5} 10.4 & 56.4 & \ua{1.7} 10.6 & \dab{1.2} 55.2 & \ua{1.5} 10.4 & \uag{1.3} 57.7 & \ua{7.7} 16.6 & \uag{0.7} 57.1 & \ua{11.1} 20.0 \\
\midrule 
\multicolumn{17}{l}{Mistral-7B-Instruct-v0.3} \\
EN & 79.4 & 7.1 & \uag{1.1} 80.5 & \ua{1.3} 8.4 & \uag{0.2} 79.6 & \ua{0.1} 7.2 & \uag{0.3} 79.7 & \ua{0.1} 7.2 & \uag{0.1} 79.5 & \ua{0.3} 7.4 & \uag{0.2} 79.6 & \da{0.1} 7.0 & \uag{0.2} 79.6 & \ua{0.2} 7.3 & \uag{0.1} 79.5 & \ua{0.6} 7.7 & \uag{0.7} 80.1 & \da{5.6} 1.5 \\
ES & 64.6 & 7.2 & \dab{0.7} 63.9 & \da{0.2} 7.0 & \uag{0.8} 65.4 & \ua{1.6} 8.8 & \dab{0.2} 64.4 & \da{0.8} 6.4 & \dab{0.5} 64.1 & \ua{0.5} 7.7 & \dab{0.5} 64.1 & \da{1.5} 5.7 & \dab{1.2} 63.4 & \da{0.1} 7.1 & 64.6 & 7.2 & \dab{1.2} 63.4 & \ua{13.0} 20.2 \\
IT & 66.0 & 3.5 & 66.0 & 3.5 & \dab{0.2} 65.8 & \da{0.2} 3.3 & \uag{0.3} 66.3 & \da{0.3} 3.2 & \uag{0.5} 66.5 & \ua{0.1} 3.6 & \dab{0.2} 65.8 & \da{0.2} 3.3 & \dab{0.2} 65.8 & \da{0.5} 3.0 & 66.0 & \ua{1.4} 4.9 & \dab{0.2} 65.8 & \da{0.1} 3.4 \\
PL & 91.0 & 19.5 & 91.0 & 19.5 & \uag{0.2} 91.2 & \ua{1.2} 20.7 & \dab{0.1} 90.9 & \da{0.6} 18.9 & \dab{0.8} 90.2 & \da{3.8} 15.7 & \uag{0.2} 91.2 & \ua{1.3} 20.8 & \uag{0.1} 91.1 & \ua{0.7} 20.2 & \uag{0.2} 91.2 & \ua{6.8} 26.3 & \dab{0.7} 90.3 & \da{0.3} 19.2 \\
PT & 59.5 & 17.2 & 59.5 & 17.2 & \dab{0.6} 58.9 & \da{1.0} 16.2 & \dab{0.6} 58.9 & \da{1.0} 16.2 & 59.5 & 17.2 & \dab{0.6} 58.9 & \ua{1.8} 19.0 & 59.5 & 17.2 & \dab{0.6} 58.9 & \da{0.4} 16.8 & \dab{1.2} 58.3 & \ua{0.2} 17.4 \\
\midrule 
\multicolumn{17}{l}{Mistral-7B-v0.3} \\
EN & 79.8 & 7.1 & \uag{0.6} 80.4 & \ua{0.4} 7.5 & \dab{0.1} 79.7 & 7.1 & 79.8 & \da{0.3} 6.8 & \dab{0.1} 79.7 & \ua{0.1} 7.2 & \dab{0.2} 79.6 & \da{0.3} 6.8 & \dab{0.1} 79.7 & \da{0.4} 6.7 & \dab{0.3} 79.5 & \da{0.7} 6.4 & \uag{0.6} 80.4 & \ua{2.2} 9.3 \\
ES & 67.1 & 12.8 & 67.1 & \da{1.4} 11.4 & \dab{0.3} 66.8 & \da{0.8} 12.0 & \dab{0.3} 66.8 & 12.8 & \dab{0.3} 66.8 & \ua{1.8} 14.6 & 67.1 & \ua{0.2} 13.0 & 67.1 & \da{0.6} 12.2 & 67.1 & \da{7.0} 5.8 & \dab{1.0} 66.1 & \da{11.5} 1.3 \\
IT & 65.8 & 5.6 & 65.8 & \da{0.9} 4.7 & \dab{0.2} 65.6 & \ua{0.5} 6.1 & \uag{0.2} 66.0 & \da{0.6} 5.0 & 65.8 & \da{0.1} 5.5 & 65.8 & \da{0.8} 4.8 & \uag{0.2} 66.0 & \da{1.3} 4.3 & \uag{0.7} 66.5 & \da{2.1} 3.5 & 65.8 & \da{3.7} 1.9 \\
PL & 90.4 & 17.8 & \uag{0.3} 90.7 & \ua{0.4} 18.2 & 90.4 & 17.8 & 90.4 & \da{0.8} 17.0 & \dab{0.9} 89.5 & \da{5.8} 12.0 & \dab{0.1} 90.3 & \ua{0.1} 17.9 & 90.4 & 17.8 & \uag{0.2} 90.6 & \da{17.2} 0.6 & \dab{0.7} 89.7 & \ua{4.0} 21.8 \\
PT & 57.7 & 12.4 & 57.7 & 12.4 & 57.7 & 12.4 & 57.7 & 12.4 & 57.7 & 12.4 & 57.7 & 12.4 & 57.7 & 12.4 & 57.7 & \ua{0.7} 13.1 & 57.7 & 12.4 \\
\midrule 
\multicolumn{17}{l}{Mistral-Nemo-Base-2407} \\
EN & 78.9 & 6.4 & \uag{0.4} 79.3 & \ua{1.1} 7.5 & \dab{0.3} 78.6 & \ua{0.1} 6.5 & \dab{0.5} 78.4 & \ua{0.2} 6.6 & 78.9 & \da{0.1} 6.3 & \dab{0.3} 78.6 & \da{0.3} 6.1 & \dab{0.5} 78.4 & \da{0.4} 6.0 & \dab{0.4} 78.5 & \ua{3.7} 10.1 & \uag{0.3} 79.2 & \ua{1.8} 8.2 \\
ES & 72.2 & 16.3 & \dab{1.2} 71.0 & \ua{1.3} 17.6 & \uag{0.7} 72.9 & \ua{1.8} 18.1 & 72.2 & \ua{2.0} 18.3 & \uag{0.2} 72.4 & \ua{0.6} 16.9 & \dab{0.5} 71.7 & \da{1.0} 15.3 & \dab{0.7} 71.5 & \ua{1.5} 17.8 & \dab{0.2} 72.0 & \da{3.9} 12.4 & \uag{0.2} 72.4 & \da{0.4} 15.9 \\
IT & 64.8 & 5.8 & \uag{0.5} 65.3 & \ua{1.1} 6.9 & \uag{0.3} 65.1 & \da{1.4} 4.4 & \dab{0.4} 64.4 & 5.8 & \dab{0.2} 64.6 & \ua{0.5} 6.3 & \uag{0.5} 65.3 & \ua{0.1} 5.9 & \uag{0.3} 65.1 & \ua{1.0} 6.8 & \uag{0.5} 65.3 & \ua{0.7} 6.5 & \uag{0.5} 65.3 & \ua{4.9} 10.7 \\
PL & 91.3 & 20.9 & \dab{0.3} 91.0 & \da{0.7} 20.2 & \dab{0.1} 91.2 & \da{0.6} 20.3 & \dab{0.1} 91.2 & \da{0.6} 20.3 & \dab{1.0} 90.3 & \da{3.4} 17.5 & \dab{0.1} 91.2 & 20.9 & \dab{0.2} 91.1 & \da{0.7} 20.2 & \dab{0.4} 90.9 & \da{8.1} 12.8 & \dab{1.1} 90.2 & \da{12.6} 8.3 \\
PT & 63.2 & 8.8 & 63.2 & 8.8 & \dab{0.6} 62.6 & \da{1.6} 7.2 & 63.2 & \da{3.0} 5.8 & \uag{0.6} 63.8 & \da{1.4} 7.4 & 63.2 & 8.8 & 63.2 & \da{3.0} 5.8 & \uag{1.2} 64.4 & \ua{2.5} 11.3 & 63.2 & \da{2.0} 6.8 \\
\midrule 
\multicolumn{17}{l}{Mistral-Nemo-Instruct-2407} \\
EN & 79.3 & 5.9 & \uag{0.3} 79.6 & \ua{1.1} 7.0 & \dab{0.2} 79.1 & \ua{0.1} 6.0 & \dab{0.2} 79.1 & \ua{0.2} 6.1 & \dab{0.2} 79.1 & \ua{0.1} 6.0 & \dab{0.3} 79.0 & \ua{0.5} 6.4 & \dab{0.3} 79.0 & \ua{0.6} 6.5 & \dab{0.2} 79.1 & \da{2.0} 3.9 & \uag{0.4} 79.7 & \da{5.2} 0.7 \\
ES & 71.0 & 10.2 & \dab{0.5} 70.5 & \ua{1.2} 11.4 & \dab{0.8} 70.2 & \ua{1.2} 11.4 & \dab{0.5} 70.5 & \ua{0.9} 11.1 & \dab{1.0} 70.0 & \da{0.6} 9.6 & \dab{0.3} 70.7 & \ua{1.6} 11.8 & \dab{0.5} 70.5 & \ua{0.2} 10.4 & \dab{0.3} 70.7 & \da{3.4} 6.8 & \dab{2.2} 68.8 & \da{2.6} 7.6 \\
IT & 65.3 & 8.0 & \dab{0.2} 65.1 & \da{0.6} 7.4 & \dab{0.2} 65.1 & \da{0.1} 7.9 & 65.3 & \da{0.4} 7.6 & 65.3 & 8.0 & 65.3 & \ua{0.5} 8.5 & \dab{0.2} 65.1 & \da{0.6} 7.4 & \uag{0.5} 65.8 & \ua{0.9} 8.9 & \dab{0.5} 64.8 & \da{1.2} 6.8 \\
PL & 90.9 & 18.9 & \dab{0.1} 90.8 & \ua{0.1} 19.0 & 90.9 & 18.9 & 90.9 & 18.9 & \dab{0.7} 90.2 & \da{1.9} 17.0 & 90.9 & 18.9 & \dab{0.3} 90.6 & \ua{0.1} 19.0 & 90.9 & \da{2.5} 16.4 & \dab{0.9} 90.0 & \ua{1.1} 20.0 \\
PT & 64.4 & 1.5 & 64.4 & 1.5 & 64.4 & 1.5 & \dab{0.6} 63.8 & \ua{1.8} 3.3 & \uag{1.2} 65.6 & \ua{2.1} 3.6 & 64.4 & 1.5 & \uag{0.6} 65.0 & \ua{1.0} 2.5 & 64.4 & \ua{3.1} 4.6 & 64.4 & \ua{1.7} 3.2 \\
\bottomrule

\end{tabular}%
}
\caption{Comprehensive evaluation of age bias mitigation on the Multilingual Hate Speech dataset across eight language models. We report hate speech prediction accuracy (Main) and True Positive Rate gap (TPR-Gap) between demographic groups for each model and debiasing approach. IMSAE (Five-Subsets) consistently achieves stronger bias reduction compared to monolingual and standard cross-lingual approaches.}
\label{Appendix:imsae-hate-speech-age}
\end{table*}

\clearpage

\begin{table*}[t]
\centering
\resizebox{\textwidth}{!}{
\begin{tabular}{lrr rr rr rr rr rr rr rr rr}
\toprule
Target & \multicolumn{2}{c}{Baseline} & \multicolumn{2}{c}{SAL (EN)} & \multicolumn{2}{c}{SAL (ES)} & \multicolumn{2}{c}{SAL (IT)} & \multicolumn{2}{c}{SAL (PL)} & \multicolumn{2}{c}{SAL (PT)} & \multicolumn{2}{c}{\subsetswithout{IMSAE (FullyJoint)}} & \multicolumn{2}{c}{\subsetswithout{IMSAE (Subsets w/o)}} & \multicolumn{2}{c}{\subsetswith{IMSAE (Five-Subsets)}}  \\

& Main  & TPR-Gap  & Main  & TPR-Gap  & Main  & TPR-Gap  & Main  & TPR-Gap  & Main  & TPR-Gap  & Main  & TPR-Gap  & Main  & TPR-Gap  & Main  & TPR-Gap  & Main  & TPR-Gap \\

\midrule 
\multicolumn{17}{l}{mBERT-uncased} \\
EN & 82.3 & 6.7 & \dab{0.1} 82.2 & \da{0.8} 5.9 & 82.3 & 6.7 & 82.3 & \da{0.1} 6.6 & 82.3 & \ua{0.1} 6.8 & 82.3 & 6.7 & 82.3 & 6.7 & \dab{0.1} 82.2 & \da{5.4} 1.3 & 82.3 & \da{6.6} 0.1 \\
ES & 65.1 & 5.1 & \dab{0.4} 64.7 & \ua{0.8} 5.9 & 65.1 & \da{0.2} 4.9 & \dab{0.2} 64.9 & \ua{0.2} 5.3 & \dab{0.2} 64.9 & \da{0.2} 4.9 & \uag{0.3} 65.4 & \ua{0.8} 5.9 & \dab{0.4} 64.7 & \ua{1.8} 6.9 & \dab{0.6} 64.5 & \da{0.4} 4.7 & \dab{0.4} 64.7 & \da{3.0} 2.1 \\
IT & 71.0 & 1.7 & \dab{0.2} 70.8 & \ua{0.4} 2.1 & 71.0 & \ua{0.8} 2.5 & \uag{0.1} 71.1 & \da{0.4} 1.3 & \dab{0.2} 70.8 & \ua{0.4} 2.1 & \dab{0.2} 70.8 & \ua{0.4} 2.1 & \dab{0.2} 70.8 & \ua{0.4} 2.1 & 71.0 & \ua{1.6} 3.3 & \dab{0.5} 70.5 & \da{1.1} 0.6 \\
PL & 99.6 & 0.0 & 99.6 & 0.0 & 99.6 & 0.0 & 99.6 & 0.0 & 99.6 & 0.0 & 99.6 & 0.0 & 99.6 & 0.0 & 99.6 & 0.0 & 99.6 & 0.0 \\
PT & 64.5 & 5.1 & 64.5 & \ua{1.5} 6.6 & \uag{1.0} 65.5 & 5.1 & 64.5 & 5.1 & \uag{1.0} 65.5 & \ua{0.5} 5.6 & \dab{0.5} 64.0 & \da{4.3} 0.8 & \uag{2.0} 66.5 & \ua{4.4} 9.5 & \uag{2.0} 66.5 & \da{4.9} 0.2 & \dab{1.0} 63.5 & \da{4.2} 0.9 \\
\midrule 
\multicolumn{17}{l}{Llama3-8B} \\
EN & 77.1 & 6.0 & 77.1 & \da{1.0} 5.0 & 77.1 & \ua{0.4} 6.4 & 77.1 & 6.0 & \dab{0.1} 77.0 & \ua{0.1} 6.1 & \dab{0.1} 77.0 & \ua{0.3} 6.3 & 77.1 & \ua{0.1} 6.1 & \dab{0.2} 76.9 & \da{3.5} 2.5 & \dab{0.3} 76.8 & \da{4.7} 1.3 \\
ES & 66.7 & 10.0 & \uag{1.2} 67.9 & \ua{1.0} 11.0 & \uag{0.3} 67.0 & \ua{0.1} 10.1 & 66.7 & 10.0 & 66.7 & 10.0 & \uag{0.7} 67.4 & \da{1.3} 8.7 & \uag{0.7} 67.4 & \ua{0.6} 10.6 & \uag{1.2} 67.9 & \da{6.0} 4.0 & \uag{1.4} 68.1 & \ua{2.9} 12.9 \\
IT & 70.5 & 12.4 & \uag{0.5} 71.0 & \ua{0.4} 12.8 & \uag{0.6} 71.1 & \ua{0.7} 13.1 & \uag{0.3} 70.8 & \ua{0.2} 12.6 & \uag{0.2} 70.7 & \ua{0.3} 12.7 & \uag{0.8} 71.3 & \ua{0.6} 13.0 & \uag{0.3} 70.8 & \ua{0.6} 13.0 & \uag{1.2} 71.7 & \da{1.1} 11.3 & \uag{0.6} 71.1 & \da{6.5} 5.9 \\
PL & 99.6 & 0.0 & 99.6 & 0.0 & 99.6 & 0.0 & 99.6 & 0.0 & 99.6 & 0.0 & 99.6 & 0.0 & 99.6 & 0.0 & 99.6 & 0.0 & 99.6 & 0.0 \\
PT & 67.5 & 2.9 & \uag{0.5} 68.0 & \ua{0.5} 3.4 & \uag{1.0} 68.5 & \ua{1.5} 4.4 & \uag{1.0} 68.5 & \da{0.2} 2.7 & \dab{0.5} 67.0 & \ua{0.1} 3.0 & \uag{0.5} 68.0 & \ua{2.5} 5.4 & \uag{0.5} 68.0 & \ua{0.5} 3.4 & 67.5 & \da{0.4} 2.5 & \dab{0.5} 67.0 & \da{1.6} 1.3 \\
\midrule 
\multicolumn{17}{l}{Llama-3.1-8B} \\
EN & 77.1 & 5.7 & \dab{0.2} 76.9 & \da{0.6} 5.1 & 77.1 & \da{0.1} 5.6 & \dab{0.1} 77.0 & \ua{0.2} 5.9 & 77.1 & \ua{0.2} 5.9 & \dab{0.1} 77.0 & 5.7 & 77.1 & \ua{0.1} 5.8 & \uag{0.1} 77.2 & \da{5.6} 0.1 & \dab{0.4} 76.7 & \da{4.0} 1.7 \\
ES & 70.4 & 22.2 & \dab{0.5} 69.9 & \ua{2.8} 25.0 & \dab{0.2} 70.2 & \da{0.6} 21.6 & \dab{0.5} 69.9 & 22.2 & 70.4 & 22.2 & \dab{0.2} 70.2 & \ua{0.3} 22.5 & \uag{0.2} 70.6 & \ua{0.6} 22.8 & \dab{0.2} 70.2 & \da{9.3} 12.9 & 70.4 & \da{6.8} 15.4 \\
IT & 68.6 & 12.7 & \uag{0.3} 68.9 & \ua{0.5} 13.2 & \uag{0.2} 68.8 & \da{0.1} 12.6 & 68.6 & \da{0.4} 12.3 & 68.6 & \da{0.4} 12.3 & \uag{0.2} 68.8 & \da{0.1} 12.6 & \uag{0.3} 68.9 & \da{0.1} 12.6 & \uag{0.3} 68.9 & \da{1.1} 11.6 & \dab{0.3} 68.3 & \da{2.1} 10.6 \\
PL & 99.6 & 0.0 & 99.6 & 0.0 & 99.6 & 0.0 & 99.6 & 0.0 & 99.6 & 0.0 & 99.6 & 0.0 & 99.6 & 0.0 & 99.6 & 0.0 & 99.6 & 0.0 \\
PT & 66.5 & 5.4 & 66.5 & \da{3.7} 1.7 & \uag{1.0} 67.5 & \da{1.6} 3.8 & 66.5 & 5.4 & \uag{0.5} 67.0 & \da{1.4} 4.0 & \uag{0.5} 67.0 & \da{0.2} 5.2 & \uag{0.5} 67.0 & \da{1.4} 4.0 & \uag{1.5} 68.0 & \da{0.3} 5.1 & \uag{0.5} 67.0 & \da{0.1} 5.3 \\
\midrule 
\multicolumn{17}{l}{Llama-3.2-3B} \\
EN & 76.5 & 4.8 & \dab{0.4} 76.1 & \da{0.7} 4.1 & 76.5 & \ua{0.4} 5.2 & \uag{0.1} 76.6 & \ua{0.3} 5.1 & \uag{0.1} 76.6 & \da{0.1} 4.7 & 76.5 & \ua{0.4} 5.2 & 76.5 & \ua{0.3} 5.1 & \dab{0.2} 76.3 & \da{4.1} 0.7 & \dab{0.4} 76.1 & \da{3.2} 1.6 \\
ES & 67.7 & 8.9 & 67.7 & \da{0.6} 8.3 & \dab{0.5} 67.2 & \da{0.6} 8.3 & \dab{0.5} 67.2 & \ua{1.0} 9.9 & \uag{0.2} 67.9 & \ua{0.7} 9.6 & 67.7 & 8.9 & \dab{0.3} 67.4 & \ua{0.4} 9.3 & \uag{0.4} 68.1 & \da{1.9} 7.0 & \uag{0.6} 68.3 & \da{1.5} 7.4 \\
IT & 66.1 & 15.6 & 66.1 & \da{0.9} 14.7 & 66.1 & \da{0.9} 14.7 & \dab{0.1} 66.0 & \da{0.4} 15.2 & \dab{0.1} 66.0 & \da{0.9} 14.7 & 66.1 & \da{0.9} 14.7 & 66.1 & \da{0.9} 14.7 & \dab{0.4} 65.7 & \ua{0.2} 15.8 & \uag{0.4} 66.5 & \da{4.1} 11.5 \\
PL & 99.6 & 0.0 & 99.6 & 0.0 & 99.6 & 0.0 & 99.6 & 0.0 & 99.6 & 0.0 & 99.6 & 0.0 & 99.6 & 0.0 & 99.6 & 0.0 & 99.6 & 0.0 \\
PT & 67.0 & 5.5 & \dab{1.0} 66.0 & \da{0.9} 4.6 & \dab{1.5} 65.5 & \ua{3.3} 8.8 & \dab{1.0} 66.0 & \ua{2.2} 7.7 & 67.0 & 5.5 & \uag{0.5} 67.5 & \ua{0.7} 6.2 & 67.0 & 5.5 & \uag{0.5} 67.5 & \da{2.7} 2.8 & \dab{1.0} 66.0 & \ua{6.0} 11.5 \\
\midrule 
\multicolumn{17}{l}{Mistral-7B-Instruct-v0.3} \\
EN & 75.2 & 6.1 & \dab{0.2} 75.0 & \da{1.2} 4.9 & 75.2 & 6.1 & 75.2 & \da{0.1} 6.0 & \uag{0.1} 75.3 & \ua{0.3} 6.4 & 75.2 & \ua{0.2} 6.3 & \uag{0.1} 75.3 & 6.1 & \uag{0.1} 75.3 & \da{2.0} 4.1 & \dab{0.4} 74.8 & \da{2.7} 3.4 \\
ES & 60.6 & 13.5 & \uag{0.9} 61.5 & \ua{0.5} 14.0 & \uag{0.4} 61.0 & \da{1.1} 12.4 & \uag{0.2} 60.8 & \da{0.4} 13.1 & \uag{0.2} 60.8 & \ua{0.5} 14.0 & \uag{0.4} 61.0 & 13.5 & \uag{1.1} 61.7 & \ua{0.3} 13.8 & \uag{0.7} 61.3 & \da{9.4} 4.1 & \uag{1.4} 62.0 & \ua{0.2} 13.7 \\
IT & 68.9 & 7.7 & \dab{0.1} 68.8 & \ua{0.5} 8.2 & 68.9 & 7.7 & 68.9 & 7.7 & \dab{0.1} 68.8 & \ua{0.1} 7.8 & \uag{0.5} 69.4 & \ua{0.1} 7.8 & \uag{0.2} 69.1 & \ua{0.8} 8.5 & \uag{0.4} 69.3 & \ua{6.1} 13.8 & \uag{0.4} 69.3 & \ua{5.8} 13.5 \\
PL & 99.6 & 0.0 & 99.6 & 0.0 & 99.6 & 0.0 & 99.6 & 0.0 & 99.6 & 0.0 & 99.6 & 0.0 & 99.6 & 0.0 & 99.6 & 0.0 & 99.6 & 0.0 \\
PT & 60.9 & 0.7 & 60.9 & 0.7 & 60.9 & \ua{1.7} 2.4 & \uag{0.5} 61.4 & \ua{2.3} 3.0 & \dab{0.5} 60.4 & \ua{0.7} 1.4 & \uag{0.5} 61.4 & \ua{1.4} 2.1 & \uag{0.5} 61.4 & \ua{1.4} 2.1 & \dab{1.0} 59.9 & \ua{3.0} 3.7 & \uag{0.5} 61.4 & \ua{3.7} 4.4 \\
\midrule 
\multicolumn{17}{l}{Mistral-7B-v0.3} \\
EN & 75.5 & 4.4 & 75.5 & \da{0.7} 3.7 & 75.5 & \da{0.1} 4.3 & \uag{0.1} 75.6 & \ua{0.4} 4.8 & \uag{0.1} 75.6 & \ua{0.2} 4.6 & \dab{0.1} 75.4 & 4.4 & \uag{0.1} 75.6 & \ua{0.2} 4.6 & \uag{0.2} 75.7 & \da{0.8} 3.6 & \dab{0.1} 75.4 & \da{0.4} 4.0 \\
ES & 64.5 & 2.7 & \uag{0.4} 64.9 & \ua{0.7} 3.4 & 64.5 & \da{0.6} 2.1 & \dab{0.3} 64.2 & \ua{0.6} 3.3 & 64.5 & 2.7 & \dab{0.3} 64.2 & \ua{0.6} 3.3 & \uag{0.6} 65.1 & \ua{0.2} 2.9 & \uag{0.2} 64.7 & \ua{4.3} 7.0 & \dab{0.3} 64.2 & \ua{1.6} 4.3 \\
IT & 67.2 & 11.0 & \dab{0.1} 67.1 & \da{0.4} 10.6 & \uag{0.2} 67.4 & \da{0.4} 10.6 & \dab{0.1} 67.1 & \da{0.4} 10.6 & \dab{0.3} 66.9 & \da{1.2} 9.8 & \uag{0.2} 67.4 & \da{0.8} 10.2 & 67.2 & 11.0 & \uag{0.3} 67.5 & \da{6.9} 4.1 & \uag{0.3} 67.5 & \da{2.7} 8.3 \\
PL & 99.6 & 0.0 & 99.6 & 0.0 & 99.6 & 0.0 & 99.6 & 0.0 & 99.6 & 0.0 & 99.6 & 0.0 & 99.6 & 0.0 & 99.6 & 0.0 & 99.6 & 0.0 \\
PT & 66.0 & 5.4 & 66.0 & 5.4 & \dab{2.0} 64.0 & \da{2.5} 2.9 & 66.0 & 5.4 & 66.0 & 5.4 & 66.0 & \da{0.3} 5.1 & 66.0 & 5.4 & \dab{2.0} 64.0 & \da{1.7} 3.7 & \dab{1.5} 64.5 & \da{1.8} 3.6 \\
\midrule 
\multicolumn{17}{l}{Mistral-Nemo-Base-2407} \\
EN & 75.8 & 4.8 & 75.8 & \da{0.6} 4.2 & \uag{0.1} 75.9 & 4.8 & \dab{0.1} 75.7 & \da{0.2} 4.6 & \uag{0.1} 75.9 & \ua{0.1} 4.9 & \uag{0.1} 75.9 & \ua{0.1} 4.9 & \dab{0.1} 75.7 & \da{0.7} 4.1 & \uag{0.2} 76.0 & \da{3.2} 1.6 & \dab{0.1} 75.7 & \da{4.3} 0.5 \\
ES & 66.1 & 10.4 & \uag{0.4} 66.5 & \ua{1.6} 12.0 & \uag{0.4} 66.5 & \ua{0.8} 11.2 & \uag{0.9} 67.0 & \da{1.5} 8.9 & 66.1 & \da{0.6} 9.8 & \dab{0.3} 65.8 & \ua{0.5} 10.9 & \uag{0.2} 66.3 & \ua{0.5} 10.9 & \uag{0.2} 66.3 & \ua{2.2} 12.6 & \uag{0.9} 67.0 & \da{0.5} 9.9 \\
IT & 69.6 & 12.7 & \uag{0.4} 70.0 & \ua{0.7} 13.4 & \uag{0.3} 69.9 & \da{0.1} 12.6 & \uag{0.3} 69.9 & \ua{0.3} 13.0 & \dab{0.2} 69.4 & \da{0.4} 12.3 & 69.6 & 12.7 & \uag{0.1} 69.7 & \ua{0.4} 13.1 & \uag{0.1} 69.7 & \da{6.9} 5.8 & \uag{0.7} 70.3 & \da{11.7} 1.0 \\
PL & 99.6 & 0.0 & 99.6 & 0.0 & 99.6 & 0.0 & 99.6 & 0.0 & 99.6 & 0.0 & 99.6 & 0.0 & 99.6 & 0.0 & 99.6 & 0.0 & 99.6 & 0.0 \\
PT & 72.6 & 3.8 & \uag{0.5} 73.1 & \ua{1.0} 4.8 & \dab{0.5} 72.1 & \da{1.5} 2.3 & \uag{0.5} 73.1 & \da{0.2} 3.6 & 72.6 & \da{1.9} 1.9 & \dab{1.5} 71.1 & 3.8 & 72.6 & 3.8 & \dab{1.5} 71.1 & \ua{11.1} 14.9 & \dab{1.5} 71.1 & \ua{10.6} 14.4 \\
\midrule 
\multicolumn{17}{l}{Mistral-Nemo-Instruct-2407} \\
EN & 76.7 & 4.6 & \dab{0.7} 76.0 & \da{0.4} 4.2 & \uag{0.3} 77.0 & \da{0.2} 4.4 & \uag{0.1} 76.8 & \ua{0.1} 4.7 & \uag{0.1} 76.8 & 4.6 & \dab{0.1} 76.6 & \da{0.1} 4.5 & \dab{0.5} 76.2 & \ua{0.3} 4.9 & \dab{0.6} 76.1 & \da{2.2} 2.4 & \dab{1.0} 75.7 & \da{3.3} 1.3 \\
ES & 70.4 & 10.2 & \dab{0.2} 70.2 & \da{2.1} 8.1 & \uag{0.2} 70.6 & \da{3.2} 7.0 & \dab{0.7} 69.7 & 10.2 & \dab{0.2} 70.2 & \ua{0.3} 10.5 & \dab{0.9} 69.5 & \ua{1.1} 11.3 & \dab{0.2} 70.2 & \da{2.1} 8.1 & \dab{0.5} 69.9 & \da{7.8} 2.4 & \dab{0.5} 69.9 & \ua{7.6} 17.8 \\
IT & 67.7 & 12.6 & \uag{0.2} 67.9 & \da{0.5} 12.1 & 67.7 & 12.6 & \uag{0.2} 67.9 & \da{0.5} 12.1 & \uag{0.2} 67.9 & \ua{0.4} 13.0 & \uag{0.2} 67.9 & \da{0.1} 12.5 & \uag{0.3} 68.0 & \ua{0.3} 12.9 & \uag{0.5} 68.2 & \ua{0.2} 12.8 & \uag{0.5} 68.2 & \da{9.0} 3.6 \\
PL & 99.6 & 0.0 & 99.6 & 0.0 & 99.6 & 0.0 & 99.6 & 0.0 & 99.6 & 0.0 & 99.6 & 0.0 & 99.6 & 0.0 & 99.6 & 0.0 & 99.6 & 0.0 \\
PT & 70.1 & 9.8 & 70.1 & 9.8 & \dab{0.6} 69.5 & \ua{3.9} 13.7 & 70.1 & \ua{2.8} 12.6 & 70.1 & \ua{2.8} 12.6 & 70.1 & \ua{3.4} 13.2 & \uag{0.5} 70.6 & \ua{2.3} 12.1 & 70.1 & \ua{4.8} 14.6 & \dab{0.6} 69.5 & \ua{4.1} 13.9 \\
\bottomrule

\end{tabular}
}

\caption{Comprehensive evaluation of Country bias mitigation on the Multilingual Hate Speech dataset across eight language models. We report hate speech prediction accuracy (Main) and True Positive Rate gap (TPR-Gap) between demographic groups for each model and debiasing approach. IMSAE (Five-Subsets) consistently achieves stronger bias reduction compared to monolingual and standard cross-lingual approaches.}
\label{Appendix:imsae-hate-speech-country}
\end{table*}

\clearpage

\begin{table*}[t]
\centering
\resizebox{\textwidth}{!}{%
\begin{tabular}{lrr rr rr rr rr rr rr rr rr}
\toprule
Target & \multicolumn{2}{c}{Baseline} & \multicolumn{2}{c}{SAL (EN)} & \multicolumn{2}{c}{SAL (ES)} & \multicolumn{2}{c}{SAL (IT)} & \multicolumn{2}{c}{SAL (PL)} & \multicolumn{2}{c}{SAL (PT)} & \multicolumn{2}{c}{\subsetswithout{IMSAE (FullyJoint)}} & \multicolumn{2}{c}{\subsetswithout{IMSAE (Subsets w/o)}} & \multicolumn{2}{c}{\subsetswith{IMSAE (Five-Subsets)}}  \\

& Main  & TPR-Gap  & Main  & TPR-Gap  & Main  & TPR-Gap  & Main  & TPR-Gap  & Main  & TPR-Gap  & Main  & TPR-Gap  & Main  & TPR-Gap  & Main  & TPR-Gap  & Main  & TPR-Gap \\

\midrule 
\multicolumn{17}{l}{mBERT-uncased} \\
EN & 86.7 & 4.4 & 86.7 & \ua{0.7} 5.1 & 86.7 & 4.4 & 86.7 & \da{0.1} 4.3 & \uag{0.1} 86.8 & \da{0.1} 4.3 & 86.7 & 4.4 & 86.7 & 4.4 & 86.7 & \da{4.3} 0.1 & \dab{0.1} 86.6 & \da{4.3} 0.1 \\
ES & 63.7 & 3.6 & \uag{0.2} 63.9 & \ua{1.0} 4.6 & \uag{0.4} 64.1 & \da{0.2} 3.4 & 63.7 & 3.6 & 63.7 & 3.6 & \dab{0.5} 63.2 & \da{0.9} 2.7 & \uag{0.9} 64.6 & \ua{0.1} 3.7 & 63.7 & \ua{2.3} 5.9 & \dab{0.3} 63.4 & \ua{0.7} 4.3 \\
IT & 68.4 & 2.1 & \dab{0.5} 67.9 & \da{1.4} 0.7 & \dab{0.5} 67.9 & \da{0.1} 2.0 & 68.4 & \da{0.6} 1.5 & \dab{0.5} 67.9 & \da{0.1} 2.0 & 68.4 & 2.1 & \dab{0.2} 68.2 & \da{1.7} 0.4 & \uag{0.3} 68.7 & \ua{1.0} 3.1 & \dab{0.5} 67.9 & \da{0.8} 1.3 \\
PL & 88.2 & 11.6 & \dab{0.1} 88.1 & 11.6 & \uag{0.2} 88.4 & 11.6 & 88.2 & \da{0.4} 11.2 & \dab{0.2} 88.0 & \da{8.8} 2.8 & 88.2 & 11.6 & 88.2 & 11.6 & \dab{0.1} 88.1 & \ua{0.4} 12.0 & \dab{0.2} 88.0 & \da{11.6} 0.0 \\
PT & 61.3 & 12.0 & \dab{0.6} 60.7 & \ua{0.7} 12.7 & \dab{1.2} 60.1 & \ua{2.4} 14.4 & \dab{0.6} 60.7 & \ua{1.2} 13.2 & \uag{0.7} 62.0 & \da{1.2} 10.8 & \uag{0.7} 62.0 & \ua{1.4} 13.4 & \dab{1.2} 60.1 & \ua{1.8} 13.8 & \uag{1.3} 62.6 & \da{5.2} 6.8 & \uag{3.1} 64.4 & \ua{2.4} 14.4 \\
\midrule 
\multicolumn{17}{l}{Llama3-8B} \\
EN & 79.4 & 4.0 & \uag{0.2} 79.6 & \ua{0.4} 4.4 & \dab{0.1} 79.3 & \ua{0.1} 4.1 & 79.4 & \ua{0.1} 4.1 & 79.4 & 4.0 & \dab{0.1} 79.3 & \ua{0.2} 4.2 & \dab{0.1} 79.3 & \da{0.1} 3.9 & \uag{0.3} 79.7 & \da{3.6} 0.4 & \uag{0.2} 79.6 & \da{2.5} 1.5 \\
ES & 70.7 & 5.5 & \uag{0.3} 71.0 & \ua{1.0} 6.5 & 70.7 & \da{0.6} 4.9 & \uag{0.5} 71.2 & \ua{0.9} 6.4 & 70.7 & \da{1.1} 4.4 & \uag{0.3} 71.0 & \ua{1.0} 6.5 & \uag{0.3} 71.0 & \ua{1.0} 6.5 & \uag{0.8} 71.5 & \ua{2.9} 8.4 & 70.7 & \da{0.3} 5.2 \\
IT & 69.4 & 3.7 & \uag{0.2} 69.6 & 3.7 & \uag{0.2} 69.6 & \da{0.8} 2.9 & \dab{0.7} 68.7 & \da{1.6} 2.1 & \uag{0.2} 69.6 & 3.7 & \uag{0.2} 69.6 & 3.7 & \uag{0.5} 69.9 & \ua{0.1} 3.8 & \uag{0.7} 70.1 & \ua{3.0} 6.7 & \dab{0.5} 68.9 & \da{2.3} 1.4 \\
PL & 88.4 & 15.3 & \dab{0.1} 88.3 & 15.3 & \dab{0.1} 88.3 & 15.3 & 88.4 & 15.3 & \dab{1.2} 87.2 & \da{2.5} 12.8 & \dab{0.3} 88.1 & \da{0.4} 14.9 & \dab{0.1} 88.3 & \da{0.4} 14.9 & \dab{0.1} 88.3 & \ua{5.7} 21.0 & \dab{1.4} 87.0 & \da{10.7} 4.6 \\
PT & 57.1 & 2.5 & 57.1 & 2.5 & \uag{1.2} 58.3 & \ua{1.5} 4.0 & \dab{0.7} 56.4 & \ua{1.9} 4.4 & 57.1 & 2.5 & \dab{0.7} 56.4 & \ua{0.2} 2.7 & \dab{1.3} 55.8 & \da{1.8} 0.7 & \dab{0.7} 56.4 & \da{1.5} 1.0 & \dab{3.7} 53.4 & \ua{0.3} 2.8 \\
\midrule 
\multicolumn{17}{l}{Llama-3.1-8B} \\
EN & 79.0 & 3.1 & \uag{0.3} 79.3 & \ua{0.6} 3.7 & 79.0 & \ua{0.2} 3.3 & \dab{0.1} 78.9 & \ua{0.1} 3.2 & 79.0 & \ua{0.1} 3.2 & 79.0 & \da{0.1} 3.0 & 79.0 & \ua{0.1} 3.2 & 79.0 & \ua{6.0} 9.1 & \uag{0.3} 79.3 & \da{2.9} 0.2 \\
ES & 72.9 & 1.4 & \dab{0.2} 72.7 & \ua{3.1} 4.5 & \uag{0.3} 73.2 & \ua{0.2} 1.6 & \dab{0.7} 72.2 & \ua{2.1} 3.5 & 72.9 & \ua{0.8} 2.2 & \dab{1.2} 71.7 & \ua{0.7} 2.1 & \dab{0.7} 72.2 & \ua{2.4} 3.8 & 72.9 & \ua{0.8} 2.2 & \uag{1.0} 73.9 & \ua{1.7} 3.1 \\
IT & 65.6 & 1.1 & \uag{0.2} 65.8 & \ua{0.1} 1.2 & \uag{0.2} 65.8 & \da{0.4} 0.7 & \uag{0.2} 65.8 & \ua{1.0} 2.1 & \uag{0.2} 65.8 & \da{0.4} 0.7 & \dab{0.3} 65.3 & 1.1 & \uag{0.2} 65.8 & \ua{0.1} 1.2 & \uag{0.7} 66.3 & \ua{3.4} 4.5 & \uag{0.7} 66.3 & \ua{1.1} 2.2 \\
PL & 87.0 & 11.6 & \dab{0.1} 86.9 & \ua{0.4} 12.0 & \dab{0.1} 86.9 & \da{0.4} 11.2 & \uag{0.1} 87.1 & \ua{0.4} 12.0 & \dab{0.3} 86.7 & \da{5.4} 6.2 & 87.0 & \ua{0.4} 12.0 & \uag{0.1} 87.1 & \ua{0.8} 12.4 & \dab{0.1} 86.9 & \ua{7.9} 19.5 & \dab{0.2} 86.8 & \da{6.6} 5.0 \\
PT & 59.5 & 2.9 & \uag{0.6} 60.1 & \da{0.8} 2.1 & \uag{0.6} 60.1 & \da{0.8} 2.1 & 59.5 & 2.9 & 59.5 & 2.9 & \dab{0.6} 58.9 & \da{0.8} 2.1 & \uag{0.6} 60.1 & \da{0.8} 2.1 & 59.5 & \da{0.9} 2.0 & \dab{1.2} 58.3 & \ua{2.7} 5.6 \\
\midrule 
\multicolumn{17}{l}{Llama-3.2-3B} \\
EN & 79.5 & 3.3 & \uag{0.1} 79.6 & \ua{0.5} 3.8 & 79.5 & \da{0.1} 3.2 & \uag{0.1} 79.6 & 3.3 & 79.5 & \ua{0.1} 3.4 & \uag{0.1} 79.6 & 3.3 & 79.5 & \da{0.1} 3.2 & 79.5 & \da{1.7} 1.6 & \uag{0.1} 79.6 & \da{2.9} 0.4 \\
ES & 68.3 & 2.4 & \dab{0.3} 68.0 & \da{0.7} 1.7 & \dab{0.5} 67.8 & \da{1.1} 1.3 & \dab{0.5} 67.8 & 2.4 & \dab{0.3} 68.0 & \ua{1.3} 3.7 & \dab{0.5} 67.8 & \ua{1.2} 3.6 & 68.3 & 2.4 & \uag{0.2} 68.5 & \da{0.4} 2.0 & \dab{0.5} 67.8 & \ua{0.3} 2.7 \\
IT & 68.2 & 4.3 & \dab{1.0} 67.2 & \da{0.7} 3.6 & \dab{0.7} 67.5 & \da{1.6} 2.7 & \dab{1.5} 66.7 & 4.3 & \dab{0.3} 67.9 & \da{1.9} 2.4 & \dab{0.3} 67.9 & \da{0.8} 3.5 & \dab{1.2} 67.0 & \da{0.9} 3.4 & \dab{1.5} 66.7 & \da{1.2} 3.1 & \dab{1.9} 66.3 & \ua{0.9} 5.2 \\
PL & 87.8 & 8.2 & \uag{0.3} 88.1 & \ua{5.8} 14.0 & \uag{0.2} 88.0 & \ua{5.8} 14.0 & \uag{0.4} 88.2 & \ua{0.4} 8.6 & \dab{0.8} 87.0 & \ua{1.2} 9.4 & \uag{0.4} 88.2 & \ua{5.8} 14.0 & \uag{0.2} 88.0 & \ua{5.8} 14.0 & \uag{0.3} 88.1 & \ua{15.1} 23.3 & \dab{0.7} 87.1 & \da{5.3} 2.9 \\
PT & 56.4 & 11.1 & \uag{0.7} 57.1 & \da{0.5} 10.6 & \uag{0.7} 57.1 & \da{0.5} 10.6 & \uag{1.3} 57.7 & \ua{0.8} 11.9 & \dab{0.6} 55.8 & \ua{0.7} 11.8 & \uag{1.9} 58.3 & \da{2.9} 8.2 & \uag{0.7} 57.1 & \da{0.5} 10.6 & 56.4 & \ua{1.0} 12.1 & \uag{1.3} 57.7 & \ua{2.3} 13.4 \\
\midrule 
\multicolumn{17}{l}{Mistral-7B-Instruct-v0.3} \\
EN & 79.9 & 2.5 & \uag{0.1} 80.0 & \ua{1.1} 3.6 & \dab{0.1} 79.8 & 2.5 & \uag{0.1} 80.0 & \ua{0.2} 2.7 & \uag{0.1} 80.0 & \ua{0.1} 2.6 & \uag{0.1} 80.0 & \ua{0.2} 2.7 & \uag{0.2} 80.1 & \ua{0.2} 2.7 & 79.9 & \da{0.3} 2.2 & 79.9 & \da{1.2} 1.3 \\
ES & 64.6 & 5.1 & \dab{0.5} 64.1 & \ua{0.7} 5.8 & 64.6 & 5.1 & \dab{0.7} 63.9 & \da{0.2} 4.9 & \dab{0.2} 64.4 & \ua{0.5} 5.6 & \dab{0.5} 64.1 & \ua{0.8} 5.9 & \dab{0.5} 64.1 & \ua{1.4} 6.5 & \dab{0.9} 63.7 & \da{2.6} 2.5 & \dab{1.7} 62.9 & \da{2.4} 2.7 \\
IT & 66.3 & 2.5 & \uag{0.4} 66.7 & \ua{0.1} 2.6 & \uag{0.2} 66.5 & \ua{0.2} 2.7 & \dab{1.0} 65.3 & \ua{1.7} 4.2 & \uag{0.2} 66.5 & \ua{0.2} 2.7 & 66.3 & \ua{3.2} 5.7 & 66.3 & \ua{2.5} 5.0 & \dab{0.3} 66.0 & \ua{4.5} 7.0 & \dab{0.5} 65.8 & \ua{2.0} 4.5 \\
PL & 87.5 & 15.8 & \uag{0.1} 87.6 & 15.8 & 87.5 & 15.8 & \uag{0.1} 87.6 & 15.8 & \uag{0.1} 87.6 & \da{0.1} 15.7 & 87.5 & \da{0.4} 15.4 & \uag{0.1} 87.6 & \ua{0.4} 16.2 & \dab{0.2} 87.3 & \da{1.1} 14.7 & \dab{0.2} 87.3 & \da{14.6} 1.2 \\
PT & 59.5 & 10.2 & 59.5 & 10.2 & 59.5 & 10.2 & 59.5 & 10.2 & 59.5 & 10.2 & 59.5 & 10.2 & \dab{0.6} 58.9 & \ua{0.1} 10.3 & \dab{0.6} 58.9 & \da{0.7} 9.5 & \dab{0.6} 58.9 & \ua{4.6} 14.8 \\
\midrule 
\multicolumn{17}{l}{Mistral-7B-v0.3} \\
EN & 79.6 & 3.6 & \uag{0.4} 80.0 & \ua{1.0} 4.6 & \dab{0.1} 79.5 & \da{0.2} 3.4 & 79.6 & \ua{0.1} 3.7 & \dab{0.1} 79.5 & \da{0.1} 3.5 & \uag{0.2} 79.8 & 3.6 & 79.6 & \ua{0.2} 3.8 & \uag{0.3} 79.9 & \da{3.1} 0.5 & \uag{0.3} 79.9 & \ua{1.3} 4.9 \\
ES & 67.1 & 6.8 & \dab{0.3} 66.8 & \ua{2.3} 9.1 & \dab{0.3} 66.8 & \ua{2.4} 9.2 & \dab{0.3} 66.8 & \ua{0.7} 7.5 & \dab{0.8} 66.3 & \ua{0.6} 7.4 & \dab{0.8} 66.3 & \da{0.1} 6.7 & \dab{0.3} 66.8 & \ua{1.6} 8.4 & 67.1 & \da{0.9} 5.9 & \dab{0.8} 66.3 & \da{0.9} 5.9 \\
IT & 65.6 & 5.7 & \uag{0.2} 65.8 & 5.7 & \dab{0.5} 65.1 & \da{0.8} 4.9 & \uag{0.7} 66.3 & 5.7 & 65.6 & 5.7 & 65.6 & 5.7 & 65.6 & \ua{0.8} 6.5 & \uag{0.2} 65.8 & \da{3.1} 2.6 & \uag{1.4} 67.0 & \ua{1.3} 7.0 \\
PL & 87.6 & 12.0 & 87.6 & \da{0.4} 11.6 & 87.6 & 12.0 & \uag{0.1} 87.7 & 12.0 & \dab{0.7} 86.9 & \ua{0.2} 12.2 & 87.6 & \da{0.1} 11.9 & 87.6 & 12.0 & 87.6 & \da{3.3} 8.7 & \dab{0.8} 86.8 & \da{9.3} 2.7 \\
PT & 57.7 & 8.3 & 57.7 & 8.3 & 57.7 & 8.3 & 57.7 & 8.3 & 57.7 & 8.3 & \uag{0.6} 58.3 & \da{0.3} 8.0 & 57.7 & 8.3 & 57.7 & \ua{1.6} 9.9 & \uag{1.2} 58.9 & \da{0.8} 7.5 \\
\midrule 
\multicolumn{17}{l}{Mistral-Nemo-Base-2407} \\
EN & 79.0 & 3.9 & \dab{0.7} 78.3 & \ua{0.2} 4.1 & \dab{0.7} 78.3 & \da{0.5} 3.4 & \uag{0.2} 79.2 & \da{0.4} 3.5 & \uag{0.3} 79.3 & \da{0.2} 3.7 & \dab{0.1} 78.9 & \da{0.4} 3.5 & \dab{0.6} 78.4 & \da{0.1} 3.8 & \dab{0.1} 78.9 & \ua{0.7} 4.6 & \dab{0.1} 78.9 & \da{3.6} 0.3 \\
ES & 72.2 & 3.1 & \dab{1.2} 71.0 & \ua{1.6} 4.7 & \uag{0.5} 72.7 & \ua{1.0} 4.1 & 72.2 & \ua{2.0} 5.1 & \dab{0.2} 72.0 & \ua{1.5} 4.6 & \dab{0.7} 71.5 & \ua{1.7} 4.8 & \dab{0.7} 71.5 & \ua{1.3} 4.4 & \dab{1.0} 71.2 & \ua{8.0} 11.1 & \dab{1.2} 71.0 & \ua{1.0} 4.1 \\
IT & 65.8 & 1.6 & 65.8 & 1.6 & 65.8 & \ua{2.3} 3.9 & \dab{0.2} 65.6 & \ua{3.4} 5.0 & \dab{0.2} 65.6 & \ua{0.7} 2.3 & 65.8 & 1.6 & \dab{0.2} 65.6 & \da{0.8} 0.8 & \dab{0.2} 65.6 & \ua{3.3} 4.9 & \uag{0.9} 66.7 & \ua{2.7} 4.3 \\
PL & 88.5 & 16.1 & 88.5 & \ua{0.4} 16.5 & 88.5 & \ua{0.4} 16.5 & \uag{0.1} 88.6 & \ua{0.4} 16.5 & \dab{0.3} 88.2 & \da{1.3} 14.8 & 88.5 & 16.1 & \uag{0.2} 88.7 & \ua{1.2} 17.3 & \uag{0.1} 88.6 & \da{10.3} 5.8 & \dab{0.3} 88.2 & \da{15.5} 0.6 \\
PT & 63.2 & 14.0 & \dab{0.6} 62.6 & \da{1.3} 12.7 & \dab{0.6} 62.6 & \da{1.3} 12.7 & \dab{0.6} 62.6 & \da{1.3} 12.7 & \uag{0.6} 63.8 & \da{1.0} 13.0 & 63.2 & \da{2.1} 11.9 & \dab{0.6} 62.6 & \da{1.3} 12.7 & \dab{1.2} 62.0 & \da{6.7} 7.3 & \uag{0.6} 63.8 & \ua{3.2} 17.2 \\
\midrule 
\multicolumn{17}{l}{Mistral-Nemo-Instruct-2407} \\
EN & 79.5 & 2.9 & \dab{0.2} 79.3 & \ua{0.6} 3.5 & \dab{0.1} 79.4 & \da{0.2} 2.7 & \dab{0.1} 79.4 & 2.9 & 79.5 & 2.9 & \dab{0.1} 79.4 & \da{0.1} 2.8 & \dab{0.1} 79.4 & \da{0.1} 2.8 & \dab{0.2} 79.3 & \da{2.3} 0.6 & \dab{0.2} 79.3 & \da{2.4} 0.5 \\
ES & 71.0 & 5.4 & 71.0 & \ua{0.9} 6.3 & 71.0 & \ua{1.3} 6.7 & \dab{0.3} 70.7 & 5.4 & 71.0 & \ua{1.3} 6.7 & \dab{0.3} 70.7 & \da{0.2} 5.2 & \dab{0.8} 70.2 & \ua{1.6} 7.0 & \dab{0.3} 70.7 & \da{2.0} 3.4 & \dab{0.3} 70.7 & \da{2.5} 2.9 \\
IT & 65.8 & 4.9 & \uag{0.2} 66.0 & \ua{0.4} 5.3 & 65.8 & 4.9 & \uag{0.2} 66.0 & \ua{1.0} 5.9 & 65.8 & 4.9 & \uag{0.5} 66.3 & \ua{0.7} 5.6 & 65.8 & 4.9 & 65.8 & \da{2.5} 2.4 & \dab{0.5} 65.3 & \da{3.5} 1.4 \\
PL & 87.6 & 10.7 & \uag{0.2} 87.8 & \ua{0.1} 10.8 & 87.6 & 10.7 & \uag{0.1} 87.7 & 10.7 & \dab{0.5} 87.1 & \ua{3.7} 14.4 & \uag{0.2} 87.8 & \ua{0.1} 10.8 & \uag{0.2} 87.8 & \ua{0.1} 10.8 & \uag{0.3} 87.9 & \da{1.6} 9.1 & \dab{0.5} 87.1 & \da{8.4} 2.3 \\
PT & 64.4 & 16.8 & \uag{0.6} 65.0 & \da{2.5} 14.3 & 64.4 & 16.8 & 64.4 & 16.8 & \uag{1.2} 65.6 & \da{1.2} 15.6 & \dab{0.6} 63.8 & \ua{0.1} 16.9 & 64.4 & \da{1.2} 15.6 & \dab{0.6} 63.8 & \da{1.8} 15.0 & \dab{1.8} 62.6 & \da{2.3} 14.5 \\
\bottomrule

\end{tabular}
}

\caption{Comprehensive evaluation of gender bias mitigation on the Multilingual Hate Speech dataset across eight language models. We report hate speech prediction accuracy (Main) and True Positive Rate gap (TPR-Gap) between demographic groups for each model and debiasing approach. IMSAE (Five-Subsets) consistently achieves stronger bias reduction compared to monolingual and standard cross-lingual approaches.}
\label{Appendix:imsae-hate-speech-gender}
\end{table*}

\clearpage

\begin{table*}[t]
\centering
\resizebox{\textwidth}{!}{%
\begin{tabular}{lrr rr rr rr rr rr rr rr rr}
\toprule
Target & \multicolumn{2}{c}{Baseline} & \multicolumn{2}{c}{SAL (EN)} & \multicolumn{2}{c}{SAL (ES)} & \multicolumn{2}{c}{SAL (IT)} & \multicolumn{2}{c}{SAL (PL)} & \multicolumn{2}{c}{SAL (PT)} & \multicolumn{2}{c}{\subsetswithout{IMSAE (FullyJoint)}} & \multicolumn{2}{c}{\subsetswithout{IMSAE (Subsets w/o)}} & \multicolumn{2}{c}{\subsetswith{IMSAE (Five-Subsets)}}  \\

& Main  & TPR-Gap  & Main  & TPR-Gap  & Main  & TPR-Gap  & Main  & TPR-Gap  & Main  & TPR-Gap  & Main  & TPR-Gap  & Main  & TPR-Gap  & Main  & TPR-Gap  & Main  & TPR-Gap \\

\midrule 
\multicolumn{17}{l}{mBERT-uncased} \\
EN & 86.8 & 4.1 & \dab{1.7} 85.1 & \da{1.3} 2.8 & 86.8 & 4.1 & 86.8 & 4.1 & 86.8 & 4.1 & \dab{0.1} 86.7 & \ua{0.3} 4.4 & \dab{0.3} 86.5 & \ua{0.8} 4.9 & \dab{0.1} 86.7 & \ua{1.9} 6.0 & \dab{2.0} 84.8 & \da{4.1} 0.0 \\
ES & 63.7 & 10.2 & \uag{0.7} 64.4 & \da{0.8} 9.4 & \uag{0.4} 64.1 & \da{0.1} 10.1 & \uag{0.7} 64.4 & 10.2 & \uag{0.2} 63.9 & 10.2 & 63.7 & 10.2 & \uag{0.4} 64.1 & \da{0.1} 10.1 & 63.7 & \da{9.2} 1.0 & \dab{0.3} 63.4 & \da{1.7} 8.5 \\
IT & 68.4 & 32.6 & \dab{0.2} 68.2 & \da{0.1} 32.5 & \dab{0.2} 68.2 & \da{0.6} 32.0 & \dab{0.9} 67.5 & \da{1.6} 31.0 & 68.4 & 32.6 & \dab{0.2} 68.2 & \da{0.6} 32.0 & \dab{0.2} 68.2 & \da{0.6} 32.0 & \dab{0.5} 67.9 & \ua{23.2} 55.8 & \dab{0.5} 67.9 & \da{25.2} 7.4 \\
PL & 91.3 & 6.2 & 91.3 & 6.2 & \uag{0.1} 91.4 & 6.2 & 91.3 & \da{0.6} 5.6 & 91.3 & \da{1.2} 5.0 & 91.3 & \da{0.6} 5.6 & 91.3 & 6.2 & 91.3 & \da{4.7} 1.5 & \dab{0.2} 91.1 & \da{6.2} 0.0 \\
PT & 61.3 & 1.0 & \uag{0.7} 62.0 & \ua{2.9} 3.9 & \dab{0.6} 60.7 & \ua{0.1} 1.1 & \dab{1.2} 60.1 & \ua{0.2} 1.2 & \dab{1.8} 59.5 & \ua{1.5} 2.5 & \dab{0.6} 60.7 & \ua{0.1} 1.1 & 61.3 & \ua{1.4} 2.4 & \uag{0.7} 62.0 & \ua{5.0} 6.0 & \uag{1.3} 62.6 & \ua{11.3} 12.3 \\
\midrule 
\multicolumn{17}{l}{Llama3-8B} \\
EN & 79.3 & 4.4 & \dab{1.7} 77.6 & \da{1.7} 2.7 & 79.3 & \ua{0.1} 4.5 & \dab{0.1} 79.2 & \ua{0.2} 4.6 & 79.3 & \ua{0.2} 4.6 & \uag{0.1} 79.4 & \ua{0.1} 4.5 & 79.3 & \ua{0.1} 4.5 & \uag{0.1} 79.4 & \da{4.3} 0.1 & \dab{1.6} 77.7 & \da{2.4} 2.0 \\
ES & 70.7 & 7.0 & \uag{0.5} 71.2 & \ua{0.9} 7.9 & 70.7 & 7.0 & \uag{0.5} 71.2 & \ua{0.8} 7.8 & \uag{0.3} 71.0 & \da{0.1} 6.9 & \uag{0.8} 71.5 & \ua{1.8} 8.8 & \uag{0.5} 71.2 & \ua{1.9} 8.9 & \dab{0.2} 70.5 & \ua{0.7} 7.7 & \uag{0.5} 71.2 & \ua{0.5} 7.5 \\
IT & 69.9 & 43.3 & \dab{0.3} 69.6 & \ua{0.1} 43.4 & \dab{0.3} 69.6 & \da{0.5} 42.8 & \dab{1.5} 68.4 & \da{1.7} 41.6 & 69.9 & 43.3 & \dab{0.3} 69.6 & \ua{0.7} 44.0 & \dab{0.5} 69.4 & \ua{0.2} 43.5 & \dab{0.5} 69.4 & \da{6.4} 36.9 & \dab{1.7} 68.2 & \da{32.7} 10.6 \\
PL & 91.0 & 16.3 & 91.0 & \da{0.6} 15.7 & \dab{0.1} 90.9 & \da{0.6} 15.7 & \uag{0.1} 91.1 & 16.3 & 91.0 & \da{0.6} 15.7 & 91.0 & \da{0.6} 15.7 & 91.0 & \da{0.6} 15.7 & 91.0 & \ua{10.5} 26.8 & \uag{0.1} 91.1 & \da{4.2} 12.1 \\
PT & 57.1 & 18.0 & 57.1 & 18.0 & 57.1 & 18.0 & \dab{0.7} 56.4 & \ua{1.3} 19.3 & 57.1 & 18.0 & 57.1 & 18.0 & \dab{0.7} 56.4 & \da{1.7} 16.3 & \dab{1.3} 55.8 & \da{4.4} 13.6 & \dab{1.3} 55.8 & \da{2.0} 16.0 \\
\midrule 
\multicolumn{17}{l}{Llama-3.1-8B} \\
EN & 79.0 & 3.1 & \dab{1.8} 77.2 & \da{2.2} 0.9 & 79.0 & \ua{0.1} 3.2 & 79.0 & \da{0.2} 2.9 & 79.0 & 3.1 & 79.0 & 3.1 & 79.0 & \ua{0.1} 3.2 & \uag{0.1} 79.1 & \ua{0.1} 3.2 & \dab{2.0} 77.0 & \da{0.6} 2.5 \\
ES & 72.9 & 1.9 & \dab{0.2} 72.7 & \da{0.2} 1.7 & \dab{0.7} 72.2 & 1.9 & \dab{0.2} 72.7 & \da{0.4} 1.5 & \dab{0.2} 72.7 & \ua{0.1} 2.0 & \dab{1.4} 71.5 & \da{0.5} 1.4 & \uag{0.5} 73.4 & \ua{0.7} 2.6 & \dab{0.2} 72.7 & \ua{2.3} 4.2 & 72.9 & \ua{2.9} 4.8 \\
IT & 66.7 & 33.0 & \uag{0.3} 67.0 & 33.0 & \uag{0.5} 67.2 & \ua{0.5} 33.5 & \uag{0.8} 67.5 & \ua{1.6} 34.6 & \uag{0.5} 67.2 & \ua{0.5} 33.5 & \uag{0.5} 67.2 & \ua{0.5} 33.5 & 66.7 & 33.0 & \uag{1.0} 67.7 & \da{12.5} 20.5 & \uag{1.0} 67.7 & \ua{1.6} 34.6 \\
PL & 90.4 & 14.5 & \dab{0.1} 90.3 & \da{0.6} 13.9 & 90.4 & 14.5 & 90.4 & 14.5 & \dab{0.1} 90.3 & \da{0.6} 13.9 & 90.4 & \ua{0.6} 15.1 & \dab{0.1} 90.3 & \da{0.6} 13.9 & 90.4 & \da{9.9} 4.6 & 90.4 & \da{9.5} 5.0 \\
PT & 59.5 & 12.1 & \uag{0.6} 60.1 & \da{1.2} 10.9 & \uag{1.2} 60.7 & \da{2.3} 9.8 & 59.5 & 12.1 & 59.5 & 12.1 & \uag{0.6} 60.1 & \da{1.2} 10.9 & \uag{0.6} 60.1 & \da{3.4} 8.7 & \uag{1.2} 60.7 & \ua{2.3} 14.4 & \dab{0.6} 58.9 & \da{4.3} 7.8 \\
\midrule 
\multicolumn{17}{l}{Llama-3.2-3B} \\
EN & 79.5 & 3.6 & \dab{2.0} 77.5 & \da{2.9} 0.7 & 79.5 & \ua{0.2} 3.8 & 79.5 & 3.6 & 79.5 & 3.6 & \uag{0.1} 79.6 & \ua{0.1} 3.7 & \dab{0.1} 79.4 & \ua{0.3} 3.9 & 79.5 & \da{3.5} 0.1 & \dab{2.0} 77.5 & \da{1.7} 1.9 \\
ES & 68.3 & 3.6 & 68.3 & 3.6 & \dab{0.3} 68.0 & \ua{0.2} 3.8 & \dab{1.0} 67.3 & \ua{0.8} 4.4 & \uag{0.2} 68.5 & \ua{1.1} 4.7 & \dab{0.3} 68.0 & \ua{0.6} 4.2 & \dab{0.5} 67.8 & \ua{0.8} 4.4 & 68.3 & \ua{0.5} 4.1 & \uag{0.5} 68.8 & \ua{1.4} 5.0 \\
IT & 68.9 & 38.3 & \dab{1.2} 67.7 & \da{1.6} 36.7 & \dab{1.0} 67.9 & \da{1.1} 37.2 & \dab{1.7} 67.2 & \da{1.1} 37.2 & \dab{0.7} 68.2 & \da{0.6} 37.7 & \dab{0.7} 68.2 & \da{0.6} 37.7 & \dab{1.2} 67.7 & \da{1.6} 36.7 & \dab{1.0} 67.9 & \da{0.4} 37.9 & \dab{1.2} 67.7 & \da{17.7} 20.6 \\
PL & 90.9 & 16.3 & 90.9 & 16.3 & \dab{0.1} 90.8 & 16.3 & \dab{0.3} 90.6 & \da{0.6} 15.7 & 90.9 & \ua{0.6} 16.9 & 90.9 & 16.3 & 90.9 & 16.3 & \dab{0.2} 90.7 & \da{10.3} 6.0 & \dab{0.1} 90.8 & \ua{19.4} 35.7 \\
PT & 56.4 & 5.4 & \uag{0.7} 57.1 & \ua{1.3} 6.7 & \uag{0.7} 57.1 & \da{4.1} 1.3 & 56.4 & 5.4 & 56.4 & \ua{1.0} 6.4 & \uag{0.7} 57.1 & \ua{0.9} 6.3 & \dab{0.6} 55.8 & \ua{1.6} 7.0 & 56.4 & \da{0.4} 5.0 & \dab{0.6} 55.8 & \ua{1.8} 7.2 \\
\midrule 
\multicolumn{17}{l}{Mistral-7B-Instruct-v0.3} \\
EN & 79.9 & 2.8 & \dab{3.1} 76.8 & \da{0.9} 1.9 & 79.9 & \ua{0.5} 3.3 & \dab{0.1} 79.8 & \ua{0.1} 2.9 & \dab{0.1} 79.8 & \ua{0.1} 2.9 & \dab{0.1} 79.8 & \ua{0.1} 2.9 & \dab{0.1} 79.8 & \ua{0.2} 3.0 & \dab{0.3} 79.6 & \ua{0.5} 3.3 & \dab{3.1} 76.8 & \da{1.1} 1.7 \\
ES & 64.6 & 8.0 & \dab{0.9} 63.7 & \da{0.9} 7.1 & 64.6 & 8.0 & \dab{0.2} 64.4 & \ua{1.7} 9.7 & \uag{0.5} 65.1 & \ua{1.4} 9.4 & \dab{0.7} 63.9 & \da{0.2} 7.8 & \dab{0.5} 64.1 & \ua{0.1} 8.1 & \uag{0.3} 64.9 & \da{1.6} 6.4 & \dab{1.7} 62.9 & \da{1.2} 6.8 \\
IT & 66.0 & 39.4 & \uag{0.3} 66.3 & \ua{0.1} 39.5 & \uag{0.3} 66.3 & \ua{0.1} 39.5 & \dab{1.4} 64.6 & \da{2.1} 37.3 & 66.0 & 39.4 & 66.0 & 39.4 & \dab{0.2} 65.8 & \da{0.4} 39.0 & \dab{0.9} 65.1 & \da{14.0} 25.4 & \dab{0.7} 65.3 & \ua{2.8} 42.2 \\
PL & 91.0 & 18.7 & \uag{0.3} 91.3 & \ua{0.6} 19.3 & 91.0 & 18.7 & 91.0 & 18.7 & 91.0 & \da{0.6} 18.1 & \dab{0.1} 90.9 & 18.7 & \uag{0.1} 91.1 & \ua{0.6} 19.3 & \uag{0.2} 91.2 & \ua{1.9} 20.6 & \dab{0.1} 90.9 & \da{8.4} 10.3 \\
PT & 59.5 & 8.8 & 59.5 & 8.8 & 59.5 & 8.8 & \dab{0.6} 58.9 & \ua{1.0} 9.8 & \dab{0.6} 58.9 & \ua{1.0} 9.8 & 59.5 & 8.8 & 59.5 & 8.8 & \dab{0.6} 58.9 & \da{5.2} 3.6 & \dab{1.2} 58.3 & \ua{4.3} 13.1 \\
\midrule 
\multicolumn{17}{l}{Mistral-7B-v0.3} \\
EN & 79.5 & 4.9 & \dab{1.6} 77.9 & \da{1.8} 3.1 & 79.5 & \da{0.1} 4.8 & 79.5 & \ua{0.1} 5.0 & 79.5 & \da{0.3} 4.6 & 79.5 & \ua{0.1} 5.0 & 79.5 & \da{0.1} 4.8 & \uag{0.1} 79.6 & \da{3.7} 1.2 & \dab{1.6} 77.9 & \da{2.0} 2.9 \\
ES & 67.1 & 3.5 & 67.1 & \da{0.1} 3.4 & 67.1 & 3.5 & \dab{0.3} 66.8 & \ua{0.3} 3.8 & \dab{0.8} 66.3 & \ua{0.5} 4.0 & \dab{0.5} 66.6 & \da{1.2} 2.3 & \dab{0.5} 66.6 & \da{0.8} 2.7 & \dab{1.2} 65.9 & \ua{0.1} 3.6 & \dab{0.3} 66.8 & \ua{5.6} 9.1 \\
IT & 66.3 & 41.1 & \dab{0.3} 66.0 & \da{0.5} 40.6 & \uag{0.2} 66.5 & \ua{0.1} 41.2 & 66.3 & \da{1.2} 39.9 & \dab{0.3} 66.0 & \da{0.1} 41.0 & \dab{0.3} 66.0 & \da{0.9} 40.2 & \uag{0.2} 66.5 & \ua{0.1} 41.2 & \dab{0.5} 65.8 & \da{22.8} 18.3 & \uag{0.7} 67.0 & \da{16.1} 25.0 \\
PL & 90.4 & 16.9 & 90.4 & \da{0.6} 16.3 & \dab{0.2} 90.2 & \da{0.6} 16.3 & \dab{0.2} 90.2 & \da{0.6} 16.3 & \uag{0.1} 90.5 & \ua{0.6} 17.5 & 90.4 & 16.9 & \dab{0.1} 90.3 & \da{0.6} 16.3 & 90.4 & \da{10.8} 6.1 & \uag{0.1} 90.5 & \da{15.1} 1.8 \\
PT & 57.7 & 2.9 & 57.7 & 2.9 & \uag{0.6} 58.3 & \ua{2.2} 5.1 & 57.7 & 2.9 & \uag{0.6} 58.3 & \ua{2.2} 5.1 & 57.7 & 2.9 & 57.7 & 2.9 & \uag{0.6} 58.3 & \ua{1.8} 4.7 & \uag{0.6} 58.3 & \da{0.4} 2.5 \\
\midrule 
\multicolumn{17}{l}{Mistral-Nemo-Base-2407} \\
EN & 79.3 & 5.3 & \dab{3.3} 76.0 & \da{1.4} 3.9 & \dab{1.0} 78.3 & \ua{0.4} 5.7 & \dab{0.8} 78.5 & \da{0.1} 5.2 & \dab{1.4} 77.9 & \ua{0.5} 5.8 & \dab{1.1} 78.2 & 5.3 & \dab{0.8} 78.5 & \da{0.4} 4.9 & \dab{0.5} 78.8 & \da{5.2} 0.1 & \dab{3.5} 75.8 & \da{4.6} 0.7 \\
ES & 72.2 & 8.9 & \dab{0.5} 71.7 & \da{1.4} 7.5 & 72.2 & \da{2.1} 6.8 & \dab{0.2} 72.0 & \da{0.9} 8.0 & \dab{0.2} 72.0 & \da{0.9} 8.0 & \dab{1.0} 71.2 & \da{0.3} 8.6 & 72.2 & \da{1.1} 7.8 & 72.2 & \da{7.3} 1.6 & \dab{0.2} 72.0 & \da{7.8} 1.1 \\
IT & 65.1 & 40.6 & \uag{0.9} 66.0 & \ua{1.0} 41.6 & \uag{0.5} 65.6 & \ua{0.5} 41.1 & \dab{0.5} 64.6 & \ua{0.1} 40.7 & 65.1 & 40.6 & \uag{0.5} 65.6 & \ua{0.5} 41.1 & \uag{0.9} 66.0 & \ua{1.0} 41.6 & \uag{0.5} 65.6 & \da{22.7} 17.9 & \uag{1.6} 66.7 & \da{22.9} 17.7 \\
PL & 91.3 & 19.9 & 91.3 & 19.9 & \dab{0.1} 91.2 & \da{0.6} 19.3 & \dab{0.1} 91.2 & \da{0.6} 19.3 & \dab{0.4} 90.9 & \da{1.8} 18.1 & 91.3 & 19.9 & \dab{0.2} 91.1 & \da{1.2} 18.7 & \dab{0.3} 91.0 & \ua{4.4} 24.3 & \dab{0.2} 91.1 & \da{16.9} 3.0 \\
PT & 63.2 & 7.6 & \dab{0.6} 62.6 & \da{1.1} 6.5 & \dab{0.6} 62.6 & \da{1.1} 6.5 & \dab{0.6} 62.6 & \da{1.1} 6.5 & \uag{0.6} 63.8 & \da{1.3} 6.3 & \uag{0.6} 63.8 & \da{1.3} 6.3 & 63.2 & \da{2.2} 5.4 & \uag{1.2} 64.4 & \da{6.3} 1.3 & \uag{0.6} 63.8 & \ua{2.0} 9.6 \\
\midrule 
\multicolumn{17}{l}{Mistral-Nemo-Instruct-2407} \\
EN & 79.5 & 5.3 & \dab{2.8} 76.7 & \da{3.3} 2.0 & \dab{0.3} 79.2 & \ua{0.3} 5.6 & \dab{0.2} 79.3 & \da{0.2} 5.1 & \dab{0.2} 79.3 & \da{0.1} 5.2 & \uag{0.2} 79.7 & 5.3 & \dab{0.3} 79.2 & \ua{0.1} 5.4 & \dab{0.2} 79.3 & \ua{1.5} 6.8 & \dab{2.6} 76.9 & \da{4.8} 0.5 \\
ES & 71.0 & 4.4 & \dab{0.5} 70.5 & \ua{0.9} 5.3 & 71.0 & 4.4 & \dab{0.5} 70.5 & 4.4 & \dab{0.5} 70.5 & \da{0.2} 4.2 & \dab{0.5} 70.5 & \da{1.1} 3.3 & \dab{0.3} 70.7 & \da{0.5} 3.9 & \dab{1.2} 69.8 & \ua{5.2} 9.6 & \dab{0.5} 70.5 & \da{0.8} 3.6 \\
IT & 65.3 & 44.8 & \dab{0.5} 64.8 & \da{0.3} 44.5 & 65.3 & \da{1.2} 43.6 & \dab{0.7} 64.6 & \da{2.0} 42.8 & 65.3 & 44.8 & 65.3 & 44.8 & \dab{0.5} 64.8 & \ua{0.3} 45.1 & \uag{0.3} 65.6 & \da{0.7} 44.1 & \uag{0.3} 65.6 & \da{43.7} 1.1 \\
PL & 90.9 & 18.1 & 90.9 & 18.1 & 90.9 & 18.1 & 90.9 & 18.1 & 90.9 & 18.1 & 90.9 & 18.1 & 90.9 & \ua{0.6} 18.7 & \dab{0.2} 90.7 & \ua{8.5} 26.6 & \uag{0.1} 91.0 & \ua{3.0} 21.1 \\
PT & 64.4 & 3.9 & \uag{0.6} 65.0 & \da{0.7} 3.2 & \dab{0.6} 63.8 & \da{1.4} 2.5 & 64.4 & \da{2.4} 1.5 & \uag{1.2} 65.6 & \ua{4.5} 8.4 & 64.4 & 3.9 & \uag{1.2} 65.6 & \da{0.3} 3.6 & \dab{0.6} 63.8 & \ua{3.5} 7.4 & \uag{0.6} 65.0 & \ua{4.2} 8.1 \\
\bottomrule

\end{tabular}
}

\caption{Comprehensive evaluation of race bias mitigation on the Multilingual Hate Speech dataset across eight language models. We report hate speech prediction accuracy (Main) and True Positive Rate gap (TPR-Gap) between demographic groups for each model and debiasing approach. IMSAE (Five-Subsets) consistently achieves stronger bias reduction compared to monolingual and standard cross-lingual approaches.}
\label{Appendix:imsae-hate-speech-race}
\end{table*}

\clearpage

\subsection{Crosslingual Baseline Performance}

This section evaluates the performance of three state-of-the-art crosslingual debiasing methods: SAL (Table \ref{appendix:sal-hate-speech}), INLP (Table \ref{appendix:inlp-hate-speech}), and SentenceDebias (Table \ref{appendix:sentence-debias-hate-speech}) on age debiasing across different source-target language pairs. Due to page limitations, we only include detailed results for age debiasing here; however, the complete results for country, gender, and race debiasing across all methods have been uploaded to our GitHub repository for reference.

\begin{table*}[b]
\centering
\resizebox{\textwidth}{!}{
\begin{tabular}{l rr rr rr rr rr rr}
\toprule
Target & \multicolumn{2}{c}{Baseline} & \multicolumn{2}{c}{SAL (EN) } & \multicolumn{2}{c}{SAL (ES)} & \multicolumn{2}{c}{SAL (IT)} & \multicolumn{2}{c}{SAL (PL)} & \multicolumn{2}{c}{SAL (PT)} \\

& Main & Ext & Main & Ext & Main & Ext & Main & Ext & Main & Ext & Main & Ext \\

\midrule 
\multicolumn{13}{l}{mBERT-uncased} \\
EN & 86.7 & 9.1 & \uag{0.3} 87.0 & \ua{0.4} 9.5 & \dab{0.5} 86.2 & \da{0.3} 8.8 & 86.7 & \ua{0.1} 9.2 & \dab{0.5} 86.2 & \da{0.1} 9.0 & 86.7 & \da{0.1} 9.0 \\
ES & 63.7 & 12.9 & \uag{0.4} 64.1 & \ua{2.1} 15.0 & \dab{0.3} 63.4 & \da{0.5} 12.4 & \uag{0.2} 63.9 & \da{0.3} 12.6 & \uag{0.2} 63.9 & \ua{0.3} 13.2 & \dab{1.0} 62.7 & \ua{0.3} 13.2 \\
IT & 68.2 & 3.6 & \dab{0.3} 67.9 & \ua{0.3} 3.9 & 68.2 & 3.6 & \dab{0.3} 67.9 & \ua{0.3} 3.9 & 68.2 & \da{0.3} 3.3 & \uag{0.2} 68.4 & \ua{0.4} 4.0 \\
PL & 91.3 & 8.8 & 91.3 & \da{0.7} 8.1 & \dab{0.1} 91.2 & \da{1.3} 7.5 & 91.3 & \da{0.7} 8.1 & \dab{0.9} 90.4 & \da{1.9} 6.9 & 91.3 & \da{0.7} 8.1 \\
PT & 61.3 & 17.6 & \dab{1.2} 60.1 & \ua{1.2} 18.8 & \dab{1.2} 60.1 & \da{0.6} 17.0 & \dab{1.2} 60.1 & \ua{1.2} 18.8 & 61.3 & 17.6 & \dab{0.6} 60.7 & \ua{1.5} 19.1 \\
\midrule 
\multicolumn{13}{l}{Llama3-8B} \\
EN & 79.6 & 7.8 & \uag{0.8} 80.4 & \ua{0.6} 8.4 & 79.6 & \da{0.4} 7.4 & \uag{0.1} 79.7 & \da{0.1} 7.7 & \uag{0.1} 79.7 & \da{0.2} 7.6 & \uag{0.3} 79.9 & \da{0.3} 7.5 \\
ES & 70.7 & 11.7 & \uag{0.3} 71.0 & \ua{0.6} 12.3 & \dab{0.2} 70.5 & \da{0.6} 11.1 & \uag{0.3} 71.0 & \ua{0.6} 12.3 & 70.7 & 11.7 & \uag{0.5} 71.2 & \ua{0.5} 12.2 \\
IT & 69.9 & 8.0 & \dab{0.3} 69.6 & \ua{0.3} 8.3 & 69.9 & 8.0 & \dab{0.5} 69.4 & \da{0.7} 7.3 & \uag{0.2} 70.1 & \da{0.9} 7.1 & \dab{0.3} 69.6 & \da{0.3} 7.7 \\
PL & 91.0 & 17.2 & 91.0 & \da{0.7} 16.5 & \dab{0.1} 90.9 & \da{0.6} 16.6 & 91.0 & \da{0.1} 17.1 & \dab{1.2} 89.8 & \da{7.8} 9.4 & 91.0 & \da{0.8} 16.4 \\
PT & 57.1 & 2.1 & \dab{0.7} 56.4 & \ua{0.9} 3.0 & \uag{1.2} 58.3 & \ua{1.9} 4.0 & \dab{0.7} 56.4 & \ua{0.9} 3.0 & 57.1 & 2.1 & 57.1 & 2.1 \\
\midrule 
\multicolumn{13}{l}{Llama-3.1-8B} \\
EN & 79.7 & 6.5 & \uag{0.7} 80.4 & \ua{0.7} 7.2 & 79.7 & \da{0.1} 6.4 & 79.7 & \ua{0.2} 6.7 & \uag{0.1} 79.8 & \ua{0.1} 6.6 & 79.7 & \da{0.2} 6.3 \\
ES & 72.9 & 8.6 & \dab{0.2} 72.7 & \ua{0.1} 8.7 & \dab{0.2} 72.7 & \da{1.0} 7.6 & 72.9 & \da{1.1} 7.5 & \dab{0.5} 72.4 & \da{0.5} 8.1 & 72.9 & \da{2.2} 6.4 \\
IT & 66.5 & 9.5 & \uag{0.2} 66.7 & \da{1.2} 8.3 & \uag{0.5} 67.0 & \da{1.0} 8.5 & \uag{0.7} 67.2 & \da{0.5} 9.0 & \uag{0.5} 67.0 & \da{1.0} 8.5 & \uag{0.5} 67.0 & \da{0.1} 9.4 \\
PL & 90.4 & 15.2 & \dab{0.1} 90.3 & \da{0.6} 14.6 & 90.4 & 15.2 & 90.4 & 15.2 & \dab{0.4} 90.0 & \da{6.9} 8.3 & \dab{0.1} 90.3 & \da{0.6} 14.6 \\
PT & 59.5 & 2.8 & \uag{0.6} 60.1 & \ua{0.5} 3.3 & \uag{0.6} 60.1 & \da{1.4} 1.4 & 59.5 & 2.8 & 59.5 & 2.8 & \uag{0.6} 60.1 & \ua{0.5} 3.3 \\
\midrule 
\multicolumn{13}{l}{Llama-3.2-3B} \\
EN & 79.7 & 6.2 & \uag{0.6} 80.3 & \ua{0.8} 7.0 & \uag{0.1} 79.8 & \ua{0.3} 6.5 & 79.7 & 6.2 & \uag{0.1} 79.8 & \ua{0.1} 6.3 & \uag{0.1} 79.8 & \ua{0.3} 6.5 \\
ES & 68.3 & 12.1 & \dab{0.3} 68.0 & \da{1.2} 10.9 & 68.3 & \ua{1.0} 13.1 & 68.3 & 12.1 & \uag{0.2} 68.5 & \ua{0.9} 13.0 & \dab{0.5} 67.8 & \ua{0.9} 13.0 \\
IT & 67.7 & 5.0 & 67.7 & \ua{0.3} 5.3 & \uag{0.5} 68.2 & \ua{2.0} 7.0 & \dab{0.2} 67.5 & \ua{0.7} 5.7 & \uag{0.2} 67.9 & \ua{1.8} 6.8 & \uag{0.2} 67.9 & \ua{1.5} 6.5 \\
PL & 90.9 & 17.1 & \dab{0.2} 90.7 & \ua{0.1} 17.2 & \dab{0.2} 90.7 & \da{0.6} 16.5 & \dab{0.3} 90.6 & \da{0.6} 16.5 & \dab{0.9} 90.0 & \da{1.4} 15.7 & 90.9 & 17.1 \\
PT & 56.4 & 8.9 & \uag{0.7} 57.1 & \ua{2.8} 11.7 & \uag{0.7} 57.1 & \ua{0.8} 9.7 & \uag{0.7} 57.1 & \ua{0.8} 9.7 & \dab{1.2} 55.2 & \ua{1.5} 10.4 & 56.4 & \ua{1.7} 10.6 \\
\midrule 
\multicolumn{13}{l}{Mistral-7B-Instruct-v0.3} \\
EN & 79.4 & 7.1 & \uag{1.1} 80.5 & \ua{1.3} 8.4 & \uag{0.2} 79.6 & \ua{0.1} 7.2 & \uag{0.3} 79.7 & \ua{0.1} 7.2 & \uag{0.1} 79.5 & \ua{0.3} 7.4 & \uag{0.2} 79.6 & \da{0.1} 7.0 \\
ES & 64.6 & 7.2 & \dab{0.7} 63.9 & \da{0.2} 7.0 & \uag{0.8} 65.4 & \ua{1.6} 8.8 & \dab{0.2} 64.4 & \da{0.8} 6.4 & \dab{0.5} 64.1 & \ua{0.5} 7.7 & \dab{0.5} 64.1 & \da{1.5} 5.7 \\
IT & 66.0 & 3.5 & 66.0 & 3.5 & \dab{0.2} 65.8 & \da{0.2} 3.3 & \uag{0.3} 66.3 & \da{0.3} 3.2 & \uag{0.5} 66.5 & \ua{0.1} 3.6 & \dab{0.2} 65.8 & \da{0.2} 3.3 \\
PL & 91.0 & 19.5 & 91.0 & 19.5 & \uag{0.2} 91.2 & \ua{1.2} 20.7 & \dab{0.1} 90.9 & \da{0.6} 18.9 & \dab{0.8} 90.2 & \da{3.8} 15.7 & \uag{0.2} 91.2 & \ua{1.3} 20.8 \\
PT & 59.5 & 17.2 & 59.5 & 17.2 & \dab{0.6} 58.9 & \da{1.0} 16.2 & \dab{0.6} 58.9 & \da{1.0} 16.2 & 59.5 & 17.2 & \dab{0.6} 58.9 & \ua{1.8} 19.0 \\
\midrule 
\multicolumn{13}{l}{Mistral-7B-v0.3} \\
EN & 79.8 & 7.1 & \uag{0.6} 80.4 & \ua{0.4} 7.5 & \dab{0.1} 79.7 & 7.1 & 79.8 & \da{0.3} 6.8 & \dab{0.1} 79.7 & \ua{0.1} 7.2 & \dab{0.2} 79.6 & \da{0.3} 6.8 \\
ES & 67.1 & 12.8 & 67.1 & \da{1.4} 11.4 & \dab{0.3} 66.8 & \da{0.8} 12.0 & \dab{0.3} 66.8 & 12.8 & \dab{0.3} 66.8 & \ua{1.8} 14.6 & 67.1 & \ua{0.2} 13.0 \\
IT & 65.8 & 5.6 & 65.8 & \da{0.9} 4.7 & \dab{0.2} 65.6 & \ua{0.5} 6.1 & \uag{0.2} 66.0 & \da{0.6} 5.0 & 65.8 & \da{0.1} 5.5 & 65.8 & \da{0.8} 4.8 \\
PL & 90.4 & 17.8 & \uag{0.3} 90.7 & \ua{0.4} 18.2 & 90.4 & 17.8 & 90.4 & \da{0.8} 17.0 & \dab{0.9} 89.5 & \da{5.8} 12.0 & \dab{0.1} 90.3 & \ua{0.1} 17.9 \\
PT & 57.7 & 12.4 & 57.7 & 12.4 & 57.7 & 12.4 & 57.7 & 12.4 & 57.7 & 12.4 & 57.7 & 12.4 \\
\midrule 
\multicolumn{13}{l}{Mistral-Nemo-Base-2407} \\
EN & 78.9 & 6.4 & \uag{0.4} 79.3 & \ua{1.1} 7.5 & \dab{0.3} 78.6 & \ua{0.1} 6.5 & \dab{0.5} 78.4 & \ua{0.2} 6.6 & 78.9 & \da{0.1} 6.3 & \dab{0.3} 78.6 & \da{0.3} 6.1 \\
ES & 72.2 & 16.3 & \dab{1.2} 71.0 & \ua{1.3} 17.6 & \uag{0.7} 72.9 & \ua{1.8} 18.1 & 72.2 & \ua{2.0} 18.3 & \uag{0.2} 72.4 & \ua{0.6} 16.9 & \dab{0.5} 71.7 & \da{1.0} 15.3 \\
IT & 64.8 & 5.8 & \uag{0.5} 65.3 & \ua{1.1} 6.9 & \uag{0.3} 65.1 & \da{1.4} 4.4 & \dab{0.4} 64.4 & 5.8 & \dab{0.2} 64.6 & \ua{0.5} 6.3 & \uag{0.5} 65.3 & \ua{0.1} 5.9 \\
PL & 91.3 & 20.9 & \dab{0.3} 91.0 & \da{0.7} 20.2 & \dab{0.1} 91.2 & \da{0.6} 20.3 & \dab{0.1} 91.2 & \da{0.6} 20.3 & \dab{1.0} 90.3 & \da{3.4} 17.5 & \dab{0.1} 91.2 & 20.9 \\
PT & 63.2 & 8.8 & 63.2 & 8.8 & \dab{0.6} 62.6 & \da{1.6} 7.2 & 63.2 & \da{3.0} 5.8 & \uag{0.6} 63.8 & \da{1.4} 7.4 & 63.2 & 8.8 \\
\midrule 
\multicolumn{13}{l}{Mistral-Nemo-Instruct-2407} \\
EN & 79.3 & 5.9 & \uag{0.3} 79.6 & \ua{1.1} 7.0 & \dab{0.2} 79.1 & \ua{0.1} 6.0 & \dab{0.2} 79.1 & \ua{0.2} 6.1 & \dab{0.2} 79.1 & \ua{0.1} 6.0 & \dab{0.3} 79.0 & \ua{0.5} 6.4 \\
ES & 71.0 & 10.2 & \dab{0.5} 70.5 & \ua{1.2} 11.4 & \dab{0.8} 70.2 & \ua{1.2} 11.4 & \dab{0.5} 70.5 & \ua{0.9} 11.1 & \dab{1.0} 70.0 & \da{0.6} 9.6 & \dab{0.3} 70.7 & \ua{1.6} 11.8 \\
IT & 65.3 & 8.0 & \dab{0.2} 65.1 & \da{0.6} 7.4 & \dab{0.2} 65.1 & \da{0.1} 7.9 & 65.3 & \da{0.4} 7.6 & 65.3 & 8.0 & 65.3 & \ua{0.5} 8.5 \\
PL & 90.9 & 18.9 & \dab{0.1} 90.8 & \ua{0.1} 19.0 & 90.9 & 18.9 & 90.9 & 18.9 & \dab{0.7} 90.2 & \da{1.9} 17.0 & 90.9 & 18.9 \\
PT & 64.4 & 1.5 & 64.4 & 1.5 & 64.4 & 1.5 & \dab{0.6} 63.8 & \ua{1.8} 3.3 & \uag{1.2} 65.6 & \ua{2.1} 3.6 & 64.4 & 1.5 \\
\bottomrule

\end{tabular}
}
\caption{Performance comparison of SAL age debiasing approaches on the Multilingual Hate Speech dataset, showing effectiveness across different source-target language pairs. These baseline methods show limited cross-lingual transfer compared to IMSAE's multilingual approach.}
\label{appendix:sal-hate-speech}
\end{table*}

\clearpage

\begin{table*}
\centering
\scalebox{0.65}{
\begin{tabular}{l rr rr rr rr rr rr}
\toprule
Target & \multicolumn{2}{c}{Baseline} & \multicolumn{2}{c}{EN - INLP} & \multicolumn{2}{c}{ES - INLP} & \multicolumn{2}{c}{IT - INLP} & \multicolumn{2}{c}{PL - INLP} & \multicolumn{2}{c}{PT - INLP} \\

& Main & Ext & Main & Ext & Main & Ext & Main & Ext & Main & Ext & Main & Ext \\

\midrule 
\multicolumn{13}{l}{mBERT-uncased} \\
EN & 86.7 & 9.1 & \uag{0.3} 87.0 & \ua{0.7} 9.8 & \uag{0.1} 86.8 & \da{0.1} 9.0 & 86.7 & \ua{0.1} 9.2 & 86.7 & 9.1 & \uag{0.1} 86.8 & \da{0.2} 8.9 \\
ES & 63.7 & 12.9 & \uag{0.4} 64.1 & \da{0.5} 12.4 & 63.7 & \ua{1.0} 13.9 & \uag{0.4} 64.1 & \ua{0.1} 13.0 & \uag{0.2} 63.9 & \da{0.3} 12.6 & 63.7 & \ua{0.6} 13.5 \\
IT & 68.2 & 3.6 & 68.2 & 3.6 & \uag{0.2} 68.4 & \da{0.2} 3.4 & \uag{0.2} 68.4 & \da{0.2} 3.4 & 68.2 & 3.6 & \uag{0.2} 68.4 & \ua{0.9} 4.5 \\
PL & 91.3 & 8.8 & 91.3 & \da{0.7} 8.1 & 91.3 & 8.8 & 91.3 & 8.8 & \dab{0.6} 90.7 & \da{1.3} 7.5 & 91.3 & 8.8 \\
PT & 61.3 & 17.6 & \uag{0.7} 62.0 & \da{1.5} 16.1 & \dab{0.6} 60.7 & \da{0.4} 17.2 & \dab{1.2} 60.1 & \ua{1.2} 18.8 & 61.3 & 17.6 & 61.3 & \ua{2.0} 19.6 \\
\midrule 
\multicolumn{13}{l}{Llama3-8B} \\
EN & 79.6 & 7.8 & \dab{0.4} 79.2 & \da{0.6} 7.2 & 79.6 & \da{0.2} 7.6 & 79.6 & \da{0.2} 7.6 & \uag{0.2} 79.8 & \ua{0.1} 7.9 & \uag{0.1} 79.7 & \da{0.2} 7.6 \\
ES & 70.7 & 11.7 & 70.7 & 11.7 & 70.7 & 11.7 & 70.7 & 11.7 & 70.7 & 11.7 & 70.7 & 11.7 \\
IT & 69.9 & 8.0 & 69.9 & 8.0 & 69.9 & 8.0 & \dab{0.5} 69.4 & \ua{1.2} 9.2 & \dab{0.3} 69.6 & \ua{1.0} 9.0 & 69.9 & 8.0 \\
PL & 91.0 & 17.2 & \dab{0.1} 90.9 & \da{0.6} 16.6 & \dab{0.1} 90.9 & \da{0.6} 16.6 & \dab{0.1} 90.9 & \da{0.6} 16.6 & \uag{0.1} 91.1 & \ua{0.6} 17.8 & 91.0 & 17.2 \\
PT & 57.1 & 2.1 & \dab{0.7} 56.4 & \ua{0.9} 3.0 & 57.1 & 2.1 & 57.1 & 2.1 & 57.1 & 2.1 & \uag{3.0} 60.1 & \ua{7.1} 9.2 \\
\midrule 
\multicolumn{13}{l}{Llama-3.1-8B} \\
EN & 79.7 & 6.5 & \dab{0.6} 79.1 & \da{0.1} 6.4 & \uag{0.1} 79.8 & 6.5 & \dab{0.1} 79.6 & 6.5 & 79.7 & \da{0.2} 6.3 & 79.7 & 6.5 \\
ES & 72.9 & 8.6 & \dab{0.5} 72.4 & \da{0.3} 8.3 & 72.9 & 8.6 & 72.9 & \da{2.3} 6.3 & \dab{0.9} 72.0 & \ua{0.3} 8.9 & 72.9 & \da{1.2} 7.4 \\
IT & 66.5 & 9.5 & 66.5 & 9.5 & 66.5 & 9.5 & \uag{0.5} 67.0 & \da{0.1} 9.4 & 66.5 & 9.5 & 66.5 & 9.5 \\
PL & 90.4 & 15.2 & 90.4 & 15.2 & \dab{0.1} 90.3 & \da{0.6} 14.6 & 90.4 & 15.2 & 90.4 & 15.2 & \dab{0.1} 90.3 & \da{0.6} 14.6 \\
PT & 59.5 & 2.8 & 59.5 & 2.8 & 59.5 & 2.8 & \uag{0.6} 60.1 & \ua{0.5} 3.3 & 59.5 & 2.8 & 59.5 & \ua{1.7} 4.5 \\
\midrule 
\multicolumn{13}{l}{Llama-3.2-3B} \\
EN & 79.7 & 6.2 & \uag{0.1} 79.8 & \ua{0.1} 6.3 & 79.7 & \ua{0.2} 6.4 & 79.7 & \ua{0.1} 6.3 & 79.7 & \ua{0.2} 6.4 & 79.7 & \ua{0.2} 6.4 \\
ES & 68.3 & 12.1 & 68.3 & 12.1 & 68.3 & 12.1 & 68.3 & 12.1 & \dab{0.3} 68.0 & \da{0.8} 11.3 & \dab{0.3} 68.0 & \da{0.8} 11.3 \\
IT & 67.7 & 5.0 & \uag{0.5} 68.2 & \ua{0.2} 5.2 & 67.7 & 5.0 & 67.7 & \ua{0.7} 5.7 & \uag{0.5} 68.2 & \ua{1.3} 6.3 & \uag{0.7} 68.4 & \ua{1.5} 6.5 \\
PL & 90.9 & 17.1 & \dab{0.2} 90.7 & \ua{0.1} 17.2 & 90.9 & 17.1 & \dab{0.1} 90.8 & 17.1 & \dab{0.2} 90.7 & \ua{1.4} 18.5 & \dab{0.2} 90.7 & \da{0.6} 16.5 \\
PT & 56.4 & 8.9 & \uag{0.7} 57.1 & \ua{0.8} 9.7 & \uag{1.3} 57.7 & \ua{2.0} 10.9 & 56.4 & \ua{1.9} 10.8 & \uag{0.7} 57.1 & \ua{0.8} 9.7 & \uag{1.3} 57.7 & \ua{3.9} 12.8 \\
\midrule 
\multicolumn{13}{l}{Mistral-7B-Instruct-v0.3} \\
EN & 79.4 & 7.1 & \uag{0.7} 80.1 & \da{0.1} 7.0 & 79.4 & \ua{0.1} 7.2 & 79.4 & \ua{0.1} 7.2 & \uag{0.2} 79.6 & \da{0.6} 6.5 & \uag{0.2} 79.6 & \da{0.4} 6.7 \\
ES & 64.6 & 7.2 & 64.6 & 7.2 & 64.6 & \da{1.2} 6.0 & 64.6 & 7.2 & \uag{0.3} 64.9 & \da{0.5} 6.7 & \uag{0.3} 64.9 & \da{0.5} 6.7 \\
IT & 66.0 & 3.5 & 66.0 & 3.5 & 66.0 & 3.5 & \dab{0.2} 65.8 & \da{0.5} 3.0 & \dab{0.7} 65.3 & \da{0.4} 3.1 & \dab{0.4} 65.6 & \da{0.7} 2.8 \\
PL & 91.0 & 19.5 & 91.0 & 19.5 & 91.0 & 19.5 & \dab{0.1} 90.9 & 19.5 & \dab{0.1} 90.9 & 19.5 & \dab{0.1} 90.9 & 19.5 \\
PT & 59.5 & 17.2 & \dab{0.6} 58.9 & \da{1.0} 16.2 & 59.5 & 17.2 & 59.5 & 17.2 & 59.5 & 17.2 & \dab{2.4} 57.1 & \da{4.2} 13.0 \\
\midrule 
\multicolumn{13}{l}{Mistral-7B-v0.3} \\
EN & 79.8 & 7.1 & \uag{0.3} 80.1 & \da{0.2} 6.9 & \dab{0.1} 79.7 & \da{0.2} 6.9 & 79.8 & 7.1 & 79.8 & \da{0.1} 7.0 & 79.8 & \da{0.1} 7.0 \\
ES & 67.1 & 12.8 & 67.1 & 12.8 & 67.1 & 12.8 & \dab{0.3} 66.8 & \da{0.8} 12.0 & \uag{0.2} 67.3 & \da{0.7} 12.1 & \dab{0.5} 66.6 & \da{0.8} 12.0 \\
IT & 65.8 & 5.6 & 65.8 & \da{0.1} 5.5 & \dab{0.2} 65.6 & \ua{0.4} 6.0 & \uag{0.5} 66.3 & \da{0.2} 5.4 & 65.8 & \da{0.1} 5.5 & \uag{0.2} 66.0 & \da{1.3} 4.3 \\
PL & 90.4 & 17.8 & 90.4 & 17.8 & 90.4 & 17.8 & \dab{0.2} 90.2 & \da{0.5} 17.3 & \uag{0.2} 90.6 & \da{0.1} 17.7 & 90.4 & 17.8 \\
PT & 57.7 & 12.4 & 57.7 & 12.4 & 57.7 & 12.4 & 57.7 & 12.4 & 57.7 & 12.4 & \dab{1.3} 56.4 & \ua{0.7} 13.1 \\
\midrule 
\multicolumn{13}{l}{Mistral-Nemo-Base-2407} \\
EN & 78.9 & 6.4 & \uag{0.2} 79.1 & \ua{0.1} 6.5 & 78.9 & 6.4 & \uag{0.3} 79.2 & \da{0.1} 6.3 & \dab{0.1} 78.8 & \ua{0.3} 6.7 & 78.9 & \da{0.3} 6.1 \\
ES & 72.2 & 16.3 & \dab{0.5} 71.7 & \ua{1.0} 17.3 & 72.2 & \ua{2.1} 18.4 & \uag{0.2} 72.4 & \ua{0.6} 16.9 & 72.2 & \ua{1.1} 17.4 & \dab{0.2} 72.0 & \ua{0.5} 16.8 \\
IT & 64.8 & 5.8 & 64.8 & 5.8 & \dab{0.2} 64.6 & \ua{0.5} 6.3 & \uag{0.5} 65.3 & \ua{0.1} 5.9 & 64.8 & 5.8 & \dab{0.2} 64.6 & \ua{0.5} 6.3 \\
PL & 91.3 & 20.9 & \dab{0.1} 91.2 & \da{0.6} 20.3 & 91.3 & 20.9 & \dab{0.1} 91.2 & \da{0.6} 20.3 & \dab{0.2} 91.1 & \da{1.3} 19.6 & \dab{0.1} 91.2 & \da{0.6} 20.3 \\
PT & 63.2 & 8.8 & 63.2 & 8.8 & 63.2 & 8.8 & 63.2 & 8.8 & 63.2 & 8.8 & 63.2 & \da{3.0} 5.8 \\
\midrule 
\multicolumn{13}{l}{Mistral-Nemo-Instruct-2407} \\
EN & 79.3 & 5.9 & \dab{0.3} 79.0 & \da{0.3} 5.6 & \dab{0.1} 79.2 & \ua{0.3} 6.2 & \dab{0.2} 79.1 & \ua{0.3} 6.2 & 79.3 & \ua{0.5} 6.4 & \dab{0.2} 79.1 & \da{0.1} 5.8 \\
ES & 71.0 & 10.2 & \dab{0.3} 70.7 & \ua{0.3} 10.5 & \dab{0.3} 70.7 & \ua{0.3} 10.5 & \dab{0.3} 70.7 & \ua{0.6} 10.8 & \dab{0.3} 70.7 & \ua{0.3} 10.5 & \dab{0.3} 70.7 & \ua{0.6} 10.8 \\
IT & 65.3 & 8.0 & 65.3 & 8.0 & 65.3 & 8.0 & 65.3 & 8.0 & 65.3 & 8.0 & \dab{0.2} 65.1 & \da{0.1} 7.9 \\
PL & 90.9 & 18.9 & 90.9 & 18.9 & 90.9 & 18.9 & 90.9 & 18.9 & 90.9 & 18.9 & 90.9 & 18.9 \\
PT & 64.4 & 1.5 & \dab{1.2} 63.2 & \ua{3.6} 5.1 & \uag{0.6} 65.0 & \ua{0.9} 2.4 & 64.4 & 1.5 & \dab{0.6} 63.8 & \ua{4.0} 5.5 & \uag{0.6} 65.0 & \ua{1.0} 2.5 \\
\bottomrule

\end{tabular}
}
\caption{Performance comparison of INLP age debiasing approaches on the Multilingual Hate Speech dataset, showing effectiveness across different source-target language pairs. These baseline methods show limited cross-lingual transfer compared to IMSAE's multilingual approach.}
\label{appendix:inlp-hate-speech}
\end{table*}

\clearpage

\begin{table*}
\centering
\scalebox{0.65}{
\begin{tabular}{l rr rr rr rr rr rr}
\toprule
Target & \multicolumn{2}{c}{Baseline} & \multicolumn{2}{c}{EN - SentenceDebias} & \multicolumn{2}{c}{ES - SentenceDebias} & \multicolumn{2}{c}{IT - SentenceDebias} & \multicolumn{2}{c}{PL - SentenceDebias} & \multicolumn{2}{c}{PT - SentenceDebias} \\

& Main & Ext & Main & Ext & Main & Ext & Main & Ext & Main & Ext & Main & Ext \\

\midrule 
\multicolumn{13}{l}{mBERT-uncased} \\
EN & 86.7 & 9.1 & \uag{0.3} 87.0 & \ua{0.9} 10.0 & 86.7 & 9.1 & 86.7 & 9.1 & 86.7 & 9.1 & \uag{0.1} 86.8 & 9.1 \\
ES & 63.7 & 12.9 & \uag{0.2} 63.9 & \ua{1.4} 14.3 & \uag{1.4} 65.1 & \ua{1.8} 14.7 & \uag{0.2} 63.9 & \ua{0.3} 13.2 & 63.7 & \ua{0.6} 13.5 & 63.7 & \ua{0.6} 13.5 \\
IT & 68.2 & 3.6 & \dab{0.3} 67.9 & \ua{0.3} 3.9 & \dab{0.3} 67.9 & \da{0.7} 2.9 & \uag{0.2} 68.4 & \ua{0.9} 4.5 & 68.2 & \da{1.0} 2.6 & 68.2 & 3.6 \\
PL & 91.3 & 8.8 & \dab{0.1} 91.2 & \da{1.3} 7.5 & \dab{0.1} 91.2 & \da{0.7} 8.1 & 91.3 & 8.8 & \dab{0.2} 91.1 & \da{0.7} 8.1 & 91.3 & 8.8 \\
PT & 61.3 & 17.6 & \dab{0.6} 60.7 & \da{0.4} 17.2 & \dab{0.6} 60.7 & \da{0.4} 17.2 & \uag{0.7} 62.0 & \ua{1.8} 19.4 & \dab{0.6} 60.7 & \da{0.4} 17.2 & \uag{1.3} 62.6 & \ua{2.2} 19.8 \\
\midrule 
\multicolumn{13}{l}{Llama3-8B} \\
EN & 79.6 & 7.8 & \uag{0.5} 80.1 & \ua{0.6} 8.4 & 79.6 & \da{0.2} 7.6 & \uag{0.1} 79.7 & \da{0.3} 7.5 & 79.6 & 7.8 & 79.6 & \ua{0.1} 7.9 \\
ES & 70.7 & 11.7 & 70.7 & 11.7 & 70.7 & 11.7 & 70.7 & 11.7 & 70.7 & 11.7 & 70.7 & 11.7 \\
IT & 69.9 & 8.0 & \dab{0.3} 69.6 & \ua{0.3} 8.3 & \dab{0.3} 69.6 & \ua{0.3} 8.3 & \dab{0.5} 69.4 & 8.0 & 69.9 & 8.0 & \dab{0.3} 69.6 & \ua{0.3} 8.3 \\
PL & 91.0 & 17.2 & 91.0 & \da{0.7} 16.5 & \dab{0.1} 90.9 & \da{0.6} 16.6 & \dab{0.1} 90.9 & 17.2 & \dab{0.2} 90.8 & \ua{0.6} 17.8 & 91.0 & \da{0.7} 16.5 \\
PT & 57.1 & 2.1 & 57.1 & 2.1 & \uag{0.6} 57.7 & \ua{0.8} 2.9 & 57.1 & 2.1 & 57.1 & 2.1 & \uag{0.6} 57.7 & \ua{5.9} 8.0 \\
\midrule 
\multicolumn{13}{l}{Llama-3.1-8B} \\
EN & 79.7 & 6.5 & \uag{0.4} 80.1 & \ua{1.3} 7.8 & \dab{0.1} 79.6 & \da{0.3} 6.2 & 79.7 & 6.5 & 79.7 & \ua{0.1} 6.6 & \uag{0.1} 79.8 & \ua{0.1} 6.6 \\
ES & 72.9 & 8.6 & \dab{0.2} 72.7 & \da{1.9} 6.7 & \dab{0.2} 72.7 & \da{0.9} 7.7 & \uag{0.3} 73.2 & \da{0.5} 8.1 & \dab{0.5} 72.4 & \da{0.3} 8.3 & 72.9 & 8.6 \\
IT & 66.5 & 9.5 & \uag{0.5} 67.0 & \da{1.0} 8.5 & \uag{0.2} 66.7 & \da{0.5} 9.0 & \uag{0.5} 67.0 & \da{0.6} 8.9 & 66.5 & 9.5 & \uag{0.2} 66.7 & \da{1.3} 8.2 \\
PL & 90.4 & 15.2 & 90.4 & 15.2 & 90.4 & 15.2 & 90.4 & 15.2 & 90.4 & \ua{0.7} 15.9 & \dab{0.1} 90.3 & \da{0.6} 14.6 \\
PT & 59.5 & 2.8 & \uag{0.6} 60.1 & \ua{0.5} 3.3 & 59.5 & 2.8 & 59.5 & 2.8 & \uag{0.6} 60.1 & \ua{0.5} 3.3 & \dab{2.4} 57.1 & \ua{0.3} 3.1 \\
\midrule 
\multicolumn{13}{l}{Llama-3.2-3B} \\
EN & 79.7 & 6.2 & \uag{0.1} 79.8 & \ua{1.2} 7.4 & 79.7 & \ua{0.2} 6.4 & \dab{0.1} 79.6 & \ua{0.2} 6.4 & 79.7 & \ua{0.2} 6.4 & \uag{0.1} 79.8 & \ua{0.2} 6.4 \\
ES & 68.3 & 12.1 & 68.3 & 12.1 & \dab{0.7} 67.6 & \da{0.3} 11.8 & \dab{0.3} 68.0 & \da{0.2} 11.9 & \dab{0.3} 68.0 & \ua{0.2} 12.3 & \dab{0.5} 67.8 & \da{0.1} 12.0 \\
IT & 67.7 & 5.0 & 67.7 & \ua{1.2} 6.2 & \uag{0.2} 67.9 & \ua{0.8} 5.8 & \dab{0.2} 67.5 & \ua{2.3} 7.3 & \uag{0.5} 68.2 & \ua{0.2} 5.2 & \uag{0.2} 67.9 & \ua{0.5} 5.5 \\
PL & 90.9 & 17.1 & 90.9 & 17.1 & \dab{0.2} 90.7 & \ua{0.1} 17.2 & \dab{0.1} 90.8 & 17.1 & \dab{0.5} 90.4 & \ua{1.2} 18.3 & 90.9 & 17.1 \\
PT & 56.4 & 8.9 & \uag{1.3} 57.7 & \ua{2.0} 10.9 & \uag{1.3} 57.7 & \ua{4.0} 12.9 & \dab{0.6} 55.8 & \ua{2.7} 11.6 & \uag{0.7} 57.1 & \ua{2.8} 11.7 & \uag{0.7} 57.1 & \ua{1.2} 10.1 \\
\midrule 
\multicolumn{13}{l}{Mistral-7B-Instruct-v0.3} \\
EN & 79.4 & 7.1 & \uag{0.6} 80.0 & \ua{0.5} 7.6 & 79.4 & \ua{0.2} 7.3 & \uag{0.2} 79.6 & \da{0.4} 6.7 & \uag{0.2} 79.6 & \da{0.4} 6.7 & \uag{0.6} 80.0 & \da{0.1} 7.0 \\
ES & 64.6 & 7.2 & \dab{0.5} 64.1 & \ua{1.0} 8.2 & 64.6 & \ua{2.4} 9.6 & \uag{0.3} 64.9 & \da{0.5} 6.7 & \uag{0.3} 64.9 & \da{0.5} 6.7 & \uag{1.0} 65.6 & \ua{0.6} 7.8 \\
IT & 66.0 & 3.5 & 66.0 & 3.5 & \dab{0.4} 65.6 & \da{0.7} 2.8 & \dab{0.7} 65.3 & \da{0.1} 3.4 & \uag{0.5} 66.5 & \ua{0.1} 3.6 & 66.0 & \da{0.6} 2.9 \\
PL & 91.0 & 19.5 & 91.0 & 19.5 & \uag{0.2} 91.2 & \ua{0.6} 20.1 & \uag{0.1} 91.1 & 19.5 & \uag{0.3} 91.3 & 19.5 & 91.0 & 19.5 \\
PT & 59.5 & 17.2 & 59.5 & 17.2 & 59.5 & 17.2 & \dab{0.6} 58.9 & \da{1.0} 16.2 & 59.5 & 17.2 & \dab{3.1} 56.4 & \da{6.3} 10.9 \\
\midrule 
\multicolumn{13}{l}{Mistral-7B-v0.3} \\
EN & 79.8 & 7.1 & \uag{0.5} 80.3 & \da{0.4} 6.7 & 79.8 & \da{0.1} 7.0 & 79.8 & \da{0.1} 7.0 & 79.8 & \ua{0.2} 7.3 & \dab{0.1} 79.7 & \da{0.2} 6.9 \\
ES & 67.1 & 12.8 & \uag{0.2} 67.3 & \da{1.7} 11.1 & \dab{0.5} 66.6 & \da{2.7} 10.1 & \dab{0.3} 66.8 & \da{0.8} 12.0 & \dab{0.5} 66.6 & \da{1.7} 11.1 & 67.1 & 12.8 \\
IT & 65.8 & 5.6 & \uag{0.2} 66.0 & \ua{0.4} 6.0 & 65.8 & \da{0.1} 5.5 & \uag{0.5} 66.3 & 5.6 & \uag{0.5} 66.3 & \da{1.0} 4.6 & 65.8 & \da{0.1} 5.5 \\
PL & 90.4 & 17.8 & 90.4 & 17.8 & 90.4 & 17.8 & 90.4 & 17.8 & 90.4 & \ua{0.1} 17.9 & \uag{0.1} 90.5 & \da{0.1} 17.7 \\
PT & 57.7 & 12.4 & 57.7 & 12.4 & 57.7 & 12.4 & 57.7 & 12.4 & \uag{0.6} 58.3 & \ua{0.8} 13.2 & 57.7 & \da{2.1} 10.3 \\
\midrule 
\multicolumn{13}{l}{Mistral-Nemo-Base-2407} \\
EN & 78.9 & 6.4 & \uag{0.7} 79.6 & \ua{1.0} 7.4 & \dab{0.5} 78.4 & \ua{0.1} 6.5 & \dab{0.1} 78.8 & \da{0.1} 6.3 & 78.9 & 6.4 & \dab{0.2} 78.7 & \ua{0.1} 6.5 \\
ES & 72.2 & 16.3 & 72.2 & \ua{1.1} 17.4 & \dab{0.2} 72.0 & \ua{1.4} 17.7 & 72.2 & \ua{1.1} 17.4 & \dab{0.5} 71.7 & \ua{3.2} 19.5 & \uag{0.5} 72.7 & \ua{1.2} 17.5 \\
IT & 64.8 & 5.8 & 64.8 & 5.8 & 64.8 & 5.8 & \uag{1.2} 66.0 & \ua{2.5} 8.3 & 64.8 & 5.8 & \uag{0.3} 65.1 & \da{0.4} 5.4 \\
PL & 91.3 & 20.9 & \dab{0.3} 91.0 & \da{0.7} 20.2 & 91.3 & 20.9 & 91.3 & 20.9 & \dab{0.4} 90.9 & \da{1.9} 19.0 & 91.3 & 20.9 \\
PT & 63.2 & 8.8 & \dab{0.6} 62.6 & \da{1.6} 7.2 & \dab{0.6} 62.6 & \da{1.6} 7.2 & 63.2 & 8.8 & 63.2 & \da{3.0} 5.8 & \uag{1.2} 64.4 & \da{4.1} 4.7 \\
\midrule 
\multicolumn{13}{l}{Mistral-Nemo-Instruct-2407} \\
EN & 79.3 & 5.9 & \uag{0.1} 79.4 & \ua{1.0} 6.9 & 79.3 & \ua{0.3} 6.2 & \dab{0.3} 79.0 & \ua{0.1} 6.0 & \uag{0.1} 79.4 & \ua{0.1} 6.0 & \dab{0.2} 79.1 & \ua{0.4} 6.3 \\
ES & 71.0 & 10.2 & \uag{0.2} 71.2 & \ua{0.4} 10.6 & \dab{1.2} 69.8 & \ua{0.4} 10.6 & \dab{0.5} 70.5 & \ua{0.9} 11.1 & \dab{0.5} 70.5 & \ua{0.9} 11.1 & 71.0 & \ua{0.4} 10.6 \\
IT & 65.3 & 8.0 & \uag{0.5} 65.8 & \ua{0.1} 8.1 & \dab{0.2} 65.1 & \da{0.1} 7.9 & \uag{0.5} 65.8 & \ua{0.3} 8.3 & 65.3 & 8.0 & \uag{0.3} 65.6 & \ua{0.5} 8.5 \\
PL & 90.9 & 18.9 & 90.9 & 18.9 & 90.9 & 18.9 & 90.9 & 18.9 & \dab{0.5} 90.4 & \da{0.6} 18.3 & 90.9 & 18.9 \\
PT & 64.4 & 1.5 & \dab{1.2} 63.2 & \ua{3.6} 5.1 & \uag{0.6} 65.0 & \ua{1.0} 2.5 & \dab{1.2} 63.2 & \ua{1.9} 3.4 & \uag{0.6} 65.0 & \ua{1.0} 2.5 & \dab{1.2} 63.2 & \ua{6.4} 7.9 \\
\bottomrule

\end{tabular}
}
\caption{Performance comparison of SentenceDebias age debiasing approaches on the Multilingual Hate Speech dataset, showing effectiveness across different source-target language pairs. These baseline methods show limited cross-lingual transfer compared to IMSAE's multilingual approach.}
\label{appendix:sentence-debias-hate-speech}
\end{table*}

\clearpage

\section{MSEFair}

This appendix provides additional results and analysis for our Multilingual Stack Exchange Fairness (MSEFair) dataset. Table \ref{appendix:imsae-sefair} presents the full reputation bias mitigation results on the MSEFair dataset across English, Portuguese and Russian, comparing Baseline, Crosslingual, IMSAE on Two-Subsets-Without (excluding target language) and IMSAE Three-Subsets (using all languages). Table \ref{appendix:crosslingual-sefair} shows a comprehensive comparison of crosslingual reputation debiasing approaches on MSEFair across different source-target language pairs.

\begin{table*}[!b]
\centering
\resizebox{\textwidth}{!}{
\begin{tabular}{lrr  rr  rr  rr  rr  rr rr}
\toprule
Target & \multicolumn{2}{c}{Baseline} & \multicolumn{2}{c}{SAL (EN)} & \multicolumn{2}{c}{SAL (PT)} & \multicolumn{2}{c}{IMSAE (RU)} & \multicolumn{2}{c}{\subsetswithout{IMSAE (FullyJoint)}} & \multicolumn{2}{c}{\subsetswithout{IMSAE (Subsets w/o)}} & \multicolumn{2}{c}{\subsetswith{IMSAE (Three-Subsets)}} \\

& Main & Ext & Main & Ext & Main & Ext & Main & Ext & Main & Ext & Main & Ext & Main & Ext \\

\midrule 
\multicolumn{15}{l}{mBERT-uncased} \\
EN & 67.5 & 10.7 & \dab{4.3} 63.2 & \da{9.4} 1.3 & \dab{0.1} 67.4 & \ua{0.3} 11.0 & \dab{0.3} 67.2 & \ua{0.2} 10.9 & \dab{0.3} 67.2 & 10.7 & \dab{4.2} 63.3 & \da{9.3} 1.4 & \dab{4.4} 63.1 & \da{9.5} 1.2 \\
PT & 78.3 & 16.2 & \uag{0.1} 78.4 & \da{0.1} 16.1 & \dab{19.6} 58.7 & \da{14.8} 1.4 & \dab{0.1} 78.2 & 16.2 & \dab{0.1} 78.2 & \da{0.8} 15.4 & \dab{19.7} 58.6 & \da{14.5} 1.7 & \dab{19.8} 58.5 & \da{15.6} 0.6 \\
RU & 70.0 & 18.0 & \dab{0.4} 69.6 & \da{0.1} 17.9 & \dab{0.2} 69.8 & \ua{0.2} 18.2 & \dab{11.9} 58.1 & \da{15.1} 2.9 & \dab{0.2} 69.8 & \da{0.3} 17.7 & \dab{0.6} 69.4 & \da{0.1} 17.9 & \dab{12.3} 57.7 & \da{14.7} 3.3 \\
\midrule 
\multicolumn{15}{l}{Llama3-8B} \\
EN & 68.1 & 11.7 & \dab{4.4} 63.7 & \da{6.6} 5.1 & \uag{0.1} 68.2 & 11.7 & 68.1 & \da{0.1} 11.6 & \dab{0.1} 68.0 & \da{0.4} 11.3 & \dab{4.3} 63.8 & \da{6.9} 4.8 & \dab{4.5} 63.6 & \da{6.9} 4.8 \\
PT & 82.6 & 21.1 & \dab{0.3} 82.3 & \da{0.1} 21.0 & \dab{18.0} 64.6 & \da{11.4} 9.7 & 82.6 & \ua{0.4} 21.5 & \dab{0.4} 82.2 & \da{0.3} 20.8 & \dab{17.9} 64.7 & \da{11.5} 9.6 & \dab{17.8} 64.8 & \da{11.2} 9.9 \\
RU & 71.6 & 17.6 & \dab{0.1} 71.5 & 17.6 & \dab{0.1} 71.5 & \da{0.2} 17.4 & \dab{8.0} 63.6 & \da{9.8} 7.8 & 71.6 & \ua{0.2} 17.8 & \dab{0.2} 71.4 & \da{0.2} 17.4 & \dab{8.1} 63.5 & \da{10.0} 7.6 \\
\midrule 
\multicolumn{15}{l}{Llama-3.1-8B} \\
EN & 68.4 & 11.2 & \dab{4.3} 64.1 & \da{6.0} 5.2 & 68.4 & \da{0.2} 11.0 & 68.4 & \ua{0.1} 11.3 & \dab{0.1} 68.3 & \ua{0.2} 11.4 & \dab{4.3} 64.1 & \da{6.1} 5.1 & \dab{4.5} 63.9 & \da{5.9} 5.3 \\
PT & 82.6 & 22.4 & \dab{0.2} 82.4 & \da{0.4} 22.0 & \dab{18.2} 64.4 & \da{13.6} 8.8 & \dab{0.3} 82.3 & \ua{0.1} 22.5 & \dab{0.7} 81.9 & \da{0.6} 21.8 & \dab{18.1} 64.5 & \da{13.1} 9.3 & \dab{18.3} 64.3 & \da{13.6} 8.8 \\
RU & 72.1 & 16.4 & \dab{0.1} 72.0 & \da{0.2} 16.2 & \dab{0.2} 71.9 & 16.4 & \dab{8.4} 63.7 & \da{10.4} 6.0 & 72.1 & \ua{0.2} 16.6 & \dab{0.1} 72.0 & \da{0.3} 16.1 & \dab{8.7} 63.4 & \da{10.6} 5.8 \\
\midrule 
\multicolumn{15}{l}{Llama-3.2-3B} \\
EN & 66.4 & 8.9 & \dab{4.1} 62.3 & \da{4.6} 4.3 & \uag{0.1} 66.5 & \da{0.2} 8.7 & \uag{0.1} 66.5 & \ua{0.2} 9.1 & 66.4 & \ua{0.2} 9.1 & \dab{4.1} 62.3 & \da{4.8} 4.1 & \dab{4.0} 62.4 & \da{4.5} 4.4 \\
PT & 82.1 & 17.7 & 82.1 & \ua{0.3} 18.0 & \dab{19.4} 62.7 & \da{11.7} 6.0 & \dab{0.1} 82.0 & 17.7 & \uag{0.1} 82.2 & \ua{0.1} 17.8 & \dab{19.7} 62.4 & \da{12.5} 5.2 & \dab{19.6} 62.5 & \da{11.5} 6.2 \\
RU & 71.7 & 17.1 & 71.7 & 17.1 & \dab{0.2} 71.5 & \ua{0.1} 17.2 & \dab{10.7} 61.0 & \da{11.9} 5.2 & \dab{0.3} 71.4 & \ua{0.1} 17.2 & \dab{0.2} 71.5 & \ua{0.1} 17.2 & \dab{10.5} 61.2 & \da{12.6} 4.5 \\
\midrule 
\multicolumn{15}{l}{Mistral-7B-Instruct-v0.3} \\
EN & 66.8 & 9.5 & \dab{3.7} 63.1 & \da{5.8} 3.7 & 66.8 & \da{0.2} 9.3 & \dab{0.1} 66.7 & \da{0.3} 9.2 & \dab{0.1} 66.7 & \da{0.5} 9.0 & \dab{3.6} 63.2 & \da{6.3} 3.2 & \dab{3.8} 63.0 & \da{6.4} 3.1 \\
PT & 81.4 & 20.9 & \dab{0.3} 81.1 & \ua{0.3} 21.2 & \dab{18.2} 63.2 & \da{9.5} 11.4 & \dab{0.2} 81.2 & \ua{0.5} 21.4 & \dab{0.1} 81.3 & \da{0.5} 20.4 & \dab{18.1} 63.3 & \da{9.7} 11.2 & \dab{17.9} 63.5 & \da{8.2} 12.7 \\
RU & 70.5 & 14.7 & 70.5 & \da{0.8} 13.9 & \dab{0.2} 70.3 & \ua{0.1} 14.8 & \dab{8.2} 62.3 & \da{9.3} 5.4 & 70.5 & \da{0.3} 14.4 & \dab{0.1} 70.4 & \da{0.9} 13.8 & \dab{8.4} 62.1 & \da{9.4} 5.3 \\
\midrule 
\multicolumn{15}{l}{Mistral-7B-v0.3} \\
EN & 67.5 & 10.6 & \dab{4.2} 63.3 & \da{4.7} 5.9 & 67.5 & \da{0.1} 10.5 & 67.5 & \da{0.2} 10.4 & \dab{0.1} 67.4 & 10.6 & \dab{4.1} 63.4 & \da{5.0} 5.6 & \dab{4.1} 63.4 & \da{4.9} 5.7 \\
PT & 82.4 & 20.1 & \dab{0.1} 82.3 & 20.1 & \dab{18.9} 63.5 & \da{10.5} 9.6 & \dab{0.2} 82.2 & \ua{0.1} 20.2 & 82.4 & \da{0.3} 19.8 & \dab{18.8} 63.6 & \da{10.8} 9.3 & \dab{19.0} 63.4 & \da{11.0} 9.1 \\
RU & 71.2 & 16.2 & \dab{0.2} 71.0 & \da{0.6} 15.6 & \dab{0.1} 71.1 & \da{0.2} 16.0 & \dab{9.7} 61.5 & \da{9.6} 6.6 & \dab{0.3} 70.9 & 16.2 & \dab{0.2} 71.0 & \da{0.7} 15.5 & \dab{9.6} 61.6 & \da{10.1} 6.1 \\
\midrule 
\multicolumn{15}{l}{Mistral-Nemo-Base-2407} \\
EN & 68.4 & 11.0 & \dab{4.2} 64.2 & \da{5.6} 5.4 & \dab{0.2} 68.2 & \da{0.2} 10.8 & \dab{0.1} 68.3 & \ua{0.3} 11.3 & \dab{0.2} 68.2 & \da{0.2} 10.8 & \dab{4.3} 64.1 & \da{5.6} 5.4 & \dab{4.1} 64.3 & \da{5.7} 5.3 \\
PT & 83.1 & 21.4 & \dab{0.2} 82.9 & \da{0.4} 21.0 & \dab{18.2} 64.9 & \da{14.8} 6.6 & \dab{0.2} 82.9 & \ua{0.2} 21.6 & \dab{0.5} 82.6 & \da{0.3} 21.1 & \dab{18.3} 64.8 & \da{15.0} 6.4 & \dab{18.4} 64.7 & \da{14.9} 6.5 \\
RU & 72.3 & 16.6 & \dab{0.2} 72.1 & \da{0.1} 16.5 & \dab{0.3} 72.0 & \ua{0.2} 16.8 & \dab{10.8} 61.5 & \da{11.3} 5.3 & \dab{0.2} 72.1 & \ua{0.1} 16.7 & \dab{0.2} 72.1 & \ua{0.4} 17.0 & \dab{11.0} 61.3 & \da{11.2} 5.4 \\
\midrule 
\multicolumn{15}{l}{Mistral-Nemo-Instruct-2407} \\
EN & 67.9 & 10.4 & \dab{3.6} 64.3 & \da{5.4} 5.0 & \dab{0.2} 67.7 & \da{0.2} 10.2 & 67.9 & \da{0.2} 10.2 & \dab{0.3} 67.6 & \da{0.7} 9.7 & \dab{3.7} 64.2 & \da{5.4} 5.0 & \dab{3.7} 64.2 & \da{5.6} 4.8 \\
PT & 82.4 & 20.8 & \uag{0.1} 82.5 & \ua{0.6} 21.4 & \dab{18.5} 63.9 & \da{16.6} 4.2 & \dab{0.1} 82.3 & \ua{0.4} 21.2 & \dab{0.3} 82.1 & 20.8 & \dab{18.6} 63.8 & \da{16.5} 4.3 & \dab{18.8} 63.6 & \da{16.0} 4.8 \\
RU & 72.0 & 15.6 & \dab{0.2} 71.8 & \da{0.1} 15.5 & \dab{0.1} 71.9 & \da{0.5} 15.1 & \dab{9.1} 62.9 & \da{11.1} 4.5 & \dab{0.1} 71.9 & \da{0.9} 14.7 & \dab{0.2} 71.8 & \da{0.2} 15.4 & \dab{9.1} 62.9 & \da{11.1} 4.5 \\
\bottomrule

\end{tabular}
}

\caption{Full reputation bias mitigation results on the MSEFair dataset across English, Portuguese, and Russian, comparing Baseline, Crosslingual, IMSAE on FullyJoint, IMSAE without target language, and IMSAE with all languages. We report helpfulness prediction accuracy (Main) and True Positive Rate gap (TPR-Gap) between low and high reputation users. Results demonstrate IMSAE's effectiveness in reducing bias while maintaining acceptable task performance.}
\label{appendix:imsae-sefair}
\end{table*}

\clearpage

\begin{table*}[t]
\centering
\resizebox{\textwidth}{!}{
\begin{tabular}{l rr rr rr  rr rr rr  rr rr rr  rr}
\toprule
Target & \multicolumn{2}{c}{Baseline} & \multicolumn{2}{c}{SAL (EN)} & \multicolumn{2}{c}{SAL (PT)} & \multicolumn{2}{c}{SAL (RU)} & \multicolumn{2}{c}{ INLP (EN)} & \multicolumn{2}{c}{INLP (PT)} & \multicolumn{2}{c}{INLP (RU)} & \multicolumn{2}{c}{SentenceDebias (EN)} & \multicolumn{2}{c}{SentenceDebias (PT)} & \multicolumn{2}{c}{SentenceDebias (RU)} \\

& Main & Ext & Main & Ext & Main & Ext & Main & Ext & Main & Ext & Main & Ext & Main & Ext & Main & Ext & Main & Ext & Main & Ext \\

\midrule
\multicolumn{21}{l}{mBERT-uncased} \\
EN & 67.5 & 10.7 & \dab{4.3} 63.2 & \da{9.4} 1.3 & \dab{0.1} 67.4 & \ua{0.3} 11.0 & \dab{0.3} 67.2 & \ua{0.2} 10.9 & \dab{0.4} 67.1 & \da{0.8} 9.9 & \dab{0.3} 67.2 & \da{0.2} 10.5 & \dab{0.2} 67.3 & \da{0.2} 10.5 & \dab{2.9} 64.6 & \ua{1.0} 11.7 & \dab{0.2} 67.3 & \da{0.1} 10.6 & \dab{0.1} 67.4 & \da{0.1} 10.6 \\
PT & 78.3 & 16.2 & \uag{0.1} 78.4 & \da{0.1} 16.1 & \dab{19.6} 58.7 & \da{14.8} 1.4 & \dab{0.1} 78.2 & 16.2 & 78.3 & \da{0.1} 16.1 & \dab{0.7} 77.6 & \da{0.6} 15.6 & \dab{0.1} 78.2 & 16.2 & \dab{0.1} 78.2 & \da{0.2} 16.0 & \dab{1.5} 76.8 & \da{1.4} 14.8 & 78.3 & 16.2 \\
RU & 70.0 & 18.0 & \dab{0.4} 69.6 & \da{0.1} 17.9 & \dab{0.2} 69.8 & \ua{0.2} 18.2 & \dab{11.9} 58.1 & \da{15.1} 2.9 & \dab{0.1} 69.9 & \da{0.7} 17.3 & \dab{0.1} 69.9 & \da{0.4} 17.6 & \dab{1.0} 69.0 & \da{1.9} 16.1 & 70.0 & \da{0.7} 17.3 & 70.0 & \ua{0.1} 18.1 & \uag{0.1} 70.1 & \da{0.1} 17.9 \\
\multicolumn{21}{l}{Llama3-8B} \\
EN & 68.1 & 11.7 & \dab{4.4} 63.7 & \da{6.6} 5.1 & \uag{0.1} 68.2 & 11.7 & 68.1 & \da{0.1} 11.6 & \dab{0.1} 68.0 & \da{0.4} 11.3 & \uag{0.1} 68.2 & \ua{0.1} 11.8 & 68.1 & 11.7 & \dab{1.4} 66.7 & \da{1.0} 10.7 & 68.1 & 11.7 & 68.1 & 11.7 \\
PT & 82.6 & 21.1 & \dab{0.3} 82.3 & \da{0.1} 21.0 & \dab{18.0} 64.6 & \da{11.4} 9.7 & 82.6 & \ua{0.4} 21.5 & 82.6 & 21.1 & \dab{0.3} 82.3 & \da{1.0} 20.1 & 82.6 & 21.1 & \dab{0.1} 82.5 & \da{0.2} 20.9 & \uag{0.3} 82.9 & \ua{0.3} 21.4 & 82.6 & 21.1 \\
RU & 71.6 & 17.6 & \dab{0.1} 71.5 & 17.6 & \dab{0.1} 71.5 & \da{0.2} 17.4 & \dab{8.0} 63.6 & \da{9.8} 7.8 & \uag{0.1} 71.7 & \ua{0.1} 17.7 & \uag{0.1} 71.7 & \ua{0.1} 17.7 & \dab{0.4} 71.2 & \da{0.5} 17.1 & 71.6 & \ua{0.4} 18.0 & 71.6 & \ua{0.2} 17.8 & \dab{0.2} 71.4 & \da{0.4} 17.2 \\
\multicolumn{21}{l}{Llama-3.1-8B} \\
EN & 68.4 & 11.2 & \dab{4.3} 64.1 & \da{6.0} 5.2 & 68.4 & \da{0.2} 11.0 & 68.4 & \ua{0.1} 11.3 & 68.4 & \ua{0.7} 11.9 & \dab{0.1} 68.3 & \da{0.1} 11.1 & 68.4 & 11.2 & \dab{1.2} 67.2 & \ua{0.1} 11.3 & 68.4 & 11.2 & 68.4 & \ua{0.1} 11.3 \\
PT & 82.6 & 22.4 & \dab{0.2} 82.4 & \da{0.4} 22.0 & \dab{18.2} 64.4 & \da{13.6} 8.8 & \dab{0.3} 82.3 & \ua{0.1} 22.5 & \uag{0.1} 82.7 & \ua{0.4} 22.8 & \dab{1.5} 81.1 & \da{1.5} 20.9 & 82.6 & \ua{0.5} 22.9 & \dab{0.1} 82.5 & \da{0.3} 22.1 & \uag{0.1} 82.7 & \ua{0.5} 22.9 & \uag{0.1} 82.7 & \ua{0.2} 22.6 \\
RU & 72.1 & 16.4 & \dab{0.1} 72.0 & \da{0.2} 16.2 & \dab{0.2} 71.9 & 16.4 & \dab{8.4} 63.7 & \da{10.4} 6.0 & 72.1 & \ua{0.2} 16.6 & \dab{0.1} 72.0 & 16.4 & \dab{0.3} 71.8 & \da{1.1} 15.3 & 72.1 & \da{0.3} 16.1 & 72.1 & \da{0.2} 16.2 & \dab{0.1} 72.0 & \da{0.1} 16.3 \\
\multicolumn{21}{l}{Llama-3.2-3B} \\
EN & 66.4 & 8.9 & \dab{4.1} 62.3 & \da{4.6} 4.3 & \uag{0.1} 66.5 & \da{0.2} 8.7 & \uag{0.1} 66.5 & \ua{0.2} 9.1 & \dab{0.1} 66.3 & \ua{0.7} 9.6 & 66.4 & 8.9 & 66.4 & \ua{0.1} 9.0 & \dab{0.6} 65.8 & \da{1.0} 7.9 & 66.4 & 8.9 & 66.4 & \ua{0.1} 9.0 \\
PT & 82.1 & 17.7 & 82.1 & \ua{0.3} 18.0 & \dab{19.4} 62.7 & \da{11.7} 6.0 & \dab{0.1} 82.0 & 17.7 & 82.1 & \ua{0.2} 17.9 & \dab{0.5} 81.6 & \ua{0.3} 18.0 & \dab{0.1} 82.0 & \ua{0.2} 17.9 & 82.1 & \ua{0.1} 17.8 & \dab{0.1} 82.0 & \ua{0.7} 18.4 & 82.1 & \ua{0.1} 17.8 \\
RU & 71.7 & 17.1 & 71.7 & 17.1 & \dab{0.2} 71.5 & \ua{0.1} 17.2 & \dab{10.7} 61.0 & \da{11.9} 5.2 & \dab{0.1} 71.6 & \da{0.1} 17.0 & 71.7 & \da{0.1} 17.0 & \dab{1.1} 70.6 & \da{1.8} 15.3 & \dab{0.1} 71.6 & \da{0.3} 16.8 & \dab{0.1} 71.6 & \da{0.2} 16.9 & \dab{0.4} 71.3 & \da{0.1} 17.0 \\
\multicolumn{21}{l}{Mistral-7B-Instruct-v0.3} \\
EN & 66.8 & 9.5 & \dab{3.7} 63.1 & \da{5.8} 3.7 & 66.8 & \da{0.2} 9.3 & \dab{0.1} 66.7 & \da{0.3} 9.2 & \uag{0.2} 67.0 & \da{0.4} 9.1 & \dab{0.1} 66.7 & \da{0.4} 9.1 & 66.8 & \da{0.1} 9.4 & \dab{0.9} 65.9 & \da{1.1} 8.4 & 66.8 & \da{0.1} 9.4 & 66.8 & 9.5 \\
PT & 81.4 & 20.9 & \dab{0.3} 81.1 & \ua{0.3} 21.2 & \dab{18.2} 63.2 & \da{9.5} 11.4 & \dab{0.2} 81.2 & \ua{0.5} 21.4 & \uag{0.1} 81.5 & \ua{0.1} 21.0 & \dab{0.7} 80.7 & \ua{0.2} 21.1 & 81.4 & \ua{0.2} 21.1 & 81.4 & \da{0.1} 20.8 & 81.4 & 20.9 & 81.4 & 20.9 \\
RU & 70.5 & 14.7 & 70.5 & \da{0.8} 13.9 & \dab{0.2} 70.3 & \ua{0.1} 14.8 & \dab{8.2} 62.3 & \da{9.3} 5.4 & \uag{0.1} 70.6 & \ua{0.1} 14.8 & \uag{0.1} 70.6 & \da{0.2} 14.5 & \dab{0.5} 70.0 & \da{0.6} 14.1 & \uag{0.1} 70.6 & 14.7 & \uag{0.1} 70.6 & \da{0.1} 14.6 & \uag{0.1} 70.6 & \da{0.2} 14.5 \\
\multicolumn{21}{l}{Mistral-7B-v0.3} \\
EN & 67.5 & 10.6 & \dab{4.2} 63.3 & \da{4.7} 5.9 & 67.5 & \da{0.1} 10.5 & 67.5 & \da{0.2} 10.4 & \uag{0.1} 67.6 & \da{0.2} 10.4 & 67.5 & \ua{0.2} 10.8 & 67.5 & \ua{0.1} 10.7 & \dab{1.4} 66.1 & \da{0.6} 10.0 & 67.5 & 10.6 & 67.5 & \ua{0.1} 10.7 \\
PT & 82.4 & 20.1 & \dab{0.1} 82.3 & 20.1 & \dab{18.9} 63.5 & \da{10.5} 9.6 & \dab{0.2} 82.2 & \ua{0.1} 20.2 & \uag{0.1} 82.5 & 20.1 & \dab{0.8} 81.6 & \ua{0.7} 20.8 & \uag{0.1} 82.5 & 20.1 & \uag{0.1} 82.5 & \ua{0.1} 20.2 & 82.4 & \da{0.3} 19.8 & 82.4 & \ua{0.1} 20.2 \\
RU & 71.2 & 16.2 & \dab{0.2} 71.0 & \da{0.6} 15.6 & \dab{0.1} 71.1 & \da{0.2} 16.0 & \dab{9.7} 61.5 & \da{9.6} 6.6 & \uag{0.1} 71.3 & \da{0.3} 15.9 & \uag{0.1} 71.3 & \ua{0.1} 16.3 & \dab{0.2} 71.0 & \da{0.7} 15.5 & \uag{0.1} 71.3 & \ua{0.2} 16.4 & 71.2 & 16.2 & \uag{0.1} 71.3 & \da{0.2} 16.0 \\
\multicolumn{21}{l}{Mistral-Nemo-Base-2407} \\
EN & 68.4 & 11.0 & \dab{4.2} 64.2 & \da{5.6} 5.4 & \dab{0.2} 68.2 & \da{0.2} 10.8 & \dab{0.1} 68.3 & \ua{0.3} 11.3 & \dab{0.2} 68.2 & \ua{0.2} 11.2 & 68.4 & \ua{0.4} 11.4 & 68.4 & \ua{0.2} 11.2 & \dab{0.2} 68.2 & \ua{0.7} 11.7 & 68.4 & \ua{0.3} 11.3 & \dab{0.1} 68.3 & \ua{0.4} 11.4 \\
PT & 83.1 & 21.4 & \dab{0.2} 82.9 & \da{0.4} 21.0 & \dab{18.2} 64.9 & \da{14.8} 6.6 & \dab{0.2} 82.9 & \ua{0.2} 21.6 & \dab{0.1} 83.0 & \ua{0.1} 21.5 & \dab{1.0} 82.1 & \da{1.2} 20.2 & 83.1 & 21.4 & \dab{0.1} 83.0 & 21.4 & 83.1 & \ua{0.4} 21.8 & \dab{0.1} 83.0 & \ua{0.2} 21.6 \\
RU & 72.3 & 16.6 & \dab{0.2} 72.1 & \da{0.1} 16.5 & \dab{0.3} 72.0 & \ua{0.2} 16.8 & \dab{10.8} 61.5 & \da{11.3} 5.3 & 72.3 & 16.6 & \dab{0.1} 72.2 & \da{0.1} 16.5 & \dab{0.3} 72.0 & \ua{0.3} 16.9 & 72.3 & \ua{0.1} 16.7 & 72.3 & 16.6 & \dab{0.3} 72.0 & \ua{0.2} 16.8 \\
\multicolumn{21}{l}{Mistral-Nemo-Instruct-2407} \\
EN & 67.9 & 10.4 & \dab{3.6} 64.3 & \da{5.4} 5.0 & \dab{0.2} 67.7 & \da{0.2} 10.2 & 67.9 & \da{0.2} 10.2 & \dab{0.1} 67.8 & \da{0.2} 10.2 & \dab{0.1} 67.8 & \da{0.1} 10.3 & 67.9 & \ua{0.1} 10.5 & \dab{0.8} 67.1 & \da{1.4} 9.0 & \dab{0.2} 67.7 & \da{0.1} 10.3 & \dab{0.1} 67.8 & 10.4 \\
PT & 82.4 & 20.8 & \uag{0.1} 82.5 & \ua{0.6} 21.4 & \dab{18.5} 63.9 & \da{16.6} 4.2 & \dab{0.1} 82.3 & \ua{0.4} 21.2 & 82.4 & \ua{0.3} 21.1 & \dab{1.0} 81.4 & \da{1.6} 19.2 & 82.4 & \ua{0.4} 21.2 & 82.4 & \ua{0.2} 21.0 & \uag{0.1} 82.5 & \ua{1.0} 21.8 & \uag{0.2} 82.6 & \ua{0.3} 21.1 \\
RU & 72.0 & 15.6 & \dab{0.2} 71.8 & \da{0.1} 15.5 & \dab{0.1} 71.9 & \da{0.5} 15.1 & \dab{9.1} 62.9 & \da{11.1} 4.5 & \dab{0.1} 71.9 & \ua{0.2} 15.8 & \dab{0.1} 71.9 & \da{0.3} 15.3 & \dab{0.5} 71.5 & \da{0.2} 15.4 & 72.0 & \da{0.2} 15.4 & 72.0 & \da{0.2} 15.4 & \dab{0.2} 71.8 & \da{0.1} 15.5 \\
\bottomrule

\end{tabular}
}

\caption{Comprehensive comparison of crosslingual reputation debiasing approaches on MSEFair, showing performance across different source-target language combinations for various debiasing methods. This analysis reveals the limitations of traditional cross-lingual approaches when dealing with typologically different languages like Russian.}
\label{appendix:crosslingual-sefair}
\end{table*}

\end{document}